\documentclass[12pt]{article}

\usepackage{newtxtext}
\usepackage{graphicx}

\usepackage[letterpaper,margin=1in]{geometry}
\usepackage{gensymb}

\usepackage{algorithm}
\usepackage{algpseudocode}

\usepackage{tabularx}
\usepackage{amsmath}
\usepackage{amsfonts}

\usepackage{multirow}
\usepackage[table]{xcolor}  % 导入 colortbl 宏包
\usepackage{booktabs}

\usepackage{rotating}

\usepackage{float}
\usepackage{placeins}

\usepackage{caption}
\usepackage{setspace}
\usepackage{afterpage}

\usepackage{ulem}
\renewenvironment{abstract}
	{\quotation}
	{\endquotation}

\date{}

\makeatletter
\renewcommand{\fnum@figure}{\textbf{Fig \thefigure}}
\renewcommand{\fnum@table}{\textbf{Table \thetable}}
\makeatother

\usepackage{scicite}

\usepackage{url}

\def\scititle{
Precise Aggressive Aerial Maneuvers with Sensorimotor Policies
}
\title{\bfseries \boldmath \scititle}

\author{
	Tianyue Wu$^{1\dagger}$,
	Guangtong Xu$^{2\dagger}$,
	Zihan Wang$^{2}$,
        Junxiao Lin$^{1}$,
        Tianyang Chen$^{1}$, \and
        Yuze Wu$^{1}$,
        Zhichao Han$^{1}$, 
        Zhiyang Liu$^{3}$,
        Fei Gao$^{1,2,3*}$ \and 
	 \small$^{1}$Institute of Cyber-Systems and Control,
\small{College of Control Science and Engineering,} \\
\small{Zhejiang University, Hangzhou, China.}\\
	\small$^{2}$Huzhou Institute of Zhejiang University, Huzhou, China. \\
    \small$^{3}$Differential Robotics Technology Co., Ltd., Hangzhou, China.\and
	\small$^\ast$Corresponding author. Email: fgaoaa@zju.edu.cn\and
	\small$^\dagger$These authors contributed equally to this work.
}

\begin{document} 

% Insert the title and author list
\maketitle

% Abstract, in bold
% There are strict length limits, and not all formats have abstracts.
% Consult the journal instructions to authors for details.
% Do not cite any references in the abstract.
\begin{abstract} \bfseries \boldmath
% Start with one or two sentences of background
Precise aggressive maneuvers with lightweight onboard sensors remains a key bottleneck in fully exploiting the maneuverability of drones. Such maneuvers are critical for expanding the systems' accessible area by navigating through narrow openings in the environment. Among the most relevant problems, a representative one is aggressive traversal through narrow gaps with quadrotors under SE(3) constraints, which require the quadrotors to leverage a momentary tilted attitude and the asymmetry of the airframe to navigate through gaps.  In this paper, we achieve such maneuvers by developing sensorimotor policies directly mapping onboard vision and proprioception into low-level control commands.  The policies are trained using reinforcement learning (RL) with end-to-end policy distillation in simulation. We mitigate the fundamental hardness of model-free RL's exploration on the restricted solution space with an initialization strategy leveraging trajectories generated by a model-based planner. Careful sim-to-real design allows the policy to control a quadrotor through narrow gaps with low clearances and high repeatability. For instance, the proposed method enables a quadrotor to navigate a rectangular gap at a 5 cm clearance, tilted at up to 90\degree \ orientation, without knowledge of the gap's position or orientation. Without training on dynamic gaps, the policy can reactively servo the quadrotor to traverse through a moving gap. The proposed method is also validated by training and deploying policies on challenging tracks of narrow gaps placed closely. The flexibility of the policy learning method is demonstrated by developing policies for geometrically diverse gaps, without relying on manually defined traversal poses and visual features.
\end{abstract}

% The first paragraph of any Science paper does NOT have a heading
% Nor is it indented
\noindent
Quadrotors are among the most maneuverable robotic systems in the world and have demonstrated remarkable ability to perform aggressive maneuvers for efficient mission accomplishment\cite{mellinger2012trajectory,richter2016polynomial,ryou2021multi,foehn2021time,kaufmann2023champion,ren2025safety}. Precise execution of these maneuvers is important for safe deployment in constrained environments, which critically depends on accurate feedback extraction for sequential decision-making. However, unlike systems that leverage external localization infrastructure to obtain reliable and direct information about the systems' states for decision making, quadrotors operating with onboard sensory data face significant challenges in acquiring state feedback with comparable accuracy and reliability.

\begin{figure}[htpb]
    \centering 
    \includegraphics[width=1\textwidth]{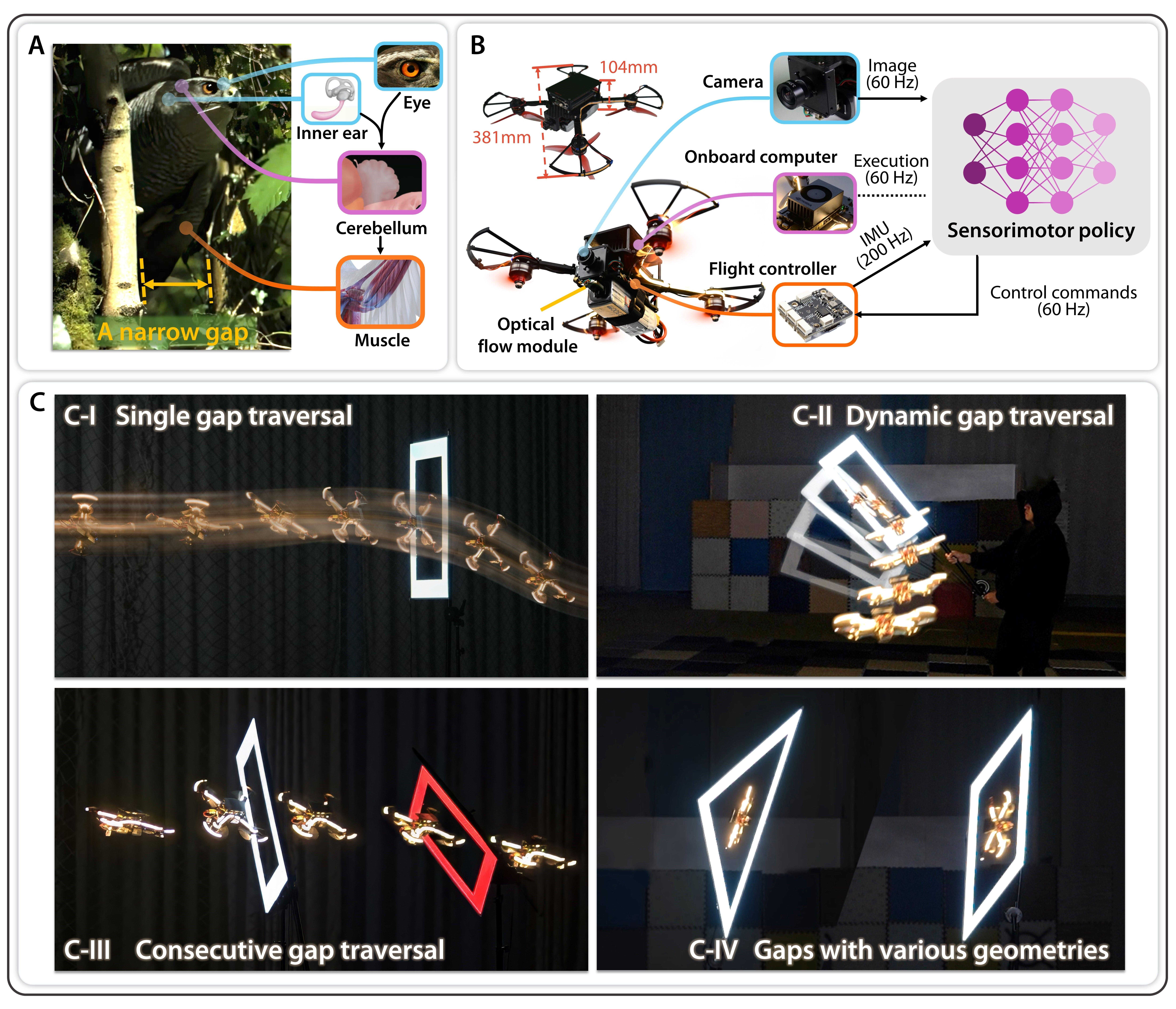}
    \caption{\textbf{Working principle and system implementation.} \textbf{(A)} Illustration of a goshawk precisely executing aggressive flight through a narrow gap between two tree trunks, guided entirely by its embodied perception system. The bird's cerebellum integrates visual input from its eyes with proprioceptive signals from its vestibular system to generate precise motor commands for gap navigation. \textbf{(B)} Illustration of the quadrotor platform equipped with a sensorimotor policy for precise aggressive flight using only onboard sensing. The policy integrates the visual sensing and proprioception that directly generate low-level control commands without explicit intermediate interfaces, such as state estimates and reference trajectory. \textbf{(C)} Key capabilities demonstrated by the proposed method through the quadrotor system: single gap traversal, dynamic gap traversal, consecutive gap traversal, and traversal through geometrically diverse gaps.}
    \label{fig:1}
\end{figure}

Despite the challenges, nature shows that the accurate sensorimotor capabilities are excellently achieved: birds demonstrate remarkable proficiency in aggressively navigating tight spaces comparable to their body dimensions, guided by their built-in sensory systems \cite{schiffner2014minding,henningsson2022flying,badger2023sideways}, as illustrated in Fig.~\ref{fig:1}A. This capability enables them to fully exploit spatial resources in complex environments and confer survival advantages in highly structured habitats \cite{martin1993nest,robinson1995regional}. Similarly, to fully leverage the maneuverability of autonomous quadrotors and maximize their accessible mission space, these robots are expected to freely perform precise aggressive flight through constrained apertures, such as building windows, inter-tree gaps, or cave entrances, despite the small clearances.  One of the most compelling demonstrations of such capabilities lies in performing highly dynamic whole-body maneuvers, where the system strategically rolls or pitches its airframe to aggressively traverse narrow openings that are impossible to traverse in a near-hovering state \cite{mellinger2012trajectory,wang2022geometrically}. 

This class of problems abstracts into the canonical challenge of \emph{narrow gap traversal} \cite{mellinger2012trajectory,loianno2016estimation,falanga2017aggressive,wang2022geometrically}, which requires the autonomous quadrotor to accurately determine both the right pose and timing to traverse a narrow passable space under tightly coupled translational and rotational dynamics. However, most previous work assumes full observability of the problem and achieves gap traversal through external localization systems that provide accurate real-time tracking of the quadrotor's state \cite{mellinger2012trajectory,wang2022geometrically,lin2019flying,xiao2021flying}. In the past decade, thanks to the development of computer vision and sensor fusion techniques \cite{bloesch2015robust,leutenegger2015keyframe}, a few efforts \cite{loianno2016estimation,falanga2017aggressive} have achieved narrow gap traversal without external localization devices using visual-inertial odometry \cite{leutenegger2015keyframe} for state estimation. However, the approach presented in \cite{loianno2016estimation} relies on a pre-specified traversal trajectory that is planned before deployment without real-time gap awareness, and the trajectory is not dynamically replanned to account for perception uncertainty \cite{smith1990estimating} and tracking deviations. To compensate for this, the researchers in \cite{falanga2017aggressive} design a two-stage rule-based trajectory planner and handcrafted visual features for rectangular gap traversal, achieving real-time replanning, gap pose detection, and successful traversals through gaps up to 45\degree \ orientations at a clearance of 8 cm.  The impressive demonstrations in \cite{falanga2017aggressive}, however, come at the cost of an algorithmic design overfitted to the setup: the method becomes inconvenient and suboptimal when extended to other variants of precise aggressive maneuvers, such as consecutive gap traversal or navigating gaps with varying geometries.  Moreover, the aforementioned approaches rely on artificially partitioned modules, such as state estimation and trajectory optimization \cite{song2023reaching,ren2023online}. Each module is independently tuned and optimized \cite{loianno2016estimation,falanga2017aggressive,ren2023online}, and the manually defined interfaces between them inevitably introduce information loss and cascading errors \cite{song2023reaching,hu2023planning,levine2016end}. While such errors are tolerable for applications with relaxed precision requirements, they can be catastrophic during aggressive flight through narrow gaps with low fault tolerance.

{This motivates a fundamental design question: Do we have to adhere the modular architecture with artificial interfaces between modules for achieving precise aggressive maneuvers through confined gaps? {The direct sensing-acting loops in biological systems} implies an alternative formulation}: birds elegantly execute aggressive navigation through narrow apertures without explicit reliance on gap pose estimation, odometric integration, or trajectory planning. As illustrated in Fig. \ref{fig:1}A, their embodied  {systems develop experience-driven memorization} {to perform reactive control that directly transforms visual and vestibular inputs into motor commands,} { achieving sensorimotor maneuvers at levels still unattainable by current engineering.} {This observation motivates us to investigate a computational approach that pursues such a direct data-driven sensing-acting mapping without explicit error propagation and static intermediate representations.}
Relevant ideas, known as sensorimotor/visuomotor learning \cite{levine2016end}, have shown promise in robotics domains such as object manipulation \cite{levine2016end}, perceptive legged locomotion \cite{agarwal2023legged}, and high-speed drone racing \cite{geles2024demonstrating}. However, these applications are either quasi-static \cite{levine2016end}, benefit from high degrees of freedom actuation thus enabling error recovery \cite{agarwal2023legged}, or enjoy a relatively forgiving solution space \cite{geles2024demonstrating}. In particular, \cite{geles2024demonstrating} is the most related work to our research. We adopt a binarized landmarks observation to surrogate raw monocular image similar to (30) to enable lightweight perception sim-to-real, and also implement an end-to-end design philosophy for autonomous quadrotor navigation. However, the problem addressed in this work introduces a unique challenge: traversal through narrow gaps requires the quadrotor to exploit the asymmetric airframe geometry and momentary tilted pose, introducing strict, non-convex constraints in special Euclidean group ($\mathrm{SE(3)}$) between the collider of quadrotor and the environment \cite{wang2022geometrically}. This differs from drone racing scenarios that consider only relative relaxed positional constraints \cite{geles2024demonstrating} but pursue time optimality. The question still remains whether a sensorimotor policy learning approach can succeed under the stringent constraints of highly dynamic maneuvers through narrow gaps where control tolerance is limited.

Here we present a sensorimotor policy learning framework that integrates reinforcement learning (RL) \cite{sutton1998reinforcement,minsky1954theory} with policy distillation \cite{rusu2015policy} in a sim-to-real transfer paradigm to achieve precise aggressive maneuvers. This method leveraging end-to-end, system-level gradient propagation to optimize the entire control pipeline \cite{levine2016end}.  Rather than employing engineered feature extraction pipelines, our method autonomously develops task-relevant representations through experiential learning \cite{herzfeld2014memory,suhaimi2022representation}. Through sensorimotor learning, the resulted policies achieve diverse precise aggressive maneuvers through narrow gaps using only onboard vision and proprioception (see Movie S1). The learned policies directly map high-dimensional visual signal and inertial data to low-level control commands (collective thrust and bodyrates), mirroring the seamless sensorimotor integration observed in avian flight.

However, direct policy training with high-dimensional and recurrent observations poses significant challenges. The complex observation space, combined with the narrow feasible solution space, creates substantial inefficiency of RL's exploration. We address the observation complexity through a decoupling approach \cite{chen2020learning} that divides the original problem into two stages: (i) identifying a robust solution for an oracle Markov Decision Process (MDP) using RL, and (ii) establishing a mapping from historical pixel observations to actions through policy distillation \cite{rusu2015policy} via online imitation learning \cite{ross2011reduction}. Even with a low-dimensional oracle observation space, exploration for feasible solutions using general model-free RL algorithms remains challenging due to the narrow feasible solution space, which poses a significant barrier to efficient learning. To address this, we leverage model-based trajectory optimization \cite{wang2022geometrically} in simulation to generate open-loop trajectories using differentiable flatness dynamics \cite{mellinger2011minimum} and initialize the agent at states along these trajectories to effectively guide exploration. 

Due to complex aerodynamic effects \cite{o2022neural,bauersfeld2021neurobem} and noisy actuation mechanisms caused by factors such as voltage fluctuations \cite{song2023reaching}, perfect simulation of quadrotor dynamics is challenging. In particular, the state distribution encountered during real-world flight can deviate from that experienced during simulation training, potentially resulting in poorly trained states that cause failure given the small control clearance. Moreover, without direct access to position and velocity feedback or knowledge of the traversal pose, the deployed policy must learn to infer the decision-making cues by developing (latent) task-relevant representation from high-dimensional exteroception. This representation, learned entirely in simulation, should demonstrate sufficient robustness to withstand the sim-to-real distribution shift over time. To this end, we conduct various types of randomization during training to expand the well-trained state space of the policy and learn a robust representation for extracting decision-making cues, as a key to successful deployment in the real world. 

The presented system successfully achieves gap traversal, where a quadrotor navigates through a rectangular gap measuring across various previously unknown poses and positions, with gap orientations up to 90\degree \ (Fig. \ref{fig:1}C-I), demonstrating very high repeatability. Notably, despite not being trained on dynamic gaps, the policy successfully servos the quadrotor through gaps with unpredictable motion patterns (Fig. \ref{fig:1}C-II), highlighting the policy's robust reactive control capabilities. By training policies on tracks containing multiple gaps, the system can traverse consecutive closely placed gaps, as an extreme validation for the capabilities of sensorimotor precise aggressive flight (Fig. \ref{fig:1}C-III). We also demonstrate that our method can develop policies that enable a quadrotor to navigate through narrow regions of various geometries in the real world (Fig. \ref{fig:1}C-IV), without requiring a manually defined traversal state or handcrafted visual features.
In summary, the primary contribution of this work lies in the realization of sensorimotor precise aggressive maneuvers under strict $\mathrm{SE(3)}$ constraints through end-to-end policies, and demonstrates that the resulting system achieves new results beyond the report from prior work and advanced performance for this classic robotics benchmark. The policy learning framework represents a sophisticated integration of the existing policy distillation framework with an improved RL process tailored for exploring narrow solution spaces; through systematic design choices of sim-to-real transfer components and real-world ablation, the produced policies achieve high repeatability in variants of the narrow gap traversal challenges in the real world.

\section*{Results}
Our primary interest lies in determining whether a quadrotor can precisely execute aggressive maneuvers through narrow passable regions with low clearances. The narrow passable region, often referred to as a gap throughout this work, is identified through a visual landmark, as illustrated in Fig. \ref{fig:1}C. In contrast to previous approaches that formulate the task as target state tracking \cite{mellinger2012trajectory,loianno2016estimation}, the traversal state is autonomously determined by the policy \cite{wang2022geometrically}, which is essential for non-rectangular gap traversal where optimal traversal states cannot be intuitively predefined.

We develop a custom quadrotor platform as shown in Fig. \ref{fig:1}B, with dimensions of 38cm×10cm (measured between the outermost propeller tips), to validate our proposed method. The quadrotor is equipped with a monocular camera featuring a Field of View (FoV) of 82°×72° and an onboard PX4 Autopilot flight controller \cite{px4autopilot}. All computational processing is performed on an NVIDIA Jetson Orin NX \cite{NX} integrated into the platform. The camera captures gap instances at a resolution of 1280×1024 pixels, which are subsequently downsampled to 320×256 pixels for neural network policy input. The flight controller executes thrust and bodyrate commands generated by the policy while providing real-time measurements of the quadrotor's roll and pitch attitude components. To hover the quadrotor before policy takeover, the flight controller integrates data from an optical flow module. This optical flow-based control system also stabilizes the drone after the policy autonomously triggers a recovery protocol upon complete gap traversal, as detailed in the \emph{Real-World Deployment} section.   In all the experiments, the range of the net thrust is set from 0.61 (i.e., 6$\mathrm{m/s^2}$) to 2.04 times the acceleration due to gravity (i.e., 20$\mathrm{m/s^2}$), whereas the maximum angular velocity is restricted to 6 radians per second.

To avoid frequent hardware damage, we use a hardware-in-the-loop (HIL) testbed for some of the experiments.  In the HIL testbed, we deploy the neural network policies to control a physical quadrotor. Rather than setting up physical gap instances in the environment, we use state information obtained by fusing the data from a motion capture system and an onboard Inertial Measurement Unit (IMU) in the flight controller to generate synthetic image inputs, thereby replacing direct landmark observations. We use the recorded data of the state estimates to determine whether the quadrotor traverses the narrow passable region without collision. 

Through these experimental platforms, we present comprehensive experimental results across three distinct settings, encompassing over 100 real-world trials of autonomous traversal through physical gaps. These experiments demonstrate several key capabilities empowered by our proposed method. First, we evaluate our approach on rectangular gaps with unknown orientations (including extreme angles up to 90\degree) and arbitrary spatial positions. Despite the stringent control tolerance requirement of only 5 cm for successful traversal, our proposed method achieves high success rates across diverse initial positioning conditions. Even without training on dynamic gaps, the policy develops the capability to control a quadrotor traversing through handheld moving gaps. We assess our method's performance on sequential gap configurations, where multiple gaps are positioned in close proximity. We also demonstrate that our proposed framework enables convenient development of policies capable of navigating gaps with diverse geometries using a unified objective function design and training pipeline, in contrast to the rectangular gap-specific planning method in \cite{loianno2016estimation,falanga2017aggressive}. After that, we identify the key ingredients for efficient policy training and sim-to-real transfer via ablation studies. We also implement baseline methods and conduct extensive experiments to benchmark their performance with the proposed method.

\begin{figure}[htpb]
    \centering  % 添加这行
    \includegraphics[width=1.0\textwidth]{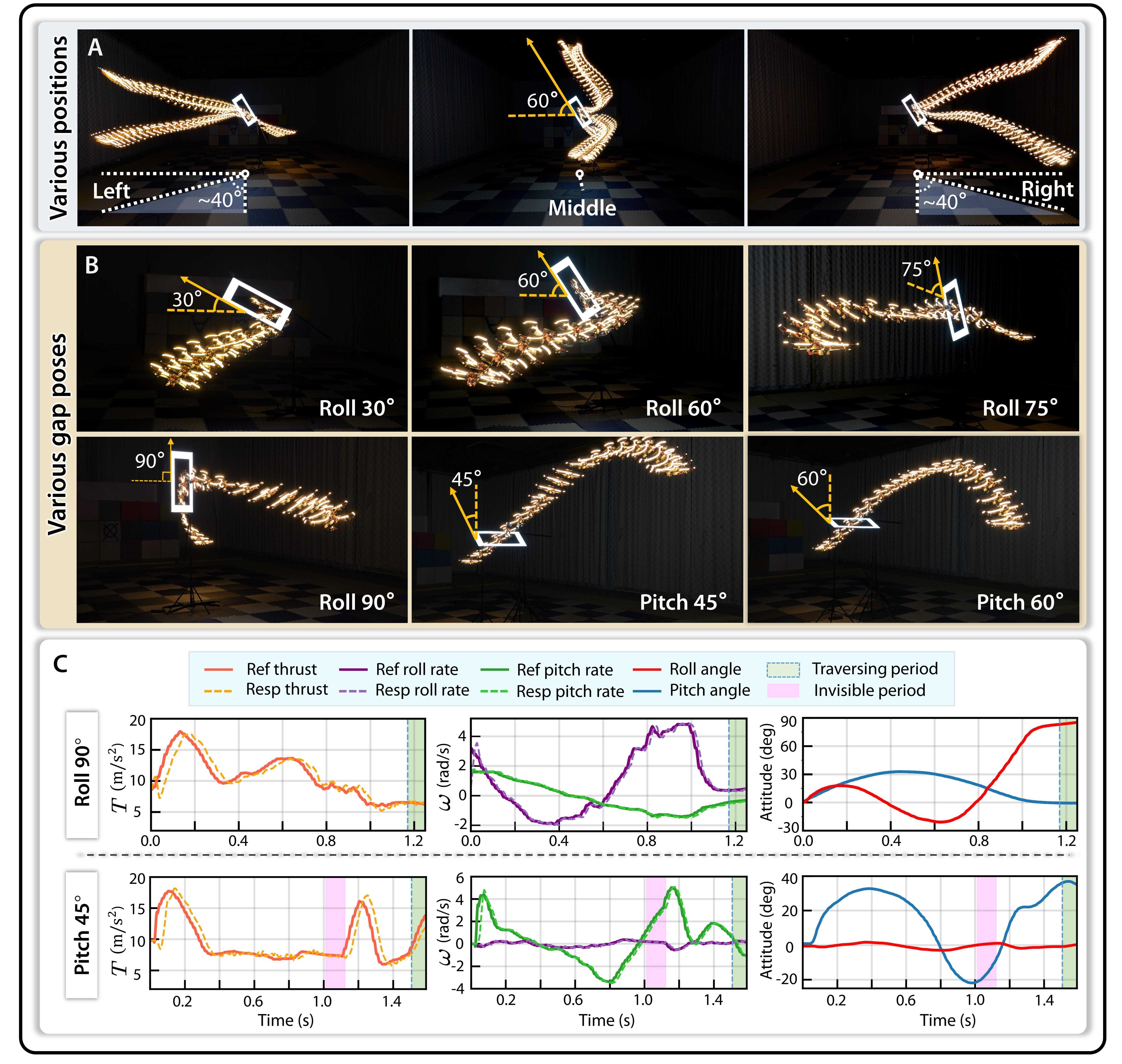}
    \vspace{-0.3cm}
    \caption{\textbf{Traversal through a rectangular gap with low clearances.} \textbf{(A)} Autonomous rollouts from a large range of initial positions through a gap with an unknown but constant tilted pose with a ground-truth of 60\degree. \textbf{(B)} Autonomous rollouts through a gap with different orientations. \textbf{(C)} Control commands output by the policy and measured responses in two trials from \textbf{(B)}: 90° rolled gap (top row) and 45\degree \ pitched gap (bottom row). In the figures, the reference control commands (policy output) are denoted as ``Ref'', and the measured responses of the corresponding quantities are denoted as ``Resp". The third column shows the measured roll and pitch angles. We plot the traversal periods as indicated by the legend, defined as starting when the vehicle first enters the gap plane and ending when it exits the gap plane. This period is determined by analyzing the recorded video. We also plot an ``invisible period" defined as the duration when the entire landmark is out of the FoV, if it occurs, prior to the traversal period.}
    \label{fig:2}
\end{figure}

\subsection*{Traversal through a Rectangular Narrow Gap with Low Clearances}
We first evaluate the proposed method on a rectangular gap with a passable area of $\mathrm{20 \ cm \times 60 \ cm}$. When the quadrotor is centered within the gap, the clearance tolerance along the short sides is only 5 cm, accounting for the vehicle's 10 cm height. This experimental setup is similar to that described in \cite{wang2022geometrically}, while a motion capture system is utilized to provide accurate quadrotor state tracking in \cite{wang2022geometrically}. Video recording for selected trials of the following experiments is provided in Movie S2. 

In Fig. \ref{fig:2}A, we show trajectories from various starting positions traversing a 60° tilted gap, where the tilted orientation is unknown prior to deployment. Some starting positions, such as the left and right ones shown in Fig. \ref{fig:2}A, are positioned approximately 40° off-axis from the gap's center line and at a distance of roughly 4m from the gap center. From a state estimation perspective, such distances amplify estimation uncertainty \cite{falanga2017aggressive}. The sensorimotor policy eliminates the need for accurate gap pose detection and localization relative to the gap, leveraging end-to-end optimization and domain randomization \cite{peng2018sim} to naturally accommodate varying perception uncertainties during flight.

Fig. \ref{fig:2}B presents trajectory rollouts through tilted gaps with various unknown roll angles. In the experimental trials, the policy achieves nearly perfect performance for gap roll angles of 60\degree or less, with 29 successful trials out of 30 total attempts. Performance slightly degrades to 90\% success rate (27 out of 30 trials) when the roll angle exceeds 60\degree.  The rollout trajectories reveal that although we do not explicitly enforce traversal pose constraints, the system autonomously aligns its body's longitudinal axis parallel to the rectangular gap's edge at the moment of passage through the gap plane, achieving this without explicit gap pose detection.  The control inputs for traversal through a 90\degree\ tilted gap are presented in Fig. \ref{fig:2}C (the top row). In this case, the policy drives the x-axis body rate to the predefined limit of 6 rad/s to navigate through the gap, enabling the quadrotor to achieve roll angles approaching 90\degree \  during traversal. At near-vertical orientations, the quadrotor can barely generate upward lift, making decision-making errors liable to cause collisions with either the long or short edges of the frame. Despite these inherent limitations of quadrotors, the policy successfully prevents such failures in the majority of experimental trials, demonstrating exceptional capabilities in achieving precise sensorimotor control.

We achieve traversal through pitched narrow space by training a separate RL policy. The deployment success rate when the pitch angle of the gap is set as 30\degree \ is 100\%, while degrading to 80\% when the pitch angles are 45\degree, and 73.3\% for 60\degree, with 15 trials for each case. The control command curves in Fig. \ref{fig:2}C (the second row) reveal that the policy learns to initially propel the quadrotor forward by increasing the pitch angle, then delicately adjusts both pitch angle and thrust to horizontally decelerate before safely navigating through the gap. The limited vertical FoV occasionally causes complete loss of gap perception (the ``invisible period'' in Fig. \ref{fig:2}C), yet the quadrotor successfully navigates through the gap in the majority of experiments. This performance is attributed to the learned belief state \cite{miki2022learning} output by the recurrent neural network (RNN) described in the \emph{Observation Space Distillation via Supervised Learning} section. Our experiments reveal that the primary failure mode is collisions between the quadrotor's upper section and the gap structure. These failures typically arise from the absence of explicit velocity feedback, which, when coupled with potential target detection loss, impairs precise x-axis velocity control within the body frame.

\subsection*{Reactive Dynamic Gap Traversal}
We also conduct experiments involving traversal through dynamic gaps, where we demonstrate that even without training in scenarios featuring moving landmarks, a policy is capable of controlling the quadrotor through gaps with unknown movements. Specifically, a handheld gap is manipulated to undergo translational or rotational motion during the quadrotor's flight, as illustrated in Fig. \ref{fig:3}A and more trials in Movie S3.  The first row of Fig. \ref{fig:3}A represents an experiment in which the narrow gap remains stationary initially, and after the quadrotor has flown for a certain distance, the frame is deliberately rotated. Despite this rotational disturbance, the quadrotor successfully adapts its flight trajectory and maintains a precise traversal pose, as demonstrated in the final frame showing successful gap navigation. The experiment shown in the bottom row of Fig. \ref{fig:3}A examines translational gap movement. When the gap is manipulated to move upwards, the policy demonstrates effective tracking behavior, where the quadrotor ascends to follow the gap's upward motion while approaching the gap plane for the final successful traversal. 

\begin{figure}[htpb]
    \centering  % 添加这行
    \includegraphics[width=1.0\textwidth]{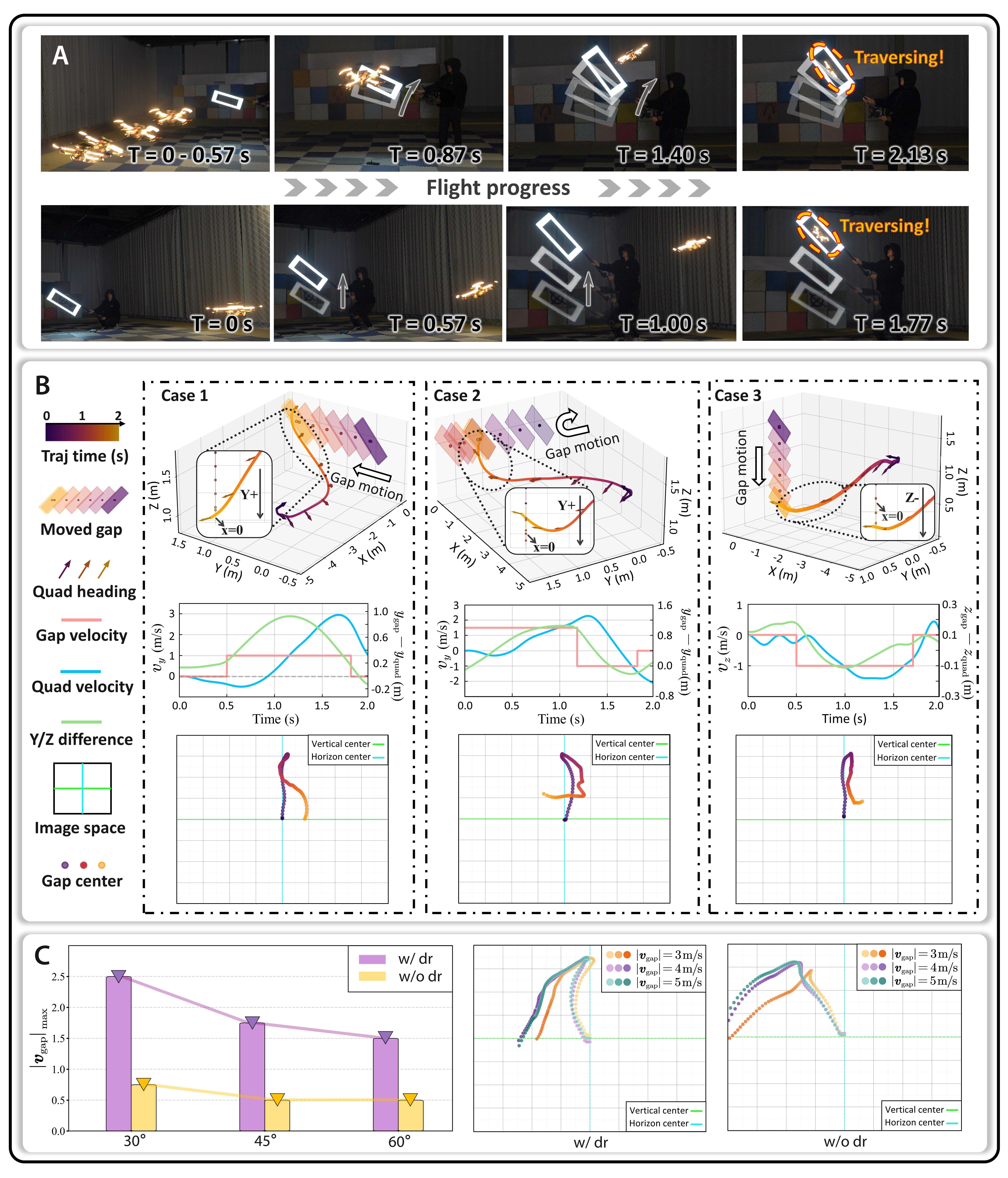}
    \captionof{figure}{\textbf{Traversal through a dynamic gap.} \textbf{(A)} Snapshots of dynamic gap traversal during flight. Each row represents temporal progression from left to right. The top row shows snapshots during flight through a rotating gap. The bottom shows snapshots during flight through an upward-moving gap. The transparent frames indicate the historical positions of the gap.  \textbf{(B)} Three simulated trials through differently moving gaps. Case 1 to Case 3 represent experiments involving a horizontal unidirectional moving gap, a horizontal bidirectional moving gap,  a downward moving gap, respectively. The figures in the first row are the spatial trajectories during the flight, where we highlight}
    \label{fig:3}
\end{figure}
\begin{figure}[h]
\ContinuedFloat
\caption[]{the X-Y view (Case 1 and Case 2) and the Y-Z view (Case 3) of the trajectories when the quadrotor approaches and traverses the gap. The middle row of figures plots the y-axis (Case 1 and Case 2) and z-axis (Case 3) velocities of the gap and quadrotor, and y-axis position difference between gap and quadrotor ($y_{\mathrm{gap}}-y_{\mathrm{quad}}$; Case 1 and Case 2) and z-axis position difference ($z_{\mathrm{gap}}-z_{\mathrm{quad}}$; Case 3). The bottom row of images shows trajectories of the rectangular gap's projected center point on image space, where the data corresponding to distances less than 0.5m between the quadrotor and the gap plane is omitted. (C) Capability limits with and without domain randomization. The left panel shows the maximum traversable gap velocity for each gap orientation. The middle and right panels visualize gap center trajectories in image space under high-speed gap motion (3$\sim$5 m/s) for policies trained with (middle) and without (right) domain randomization, where trajectory points progress from lightest (earliest) to darkest (latest) shading.}
\end{figure}

We conduct controlled experiments in simulation with results shown in Fig. \ref{fig:3}B, visualizing the spatial trajectories of both the gap and quadrotor (top row), temporal profiles of gap and quadrotor's motion (middle row), and the projected gap center trajectory on the image plane (bottom row). The three experimental cases examine horizontal unidirectional motion, horizontal bidirectional motion, and downward motion of the gap, respectively. The results demonstrate that the policy consistently exhibits robust tracking capability regardless of gap motion patterns. As shown in the bottom row of figures, when the quadrotor approaches the moving gap, the gap center consistently remains near the horizontal centerline of the image, indicating that the learned policy effectively maintains visual alignment with the target throughout the maneuver. This behavior confirms that the perception-aware behavior implemented through the reward structure (see the \emph{Reward Formulas} section in the Supplementary Material) is well-performed, even in the dynamic gap scenario, which is unseen during training. The green curves in the middle row of Fig. \ref{fig:3}B show that the servo-like behavior arises from the coordinated interplay of position tracking and active heading compensation, rather than simple position alignment.

These experiments demonstrate that the policy successfully learns task-relevant representations for goal-conditioned flight while exhibiting strong reactive control capabilities that respond to real-time observations rather than merely reproducing learned trajectories in simulation. To understand what enables this reactive capability, we conduct controlled ablation experiments comparing policies distilled with and without domain randomization (both using the same RL expert trained with domain randomization).

The first experiment evaluates maximum traversable gap speeds across different gap orientations. Initial conditions are standardized, where the quadrotor starts 5 meters from the gap with the gap center aligned to the FoV center. The gap moves laterally at constant velocity parallel to the ground and is located within the gap plane (varied across trials in 0.25 m/s increments). Results (left panel, Fig. 3C) reveal a striking performance gap: the policy employing domain randomization exhibits significantly stronger dynamic target adaptation capabilities compared to its non-randomized counterpart.  Moreover, at extreme speeds ($\geq$3 m/s), the policy distilled under domain randomization maintains the gap center within the FoV (Fig. 3C, middle), while the policy without domain randomization experiences rapid gap center trajectories drift out of view (Fig. 3C, right).

Based on these results, we conclude that domain randomization is the key enabler of reactive behavior, by expanding the observation sequence distribution to cover diverse relative-motion patterns between the quadrotor and the target during training. This adaptive capability cannot be achieved through methods that rely on pre-planned trajectories without replanning \cite{mellinger2011minimum,loianno2016estimation,wang2022geometrically}. To our knowledge, this represents the first demonstration of dynamic narrow traversal using exclusively onboard sensing.  Failure mode analysis for this purely reactive approach is provided in Supplementary Materials, the \emph{Failure Modes in Dynamic Gap Traversal} section.

\begin{figure}[htpb]
    \centering 
    \includegraphics[width=1\textwidth]{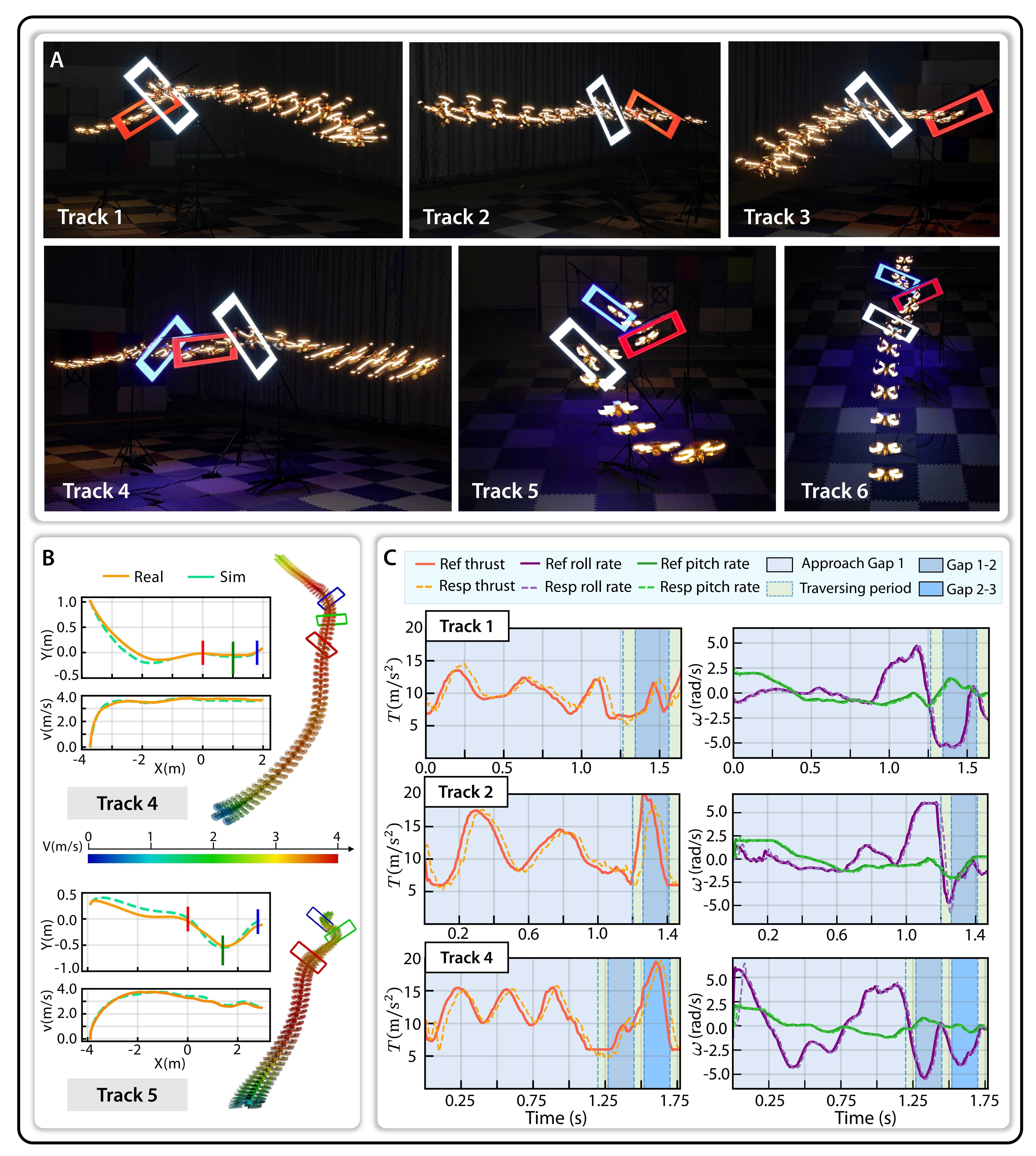}
    \caption{\textbf{Traversal through consecutive gaps with low clearances.} \textbf{(A)} Autonomous rollouts of policies across 6 different tracks containing two or three closely positioned gaps. \textbf{(B)} Comparison of real-world and simulated trajectories (position and velocity magnitude) for Track 4 and Track 5, with real-world data recorded using the HIL setup. Trajectory color represents velocity magnitude. \textbf{(C)} Policy outputs and measured system responses during experiments. The three rows correspond to results from Track 1, Track 2, and Track 4, respectively. Legend abbreviations follow the same conventions as Fig. \ref{fig:2}C, with additional color blocks defined as follows: ``Approach Gap 1" denotes the pre-traversal period before the first gap, ``Gap 1-2" represents the inter-gap period between the first and second gaps, and ``Gap 2-3" indicates the period between the second and third gaps.}
    \label{fig:4}
\end{figure}

\subsection*{Traversal through Consecutive Narrow Gaps with Low Clearances}
Traversing consecutive gaps positioned in close proximity presents greater challenges for both RL exploration and real-world deployment compared to single gap traversal.  In this section, we present, to our best knowledge, the first demonstration of precise traversal through consecutive narrow gaps using only onboard sensing in the real world (see Movie S4). The objective function design and training pipeline used in the above section are easily extended to this setup. 

We design tracks consisting of two or three $\mathrm{20 \ cm \times 60 \ cm}$ rectangular gaps that ensure the existence of feasible open-loop solutions using trajectory optimization methods. During training, the poses and positions of the gaps are slightly randomized for each track configuration. This approach eliminates the need for highly accurate measurement of relative positions and orientations between gaps during real-world deployment. Fig. \ref{fig:4} illustrates the track configurations, trajectory rollouts generated by the policies (Fig. \ref{fig:4}A), and the recorded states (Fig. \ref{fig:4}B) and control commands (Fig. \ref{fig:4}C). Table \ref{tab:S1} provides a quantitative description of the track configurations during training. 

In Tracks 1 to 3, there are two gaps placed in the tracks. For instance, along Track 1, where the difference in tilted roll angles between the two gaps reaches approximately 75\degree \ and the inter-gap distance is merely 0.8m, the policy drives the roll rate to the predefined limit of 6 rad/s to successfully navigate the second gap, as demonstrated by the command trajectories in Fig. \ref{fig:4} (top row). In the track shown in Track 2, since the first gap is tilted at 60\degree, the quadrotor executes a sharp roll to approximately 60\degree \ when traversing the gap, inevitably losing altitude control due to the downward gravitational acceleration. The policy must therefore time the roll precisely: as shown in Fig. \ref{fig:4}C (the middle row), the policy begins increasing the roll rate to the predefined limit only when approaching the first gap, then rapidly increases thrust after the entire airframe passes through the first gap to prevent excessive altitude loss and ensure safe traversal of the second gap.

In Tracks 4 to 6, there are three gaps to traverse in the tracks where we significantly stagger the gaps laterally in Track 5 and Track 6. For laterally staggered tracks, beyond the requirement for precise attitude responses, the quadrotor needs to finely laterally maneuver for successful traversal. Consequently, these tracks are more prone to collisions between the quadrotor's propellers and the gaps compared to tracks where the centerlines of each gap are more aligned, such as Track 4. Nevertheless, the policy demonstrates high repeatability in controlling the quadrotor through these gaps by executing precise attitude maneuvers and velocity modulation at critical moments. Sim-to-real comparison in HIL setup in Fig. \ref{fig:4}B. The real-world data is recorded using the HIL deployment setup. The results demonstrate that rollout trajectories in simulation and the real world are similar, with differences remaining bounded rather than accumulating, thereby validating the closed-loop control capability of the policy. The velocity curves corresponding to Track 5 in Fig. \ref{fig:4}B reveal that the policy can precisely reproduce the deceleration behavior observed in simulation, enhancing the opportunity to execute sufficient lateral maneuvering while performing sharp body rolls to traverse subsequent gaps.

\begin{figure}[htpb]
    \centering 
    \includegraphics[width=1\textwidth]{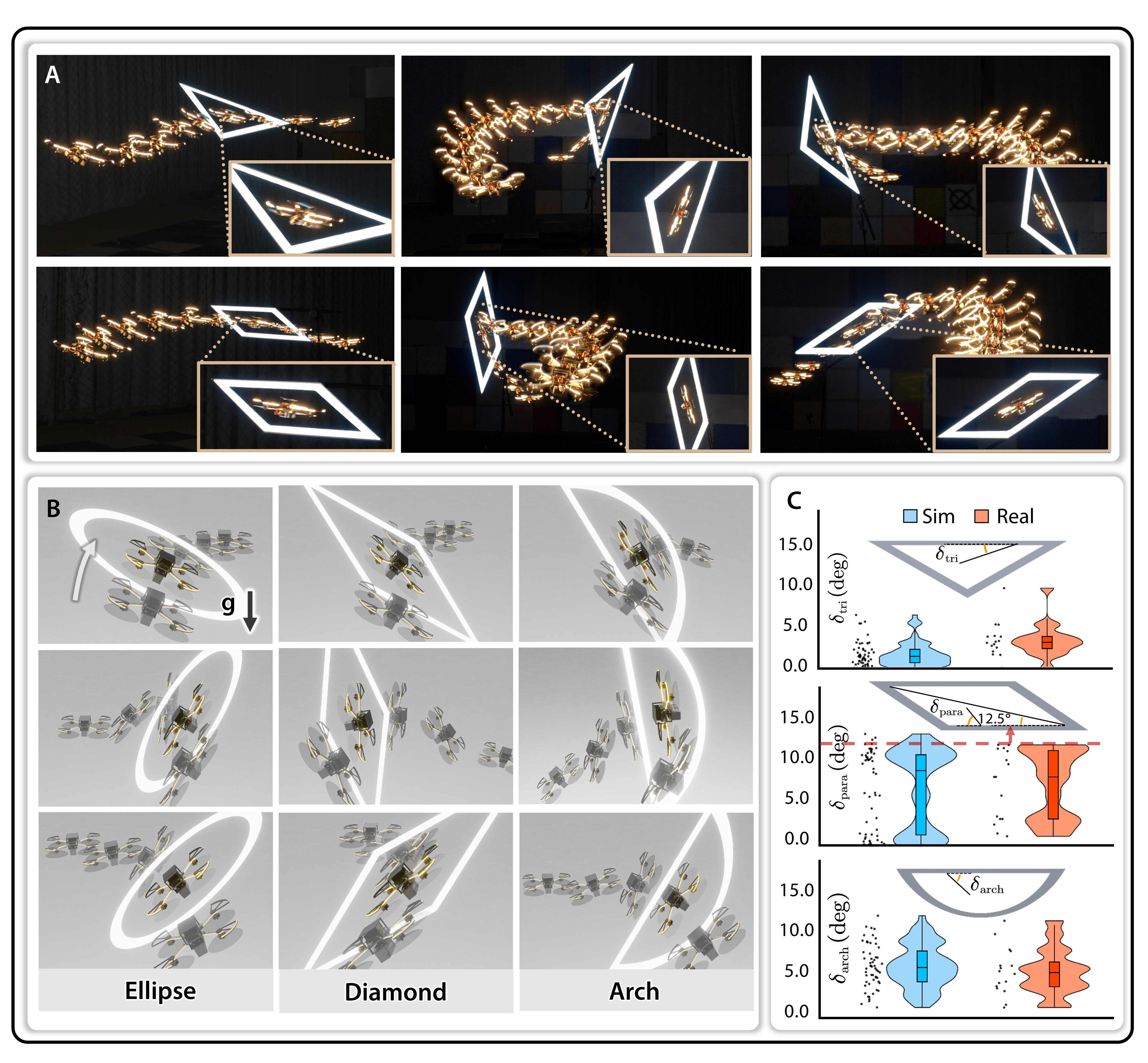}
    % \caption{\textbf{Fig. S1 More hardware-in-the-loop trajectories.} More results for gaps with ellipse (Fig_\textbf{(A)}), diamond (Fig_\textbf{(B)}), and arch (Fig_\textbf{(C)}) shapes.}
    \caption{\textbf{Traversal through narrow openings with various geometries.}  \textbf{(A)} Autonomous rollouts through physical triangular and parallelogram gaps. We highlight the state of the quadrotor when it is traversing the gap. \textbf{(B)} Simulated rollouts through elliptical, diamond-shaped, and arch-shaped gaps. Trajectories are visualized using Blender \cite{blender}, with the quadrotor's traversal frame highlighted and other trajectory points shown with white masks. The traversal frame is defined as the point on the trajectory closest to the gap plane. \textbf{(C)} Distribution of traversal roll angles relative to the gaps in simulation and HIL (real-world) experiments. The reference angles used for statistical analysis are marked in the figures for each type of gap.}
    \label{fig:5}
\end{figure}

\subsection*{Traversal through Narrow Passable Regions with Various Geometries}
In real-world scenarios, gap geometries vary significantly and optimal traversal orientations are not always apparent, motivating policies trained on diverse gap shapes. In this section, we demonstrate policies trained on diverse geometries of passable regions and landmark appearances, showcasing the flexibility of the proposed policy learning method in achieving various precise aggressive maneuvers.  The geometric parameters of the narrow passable regions are presented in Table \ref{tab:S2}.  Rollout trajectories through physical triangular and parallelogram gaps are shown in Fig. \ref{fig:5}A, with additional trials at various tilt angles presented in Movie S5. Simulated demonstrations through elliptical, diamond-shaped, and arch-shaped gaps are presented in Fig. \ref{fig:5}B. More and complete simulated trajectories are shown in Fig. \ref{fig:S1}  

For triangular gaps, we observe that the traversal orientations relative to the gap are highly consistent, with the quadrotor aligning its body-fixed x-y plane parallel to the triangle's longest edge, as shown in the top row of Fig. \ref{fig:5}A. This behavior is quantitatively validated in the upper panel of Fig. \ref{fig:5}C, which shows small angular deviations (predominantly $<$ 5\degree) in both simulation and real-world experiments. In contrast, the middle panel of Fig. \ref{fig:5}C reveals that traversal orientations through parallelogram gaps exhibit multimodality. The quadrotor adopts one of two preferred orientations: alignment with either the parallelogram's long edge or its long diagonal. Intermediate orientations occasionally occur when the parallelogram's long edge is nearly horizontal, as demonstrated in the bottom-left panel of Fig. \ref{fig:5}A.

The lower panel of Fig. \ref{fig:5}C displays the statistical distribution of traversal poses through the arch-shaped gap, revealing greater dispersion in relative traversal poses compared to the other gaps. This variability of the traversal poses reflects the relatively weaker geometric constraints imposed by the arch's feasible traversal space on vehicle orientation. Unlike vertex-based state estimation methods (e.g., PnP \cite{haralick1989pose}) that require manual feature design and struggle with elliptical or arch landmarks, the flexible sensorimotor policy learning method proposed in this work can automatically extract landmark observation representations without handcrafted features, making it more convenient to extend to general visual landmarks in practical applications.

\subsection*{Validation under Learned Segmentation as Noisier Landmark Observations}
The main results rely on illuminated frames for robust color-threshold segmentation, limiting applicability to simplified settings that allows focused validation of the end-to-end control approach. To evaluate robustness under more realistic perceptual conditions, we test the same policy on a non-illuminated rectangular gap in visually diverse backgrounds where color segmentation fails.
	
\begin{figure}[htpb]
    \centering 
    \includegraphics[width=1\textwidth]{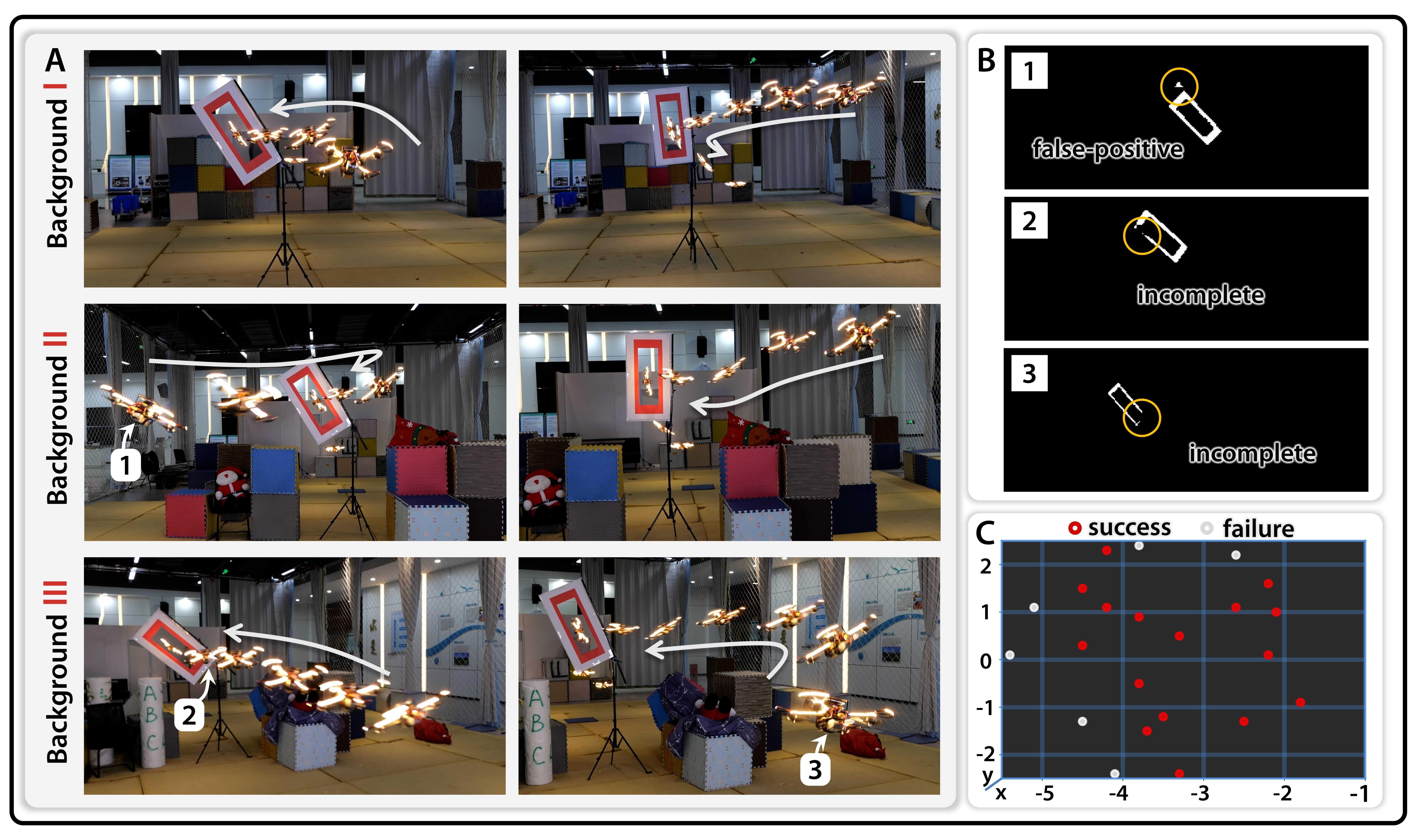}
    \caption{{\textbf{Narrow gap traversal with learned segmentation in different visual backgrounds.} \textbf{(A)} Autonomous rollouts with 3 different visual backgrounds with different gap orientations. \textbf{(B)} Flawed segmentation results during flight, where the labeled numbers in the panels correspond to the snapshots of the quadrotor marked in (A). \textbf{(C)} Records of experimental success, with each point indicating the approximately measured horizontal (X-Y) starting position. Among these trials, the gap orientation is greater than or equal to 60\degree and the gap is positioned at (0,0) while its lies in $x=0$ plane. Red points denote successful trials, while white indicate unsuccessful ones.}}
    \label{fig:6}
\end{figure}

A lightweight segmentation model consisted of MobileNetv3 encoder \cite{howard2019searching} and Atrous Spatial Pyramid Pooling (ASPP) module \cite{chen2018encoder} is trained and deployed for the experiments. The network is ported to TensorRT, and inference takes an average of 4 ms on the onboard device. As the deployment environments are unseen during model training and the labeled data is limited, the learned segmentation produces notably imperfect outputs like larger mask edge errors, false positives from background clutter (the top panel of Fig. \ref{fig:6}B), and occasional incomplete landmark observations (the middle and bottom panels of Fig. \ref{fig:6}B) compared to the illuminated-gap setup.
	
Results presented in Fig. \ref{fig:6}C demonstrate that at a relatively close range (2-4 meters), the policy succeeds in most trials with noisy masks with the gap's orientation equal to or greater than 60\degree. At longer ranges or under skewed perspectives, however, failures increase as segmentation quality degrades, leading to suboptimal early-phase trajectories whose errors accumulate into the final traversal. Nevertheless, these results validate that the learned control behaviors tolerate realistic segmentation noise, a necessary prerequisite for extension to more general perception systems.

\begin{figure}[htpb]
    \centering 
    \includegraphics[width=1\textwidth]{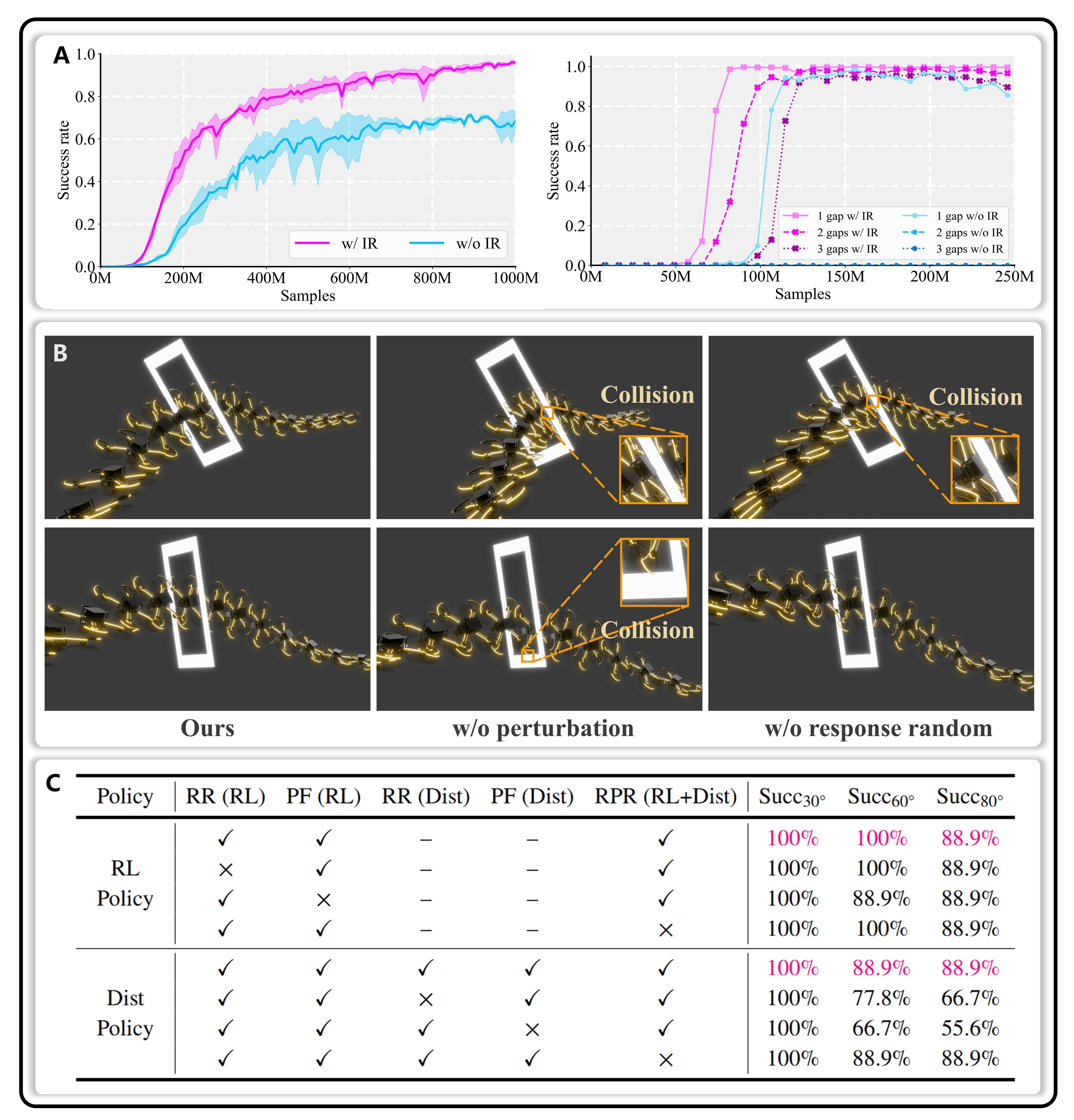}
    \caption{\textbf{Key ingredient ablation for policy learning and sim-to-real transfer.} \textbf{(A)} Ablation of the informed reset (IR) strategy. The left picture shows the success rate evolution for RL training on a rectangular gap, where we compute the mean (solid line) and standard deviation (shadow region) across three trials, and the right shows success rate evolution on Track 5 including three gaps. In the right picture, the ``1 gap", ``2 gaps", and ``3 gaps" legends represent the rates of successful traversal of the first gap, second gap, and third gap, respectively.   \textbf{(B)} Ablation study of the sim-to-real techniques. We visualize several failure cases when ablating some components of the randomization. The three columns of the figures from left to right correspond to using all the randomization components,ablation of the {perturbation force ({PF})}, ablation of {response randomization ({RR})}, and ablation of {response parameter randomization (\textcolor{black}{RPR})}.  The experiment setups corresponding to the two rows of figures are test point 4 of a 60\degree \ tilted gap for the top row, and test point 8 of an 80\degree \ tilted gap for the bottom row of figures. The test point configuration is }
    \label{fig:7}
\end{figure}

\begin{figure}[h]
\ContinuedFloat
\caption[]{ illustrated in Fig. \ref{fig:S3}. \textbf{(C)} Ablation study quantitative results. The figure shows the impact of different randomization components on success rates across three settings where the gap is tilted at 30\degree, 60\degree, and 80\degree, respectively. Each row represents a different combination where A(B) denotes: A = the specific randomization type (RR, PF, RPR), and B = the implementation stage (RL = RL stage, Dist = distillation stage, RL+Dist = both stages). $\checkmark$ indicates the component is enabled, $\times$ indicates disabled.}
\end{figure}

\subsection*{Key Ingredients for Policy Learning and Sim-to-Real Transfer}
Beyond the vanilla teacher-student RL framework, we employ two categories of training techniques to address inefficient policy learning and sim-to-real performance gap. In this section, we validate the necessity of these features through ablation experiments. In this section, we validate the necessity of these features through ablation experiments.

\subsubsection*{Informed Reset}
Since narrow gaps constrain the feasible solution space, standard RL algorithms can suffer from inefficient exploration. To address this challenge, we propose an \emph{informed reset} (IR) scheme that strategically initializes the agent to states that facilitate effective exploration. Fig. \ref{fig:7}A illustrates the success rate evolution for two RL problems with and without IR. The left panel presents results for a rectangular $\mathrm{60 \ cm\times20 \ cm}$ gap with arbitrary roll orientation, while the right panel shows training results for Track 5 (the \emph{Traversal through Consecutive Narrow Gaps with Low Clearance} section), which contains three closely positioned gaps. 

For the single rectangular gap, policy learning without IR converges to a maximum average success rate of 70\% within a 1G sample budget, but requires approximately three times the sample budget of IR-enabled learning to reach this performance level. More importantly, IR achieves an average success rate around 96\% within the same 1G sample budget, effectively mitigating the fundamental exploration difficulties inherent in low-tolerance control problems without requiring explicit manual curriculum design \cite{xiao2021flying}. The best performance usually occurs between sample number of 1G and 1.2G. Therefore, under our device condition (as mentioned in the \emph{Methodology Overview} section), the RL stage takes around 1.5 hours.

In the consecutive gap traversal problem, IR not only reduces the sample requirements for achieving initial success rates to traverse the first gap but also addresses critical exploration challenges to traverse the subsequent gaps. Specifically, without IR, the policy fails to discover feasible solutions for successfully navigating the second narrow gap — the required pre-gap deceleration conflicts with the immediate reward structure, causing the RL algorithm \cite{schulman2017proximal} to become trapped in suboptimal solutions that prioritize immediate forward progress reward (see the \emph{Reward Formulas} section). When IR is enabled, the RL agent starts the episode in more informative states (including but not limited to hover states), thus encountering more high-return states during early exploration, receiving richer feedback about the long-term consequences of different action sequences. This enhanced state coverage enables the policy to discover the deceleration strategy and escape the local optimum.

\subsubsection*{Sim-to-Real Techniques}
As described in the \emph{Sim-to-Real Transfer for Goal-Conditioned Precise Aggressive Flight} section (hereinafter referred to as the \emph{Sim-to-Real Transfer} section), we apply several randomization techniques to achieve high repeatability in real-world deployment. In this section, we ablate some of the applied randomization components to identify their contributions to performance. Specifically, we ablate (i) perturbation force (PF), (ii) response randomization (RR), which is applied using factor $\boldsymbol{c}$ described in the \emph{Sim-to-Real Transfer} section, and (iii) response parameter randomization (RPR), which randomizes the calibrated low-level control delay parameters {$\boldsymbol{h}$} described in the \emph{Sim-to-Real Transfer} section. 
We note that a similar RR implementation is demonstrated to be important for sim-to-real transfer in previous work on RL-based drone racing \cite{kaufmann2022benchmark,song2023reaching}. It is expected to help the sim-to-real transfer, as the problem we are considering—narrow gap traversal—shares similarity with racing gate traversal or waypoint navigation in drone racing scenarios.
However, PF is not commonly used in RL-based autonomous flight, so it is currently unclear whether PF can enhance the system's performance of the task at hand in the physical world. We conduct extensive HIL experiments to evaluate sim-to-real transfer performance {with or without these components}. Typical failure cases for ablation of different components are visualized in Fig. \ref{fig:7}B. The initial position distribution and additional detailed results for this experiment are provided in Fig. \ref{fig:S3}.

The results in Fig. \ref{fig:7}C demonstrate that removing any single randomization component has little impact on the RL policies' performance, while, of particular interest, the same ablations can degrade the performance of the distillation policies. Notably, all distillation policies are supervised by the RL policy trained with both RR and PF, as illustrated in Fig. \ref{fig:7}C. The heightened sensitivity of the distillation policies to randomization design during sim-to-real transfer likely stems from its more challenging operational conditions: it must solve a partially observable problem without explicit position or velocity feedback, where historical noisy observations and dynamics gaps between simulation and reality can influence current decisions through the latent representation for extracting decision-making cues which is learned in simulation. When the tilted angle reaches 60\degree \ and 80\degree, the success rate drops significantly if RR or PF are not applied.  Analysis of failure modes reveals two primary patterns: lateral deviation causing collisions with the long edges (Fig. 7B, top row), and altitude loss during attitude adjustment causing collisions with the short edges (Fig. 7B, bottom row). From Fig. \ref{fig:7}C, we also observe that applying RPR does not significantly impact real-world performance when the other two randomization techniques are employed. We attribute this to the low-latency characteristics of our chosen control interface, collective thrust and bodyrates. For instance, our hardware platform achieves {an average} bodyrate latency of approximately 20 ms. {This relatively short latency allows the control system to issue corrective commands within subsequent control cycles, mitigating the impact on the task performance of previous command latency fluctuations.}

\subsection*{{Performance Evaluation with Reference Baseline Systems}}
{To contextualize the performance and validate capabilities beyond existing systems, we implement two representative baselines from prior work:}
		
{\textbf{Baseline 1 - Wang et al. \cite{wang2022geometrically}:} This approach uses known gap position and pose to formulate trajectory optimization incorporating $\mathrm{SE(3)}$ geometric and dynamical constraints. A reference trajectory is generated and tracked by a low-level controller with full-state feedback from an external motion capture system.}
		
{\textbf{Baseline 2 - Falanga et al. \cite{falanga2017aggressive}:} This system achieves narrow gap traversal without external localization or prior knowledge of gap orientation. It uses vision-inertial fusion for state estimation, with gap corners detected via rectangle and point detection for Perspective-n-Point (PnP) pose estimation \cite{haralick1989pose}.} {A specialized trajectory generation scheme enables rapid online replanning and traversal through gaps with up to 45\degree \ orientation and 8 cm clearance \cite{falanga2017aggressive}. The localization method in this system achieves higher accuracy as the quadrotor approaches the gap.} \\
{We postpone the implementation details of the baseline systems and a discussion of the two trajectory generation methods to the \emph{Implementation and Discussion of the Baseline Systems} section in Supplementary Materials.} 
		
Fig. \ref{fig:8}A visualizes rollout trajectories across different approaches, revealing qualitatively distinct trajectory patterns between the learned policy and model-based trajectory optimization. This difference stems primarily from optimization formulation. For instance, the baseline planners do not incorporate SE(3) perception constraints for gap visibility (unlike the RL reward scheme), and require auxiliary constraints beyond collision avoidance to enhance trajectory generation stability and real-world tracking performance.

\begin{figure}[H]
    \centering 
    \includegraphics[width=1\textwidth]{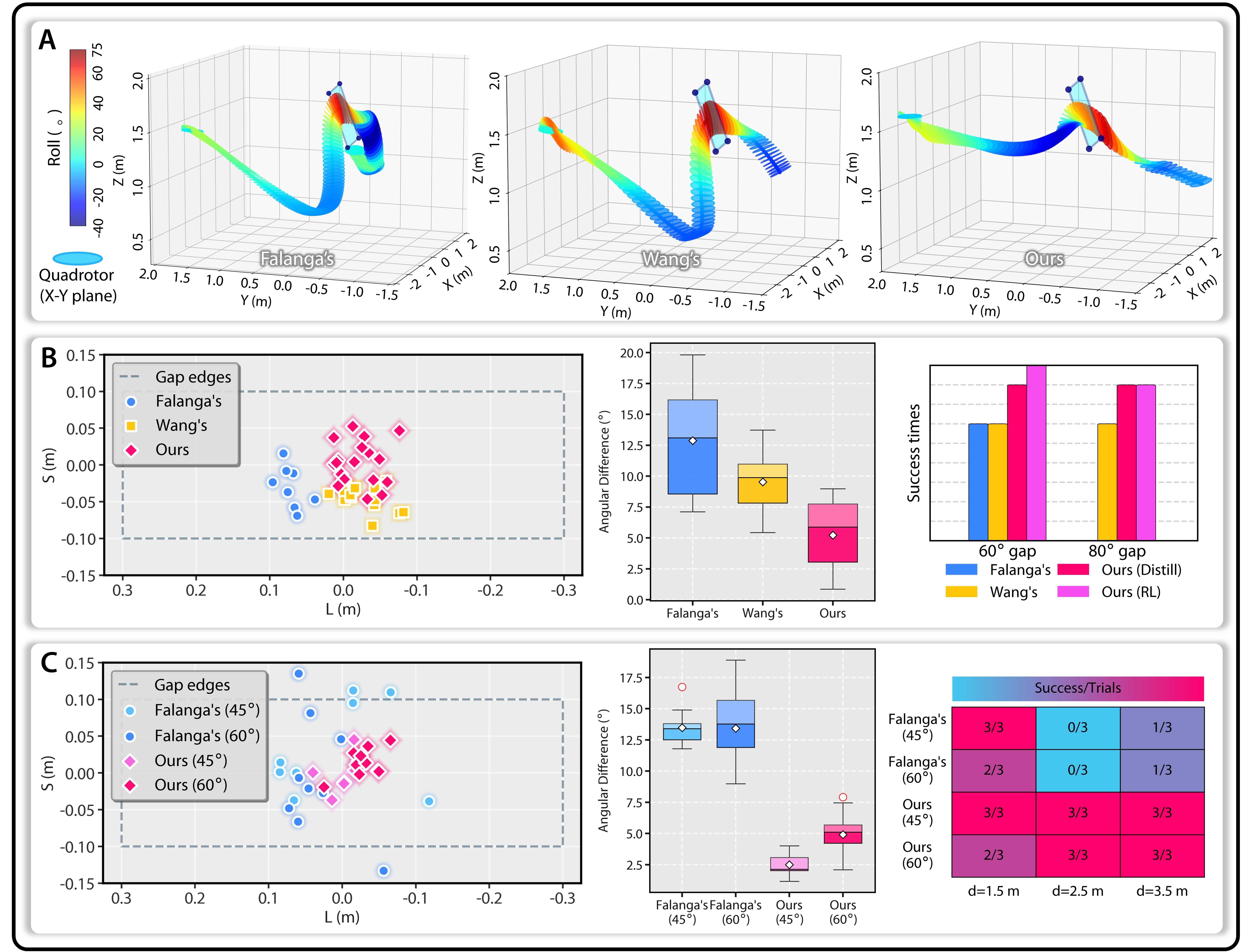}
    \caption{\textbf{Baseline performance evaluation.} \textbf{(A)} Trajectories tracking visualization of Falanga's planner, Wang's planner, and the rollout trajectory of the distilled (vision-based) policy. \textbf{(B)} Traversal position (left) and tilted angle difference (middle) relative to the gap, and success rate in 9 trials for each case. Here, Falanga’s and Wang’s indicate trajectory generation and tracking methods using privileged states, ``Ours" in the left and middle panels of the figure indicates the distilled (vision-based) policy, and ``Ours (RL)" indicates the RL policy with privileged information. \textbf{(C)} Traversal position (left) and tilted angle difference (middle) relative to the gap, and success trials distribution on initial distances to the gap $d_0$ (right), comparing ours method and Falanga’s system in an onboard-sensing setup.}
    \label{fig:8}
\end{figure}
		
{Figs. \ref{fig:8}B and \ref{fig:8}C include privileged state-based baselines and a vision-based baseline, respectively. In particular, results in Fig. 8B also include results of tracking the trajectory generated by the method \cite{mueller2015computationally} adopted by Falanga's system with the same low-level controller implementation as in Wang's. Since Falanga's trajectory generation method does not generate dynamically feasible trajectories at 80\degree \ tilt, results are shown only up to 60\degree. The experimental setup corresponding to Fig. 8B is the same as that of the \emph{Sim-to-Real Techniques} section.} 
		
{The results in Fig. \ref{fig:8}B show that the distilled policy, despite operating with indirect visual observations and no external localization, demonstrates higher success rates comparable to trajectory tracking methods with privileged information, with slightly lower success rate than the RL policy with privileged information under the specified gap size. Under specific initial conditions (i.e., test point 1 in Fig. \ref{fig:S3}), Wang's system plans distorted, infeasible trajectories difficult to follow, which we do not display in the pictures. Tracking Falanga's planner exhibits the largest attitude errors during traversal in our implementation, contributing to failures.}
		
Figure \ref{fig:8}C compares the distilled policy and Falanga's system under matched visual conditions, both using binary images in HIL experiments. To accommodate Falanga's requirement for continuous corner observation, we use 120\degree \ FoV and 512×512 resolution (versus 320×256), retraining the distilled policy for fair comparison. Falanga's system fails in most trials from longer initial distances $d_0$, with substantial position errors. As the quadrotor approaches the gap, corner points leave the FoV and replanning ceases, requiring both accurate state estimation and precise tracking at the approach-to-traverse transition as described in \cite{falanga2017aggressive} — violations of either condition accumulate into traversal errors. Tracking errors exceeding 5 cm are often observed, likely from inter-module latency and simplified dynamic models. The system's formulation that requires a hard state constraint (position, velocity, and acceleration) at the transition point sometimes struggles to solve for dynamically feasible solutions during flight, forcing the final successful replan to occur at a greater distance from the gap where state estimates are noisier. We demonstrate this replanning limitation in the \emph{Implementation and Discussion of the Baseline Systems} section in Supplementary Materials.  In contrast, the proposed method avoids manual design biases by learning from rollout outcomes and end-to-end optimization in randomized simulation, thereby alleviating explicit error propagation and model mismatch.

\begin{figure}[htpb]
    \centering 
    \includegraphics[width=1\textwidth]{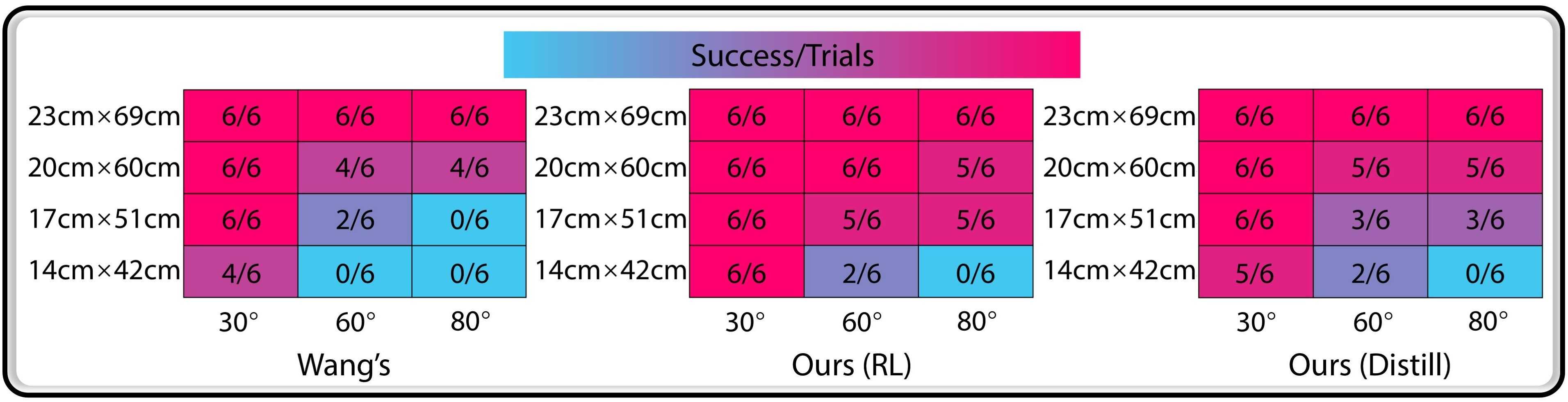}
    \caption{\textbf{Success rates of different approaches across gaps of varying size and tilt angle.} Each cell reports the number of successful traversals out of 6 trials for a given gap dimension (height $\times$ width) and tilt angle combination, corresponding to test points 1–6 defined in Fig. S3.}
    \label{fig:9}
\end{figure}

We also conduct hardware-in-the-loop experiments configuring gaps of varying dimensions to compare privileged information-based policies, vision-based policies, and Wang's system. Fig. \ref{fig:9} displays the statistical success rates for each method in different setups (detailed in the figure caption).

\subsubsection*{How tight is too tight for the SE(3) constraints?}

We can characterize the tightness of the SE(3) constraints by the size of the navigable geometric region and the magnitude of the gap's tilt angle. By comparing the performance of the RL policies against the trajectory planner baseline (Wang's system) using privileged states, we observe in Fig. \ref{fig:9} that the success rates of both systems are sensitive to constraint tightness. Under relatively loose constraints, i.e., with the 23cm$\times$69cm gap, the performance difference between the two approaches is minimal; when the gap dimensions are reduced to 14cm$\times$42cm with an 80\degree tilt, both systems fail completely. However, for the baseline system, a gap of 17cm$\times$51cm already proves excessively stringent, while the RL policy maintains significantly better performance in this regime. This indicates that the threshold for ``too tight" differs substantially between the methods.

\subsubsection*{What primarily contributes to the traversal challenge: SE(3) constraints, vision, or both?}

By comparing the middle and right panels of Fig. \ref{fig:9}, we observe that while vision-based and state-based RL policies perform similarly for the 23cm$\times$69cm gap, their success rates diverge significantly at 60\degree and 80\degree tilts in the challenging 17cm$\times$51cm setup. At the extreme 14cm$\times$42cm dimension, both policies exhibit low success rates. These results indicate a compounded effect: as geometric constraints tighten progressively, the vision-based aspect of the task acts as the primary performance bottleneck, until the geometric constraints become so stringent that both approaches fail regardless of the sensing modality. Furthermore, comparing Fig. \ref{fig:8}B and Fig. \ref{fig:8}C reveals that relying only on onboard sensing negatively impacts the traditional modular trajectory planner much more severely than it impacts the learned policies.

\section*{Discussion}
The sensorimotor policy training approach introduced in this research pushes new boundaries in achieving autonomous, precise, and aggressive maneuvers for underactuated multirotors. However, in the field of robotics, modular control architectures are widely employed in academic work \cite{kuindersma2016optimization,zhou2022swarm} and industry solutions\cite{atlas,mobileeye}, which facilitate  research and development (R\&D) across different development groups. In such a paradigm, practitioners often struggle when robots must perform autonomous precise control using only onboard sensing: engineers must meticulously tune each module to suppress accumulative errors, and these systems may lack a unified parameter set that functions consistently across the whole task space. The presented results in this work demonstrate that a controller based on direct sensorimotor mapping has the potential to eliminate the need for painstaking tuning to achieve near-perfect state estimation and accurate trajectory tracking. The enhanced deployment readiness of this method largely stems from learning from experience, end-to-end optimization, and simulation randomization, as it develops policies that extract robust task-relevant representations from observations, rather than relying on {static} module interfaces like state estimates {and introducing potentially sub-optimal human bias through manual design}. 

Extending the demonstrated capabilities to general unstructured environments remains the primary limitation and long-term goal. Achieving this vision requires first moving beyond the simplified perception structure provided by artificial landmarks in this work, which enable straightforward visual mask extraction through color thresholding or lightweight learned segmentation. In more general perception scenarios, one pathway is deploying robust, general-purpose visual foundation models \cite{kirillov2023segment} to identify and track traversable gaps from natural features, though highly dynamic flight demands high-frequency, low-latency perception that current models struggle to provide within onboard computational budgets. Another alternative is learning directly from raw sensory modalities such as depth images \cite{zhang2024back}  or LiDAR \cite{xu2025flying}, which enables direct sim-to-real transfer but introduces challenges in extracting compact, generalizable representations from high-dimensional data in real time. From a algorithmic design perspective, the current framework also offers a natural extension path towards generalized precise aggressive whole-body maneuvers: low-dimensional free-space representations (e.g., convex polyhedra \cite{deits2015computing,wang2025fast}) can serve as oracle surrogates for efficient RL training, and policy distillation's modality-agnostic nature allows subsequent integration of diverse sensor inputs.  Nevertheless, the fundamental challenge remains: breaking the \emph{precision-generalization dilemma} \cite{bauza2024simple}—where context-specific excellence and broad versatility still remain mutually exclusive—represents a grand challenge in robot learning. We regard this as a compelling direction for sustained, long-term community research.

On the other hand, sim-to-real discrepancies can also limit the precision achievable: for instance, a policy that completes a consecutive-gap track in simulation may fail under matched real-world conditions (see \emph{Failure Modes in Consecutive Gap Traversal} in Supplementary Materials). One promising approach is interactive learning on physical platforms \cite{levine2016end,pan2017agile,luo2024serl}, fine-tuning simulation-trained policies in the real world. Without a physical gap, the HIL setup used in the \emph{Results} section — generating synthetic exteroceptive observations via external localization — can mitigate collision risk during real-world rollout but still enable learning with real-world dynamics. However, this method cannot eliminate the sim-to-real gap in exteroception and remains constrained by the coverage area of external localization devices.

\section*{Method and Materials}
\subsection*{Problem Description}
The primary objective of this research is to achieve diverse goal-conditioned, precise aggressive maneuvers, including rectangular gap traversal, consecutive gap traversal, and traversal through gaps with various geometries, using only onboard sensory data. Visual landmarks \cite{falanga2017aggressive,jung2018direct} inform the system of the narrow areas to traverse. We train policies to control the quadrotor through these specified narrow passable regions in the environment based on landmark observations, without explicit position or velocity feedback. A successful traversal through the passable region is defined as achieving no interaction between the collider corresponding to the quadrotor and the gap plane outside the designated passable areas during simulation training, or no collision between the physical quadrotor and the gap structure in real-world experiments. We formalize the described control problem in the \emph{Problem Formulation} section in Supplementary Materials.

\begin{figure}[htpb]
    \centering 
    \includegraphics[width=1\textwidth]{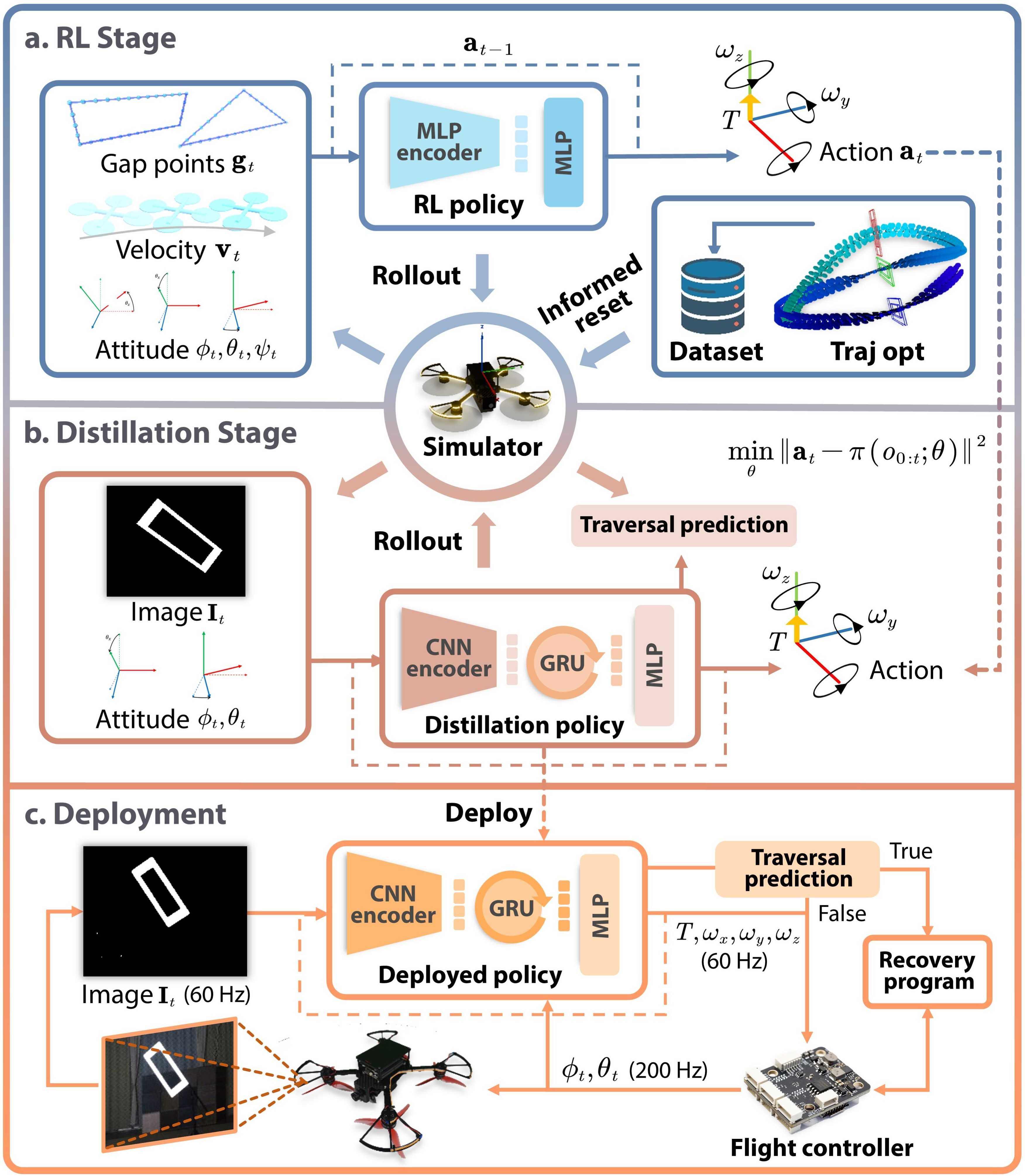}
    \caption{\textbf{Policy training framework.} We use a two-stage sim-to-real policy learning method where the agent learn policies by interacting with the simulated environment. First, an RL policy learns solutions for an oracle MDP, utilizing an informed reset strategy to guide exploration. Subsequently, the RL policy's behavior is distilled into a recurrent neural policy (denoted as SL policy in the figure) using SL. The SL policy can be directly deployed on the quadrotor without fine-tuning. Upon completion of gap traversal, as determined by a traversal prediction head, a recovery program is triggered to stabilize the quadrotor.}
    \label{fig:10}
\end{figure}

\subsection*{Methodology Overview}
We follow the general sim-to-real online policy learning paradigm, training neural network policies using simulated samples and then deploying the policies to physical quadrotors in a zero-shot manner \cite{tan2018sim,peng2018sim}.

Our problem involves several challenging characteristics that render standard model-free RL approaches inefficient: (i) high-dimensional and recurrent inputs, (ii) partial observability, and (iii) constrained solution spaces with sparse rewards \cite{ng1999policy}. To address these challenges, we adopt a teacher-student training approach \cite{chen2020learning}, also known as policy distillation \cite{rusu2015policy}.
As illustrated in Fig. \ref{fig:10}, we decouple the original RL problem—formulated as a POMDP with pixel-based recurrent observations—into two subproblems to overcome the challenges (i) and (ii): first, an MDP with a low-dimensional surrogate observation space, and second, a supervised learning (SL) problem that maps from the original observation space to optimal actions end-to-end. In particular, we design an oracle observation space and construct an MDP to approximate the original POMDP. If the optimal solution of the MDP closely approximates that of the POMDP, an SL stage that distills the MDP solution into a policy operating on the original observation space can approximately recover the optimal solution \cite{agarwal2023legged}. We address challenge (iii) in the surrogate MDP using an \emph{informed reset} strategy to improve exploration efficiency.

We use an Intel i9-14900K Central Processing Unit (CPU) for {state} transitions {between control steps} in both stages and observation generation in the RL stage, and an NVIDIA RTX 4090 Graphics Processing Unit (GPU) for neural network optimization in both stages and image generation in the distillation stage.

\subsection*{Online Reinforcement Learning from Surrogate Observations}
\subsubsection*{Observation and Action Space}
We choose $n_\mathbf{g}$ 3D points $\mathbf{g}_t$ uniformly sampled along the gap edge to surrogate the image observation $\mathbf{I}_t$, where $n_\mathbf{g}=32$ in our implementation. For consecutive gap traversal, only the gap points corresponding to the gap to be immediately traversed are input to the policy. Only if the entire collider of the quadrotor traverses that gap plane will the input points be updated to points on the subsequent gap. Roll and pitch angles $\mathbf{\phi}_t$ and $\mathbf{\theta}_t$ of the quadrotor can be read with high precision from the onboard flight controller, which are adopted in both RL and distillation stages. The oracle observation also includes privileged information, the body-frame linear velocity $\mathbf{v}_t$ that is unavailable during deployment.  We also include the previous action $\mathbf{a}_{t-1}$ as inputs. The outputs are low-level control commands $\mathbf{a}_t:=[T_t, \omega^{x}_t, \omega^{y}_t, \omega^{z}_t]$, collective thrust $T_t$ and 3-axis bodyrate $\omega^{x}_t$, $\omega^{y}_t$ and $\omega^{z}_t$, to be executed by the flight controller. We train a single RL policy initialized from the actor and critic networks previously trained on the rectangular gap with arbitrary roll angles, across all geometries in the \emph{Traversal through Narrow Passable Regions with Various Geometries} section, rather than training separate RL policies from scratch for each gap geometry. The generalized point representation of gap observations informs the policy of the passable region's geometry, supporting this joint training procedure and encouraging knowledge transfer across different task variants for a scalable RL training procedure.

\subsubsection*{Reward Signal}
RL optimizes the problem in the form of $
\max_{\pi} \mathbb{E}_{\mathbf{a}_t\sim \pi \left(  \cdot |\mathbf{o}_t \right)}\left[ \sum\nolimits_t^{}{\gamma ^tr_t} \right]$, where the immediate reward $r_t$ should be designed to find the solution of the control problem at hand. The rewards used in this work are categorized into a precision reward, the main reward component, a shaping reward, smoothness rewards, distillation regularization rewards, and a speed constraint. The formulas for each reward component are postponed to the \emph{Reward Formulas} section. The actor can obtain the precision rewards only when the collider of the quadrotor, which is conservatively modeled as a $\mathrm{34 \ cm \times 34 \ cm \times 11 \ cm}$ cuboid, is traversing the gap without collision. The shaping reward helps exploration by encouraging the quadrotor to fly towards the gap. The smoothness reward punishes action changes between control steps and constrains the magnitude of the action. These rewards make the flight maneuver smooth and natural, and facilitate sim-to-real transfer. The distillation regularization rewards encourage the quadrotor to actively detect the gap to ensure the information input to the RL policy can be mostly inferable during distillation. The speed constraint reward punishes when the speed of the quadrotor is over a constraint, set as 4m/s in this research.

\subsubsection*{Policy Representation}
The policy (actor and critic) is represented by neural networks consisting of a simple gap point encoder and a feedforward output network. The point encoder, which receives gap points as input, is a multilayer perceptron (MLP) with an intermediate global max-pooling layer to encode permutation-invariant features \cite{qi2017pointnet}. The feedforward network is an MLP that fuses the point features with other observations to output actions. The detail of the policy architecture is illustrated in Fig. \ref{fig:S4}(A).

\subsubsection*{Informed Reset}
We propose initializing the quadrotor to informative states during simulation training, rather than always resetting it to the hovering take-off state at each episode, to bias RL's exploration distribution. Using a simplified differentiable flatness model, our previous work, a quotient space-based trajectory optimization method \cite{wang2022geometrically}, can efficiently generate full-state trajectories {under $\mathrm{SE(3)}$ geometric constraints between quadrotor's tilted collider and the gap}. With carefully designed flight corridors and parameter tuning, we can apply this method for trajectory generation through narrow gaps. After collecting an offline dataset consisting of these trajectories, we {can} initialize the quadrotor by sampling states, i.e., position, velocity, and attitude, on these planned trajectories{, while we still initialize the quadrotor at hovering states with a certain probability in our implementation (see the \emph{Procedure of Informed Reset} section in the Supplementary Materials)}. The rationale behind this reset strategy is logically intuitive: high-quality trajectories concentrate the exploration distribution around crucial states, and starting episodes from these states makes random exploration actions more likely to yield high rewards. Suppose the policy learns to achieve high returns when initialized with informative states. In that case, it becomes more likely to produce complete collision-free trajectories when encountering similar states later, thereby efficiently biasing the explored distribution. This guidance approach enables the policy to learn a more robust solution rather than merely imitating trajectories obtained through local optimization with arbitrary suboptimality and simplified deterministic dynamics. {However, the efficacy of IR inherently relies on assumptions about the quality of reference trajectories. Intuitively, low-quality trajectories can reduce exploration efficiency or cause the RL agent to become trapped in shallow local optima. An interesting future direction is to develop formal verification methods for trajectory planners or reference trajectory quality, ensuring that the reduction in sample complexity for exploration remains beneficial and does not inadvertently hinder performance.}

\subsubsection*{Policy Optimization}
\label{sec:implementation}
Thanks to the surrogate gap observation that is lightweight to compute, the RL policies are trained with efficient data generation. We use proximal policy optimization (PPO) \cite{schulman2017proximal} algorithm for policy optimization that is demonstrated to be powerful with a large throughput of data \cite{makoviychuk2021isaac}. The actor and critic networks are not shared. The hyperparameter setup of the RL training is in Table \ref{tab:S3}. 

\subsection*{Observation Space Distillation via Supervised Learning}
\label{sec:distillation}
\subsubsection*{Observation and Action Space}
We use a masked gap image with a resolution of 320$\times$256 as the image observation, and roll and pitch angles directly read from the onboard flight controller as the proprioception input. The choice of a masked image is based on the fact that it is easier to transfer the policy using the simplified image from simulation to the real world than one trained on raw images, and the assumption about the scenario that the backgrounds and textures are irrelevant to the task. This choice is inspired by previous work \cite{xiong2024adaptive,liu2024visual} where a similar object-level visual mask is used to simplify the image data processing while minimizing information loss. For consecutive gap traversal, we compute only the mask of the gap that the quadrotor is about to traverse in the input image. The program detects the mask of the next gap only when the previous gap becomes undetectable, indicating that the quadrotor is closely approaching or traversing the gap plane. This design enables a simpler simulation procedure compared to including masks of all gaps in the image, avoiding issues that arise when gaps overlap in the FoV, e.g., obstruction caused by the suspension brackets.

The output action, in addition to the 4-dimensional control commands, includes an additional dimension that determines whether the entire collider has traversed the gap plane to trigger the recovery program in real-world experiments. This output head produces a value ranging from -1 to 1, where values below 0 indicate that the quadrotor's collider has not yet fully passed through the gap plane, while values above 0 indicate complete traversal. The supervision signal can be readily computed in the simulator.

\subsubsection*{Policy Representation}
As shown in Figure \ref{fig:6}, the current image input $\mathbf{I}_t$ is encoded using a lightweight convolutional neural network (CNN) module. The resulting visual features are concatenated with attitude measurements and the previous action, fed into a single-layer gated recurrent unit (GRU) to maintain a belief state from historical observations \cite{miki2022learning}. We choose an RNN instead of explicitly stacking historical frames of observations \cite{li2025reinforcement} due to the difficulty of finding a constant number of historical frames that trade off onboard inference efficiency and memory horizon in all task variants.  A following feedforward MLP module fuses the features output by the GRU together with the attitude measurements and the last action to generate the action. The detail of the policy architecture is illustrated in Fig. \ref{fig:S4}(B).

\subsubsection*{Policy Optimization}
We use the dataset aggregation (DAgger) algorithm \cite{ross2011reduction} as a meta-algorithm to train the SL policy. DAgger is an online/interactive IL method that effectively suppresses the covariate shift problem of offline learning, both in theory \cite{ross2011reduction} and practice \cite{pan2017agile}. We adopt the on-policy variant of DAgger, where we only use samples collected by the current policy for policy optimization. The hyperparameters of the DAgger training are presented in Table \ref{tab:S3}.

\subsection*{Sim-to-Real Transfer for Goal-Conditioned Precise Aggressive Flight}
For such a low fault-tolerant maneuver, effective sim-to-real techniques need to be explored.  We introduce the following sim-to-real transfer techniques based on the ideas of system identification \cite{lee2024robot} and domain randomization \cite{peng2018sim} for a successful sim-to-real transfer:
\\ \\
\textbf{Perturbation force:} A key factor in improving the success rate of the trained policy during deployment is applying perturbations to the simulated quadrotor. Such a force is applied to simulate unmodeled dynamics to extend the range of states the policy supports \cite{andrychowicz2020learning,li2025reinforcement} and to prevent the policy from overfitting specific simulator characteristics. For instance, in the distillation phase, we aim to prevent the policy from learning to make decisions that rely exclusively on historical action inputs, which potentially leads to overfitting to {simulated} dynamics \cite{nisar2019vimo,ding2021vid} while trivializing higher-fidelity visual inputs. The unobservable perturbation forces encourage the policy to rely more heavily on visual observations. Specifically, we apply a random perturbation force that persists for tens of simulation steps along each axis with a certain probability. The implementation details of the perturbation forces are provided in the \emph{Implementation of Simulation and Domain Randomizations} section.
\\ \\
\textbf{Response simulation for the flight controller:}
Accurately simulating the system dynamics of a real quadrotor, including flight controller execution and propeller force generation, is far from trivial \cite{kaufmann2023champion}, particularly for systems equipped with consumer-grade flight controllers that implement complex mechanisms to ensure stability \cite{px4autopilot}. To address this, we carefully tune the flight controller parameters and obtain a relatively consistent response delay for smooth commands. In this way, we simulate the response of the collective thrust and bodyrates in a simple way: we directly fit a set of structured parameters using real-world data to simulate the response at the moment from historical reference thrusts and body rates output by the policy. In particular, we define { the discrete-time dynamics of the actuator response as} {$\hat{a}_k^{(n)} = \frac{1}{w}\sum_{i=h^{(n)}} ^{h^{(n)}+w-1} a_{k-i}^{(n)}, \quad n = 0, 1, 2, 3$}, where $\hat{\boldsymbol{a}}_k := [\hat{a}_k^{(0)}, \hat{a}_k^{(1)}, \hat{a}_k^{(2)}, \hat{a}_k^{(3)}]$ represents the {actuator response} of the collective thrust and body rates at {discrete time step $k$, where the continuous dynamics are integrated using a 4th-order Runge-Kutta method with fixed time step following \cite{song2021flightmare}}. $\boldsymbol{a}_k := [a_k^{(0)}, a_k^{(1)}, a_k^{(2)}, a_k^{(3)}]$ denotes the command setpoint, which is assigned as the latest action output by the policy. The parameter {$\boldsymbol{h} := [h^{(0)}, h^{(1)}, h^{(2)}, h^{(3)}] \in (\mathbb{N}^+)^4$} represents the actuation delay {in discrete time steps}, and {$w \in \mathbb{N}^+$} is the {averaging window length used to model the low-pass filtering characteristics of the actuator response.} 

The collective thrust is fulfilled in the PX4 Autopilot flight controller by computing a throttle, whose value is from 0 to 1 \cite{px4autopilot}. Therefore, a precise mapping from the desired thrust and the throttle is the key for restricting the sim-to-real gap introduced by thrust execution, as we only simulate the thrust response according to the historical thrust commands output by the policy in the simulator. The method we calibrate this mapping is provided in the \emph{Low-Level Control of the Quadrotor System} section in Supplementary Materials.
\\ \\
\textbf{Response randomization:} Inspired by previous works \cite{song2023reaching,kaufmann2022benchmark} that randomize the outputs of the policy by multiplying randomization factors on them before they are input to the dynamics model, we propose to randomize the  {quadrotor} dynamics  by adding randomness in the computed flight controller response. In particular, the simulated response with randomization is $\tilde{{a}}_k^{(n)} = \hat{{a}}_k^{(n)} \cdot \boldsymbol{c}, \boldsymbol{c} \in \mathcal{U}(1-c^{(n)}, 1+c^{(n)})$, where $\mathcal{U}(\cdot, \cdot)$ represents uniform distribution. Moreover, we maintain these randomization factors $\boldsymbol{c}$ for a period of time, such as tens of decision steps, instead of refreshing them at every step. This can make it easier for the policy to encounter novel but relevant states, resulting in a larger space of states supported by the policy. The fitted parameters for response calculation are also randomized similarly. The values of the randomization factors can be found in the \emph{Implementation of Simulation and Domain Randomization} section in Supplementary Materials.  
\\ \\ 
\textbf{Perceptual latency simulation:}  We calibrate (i) the image generation time, i.e., the time from when the camera starts capturing the image to when it is transmitted to input to the policy, and (ii) the time for decision making, i.e., the time cost for neural network inference, and model these latencies in the simulator.  
\\ \\
\textbf{Mask observation randomization:}  Although we use mask images as the visual input to narrow the sim-to-real gap, appearance mismatch between the simulation and the real world can also be introduced. We apply the following two types of domain randomization to randomize the captured images: (i) randomization on the camera's intrinsic parameters, where the randomization factor is kept constant in each episode, and (ii) taking the maximum or minimum of the rendered image randomly with 50\% probability for every $\mathrm{2\times2}$ or $\mathrm{4\times4 }$ adjacent pixel squares to obtain the input image, resulting in a mask with edge noise.  We observe that using such randomization can improve the repeatability of the experiments.

\subsection*{Real-World Deployment}
Colored glowing frames are used to enable robust mask calculation via HSV (Hue, Saturation, Value) segmentation {in most of the experiments}, while we also employ segmentation model in the \emph{Validation under Learned Segmentation as Noisier Landmark Observations} section.

Real-world experiments are conducted in two distinct stages. Upon receiving a trigger signal, the onboard computer initiates the policy to control the quadrotor to traverse through the gap, marking the beginning of the first stage. The system transitions to the second stage and is controlled by a recovery program when the neural network head responsible for gap traversal detection outputs a predetermined number of positive values (set to 4 in our experiments), indicating successful passage through the gap plane. During the recovery stage, a Proportional-Derivative (PD) controller utilizes attitude feedback from the flight controller to restore the quadrotor to a stable orientation with near-zero pitch and roll angles. Subsequently, an optical flow-based control module integrated within the PX4 flight controller is employed to decelerate the quadrotor and maintain hovering.

\bibliographystyle{unsrt}
\bibliography{references}

\newpage

\section*{Acknowledgments}
We express our gratitude to Weijie Kong, Jiarui Zhang, Rui Jin, and Yuman Gao for their invaluable photography and videography services, as well as their advice on enhancing images and multimedia content. Our heartfelt thanks go to Yuhang Zhong, Mingyang Wang, and Donglai Xue for proofreading the article. We thank Yeke Chen and Guangyu Zhao for their contributions to the codebase. We thank Zhenyu Hou and Mengze Tian for their crucial suggestions on the response letter during rebuttle. We would like to convey our special thanks to Kaihan Chen, who has greatly helped with the development of the segmentation model, and Yuhan Xie, who has selflessly provided the trajectory generation codebase based on which we implement the Falanga's system.  We respectfully thank the researchers who cultivated the field of aerial robotics, especially the members from the Robotics and Perception Group, whose work pioneers in data-driven drone control, and Zhepei Wang, who developed the representative narrow gap traversal demonstration that inspires us. Without these people, this work would not have been possible. We would like to express our sincere appreciation to our lab, where we study robotics and conduct research. We appreciate the advances in LLM agents today and have used Claude Opus to refine the manuscript. Lastly, we would like to express our sincere respect to the editor and reviewers for their fair comments and impressive efforts to enhance the quality of the article.
\textbf{Funding:}
This project is funded by the National Natural Science Foundation of China No. 6232231 and National Natural Science Foundation of China under Grant No. 62203256.
\textbf{Author contributions:}
T.W. conceived the project, formulated the main idea, determined the project scope, developed training and deployment codes, conducted experiments, and wrote the manuscript. G.X. implemented the system, developed deployment code, refined training algorithms, created multimedia materials, conducted hundreds of experiments, and revised the manuscript. Z.W. designed the hardware platform, contributed to system implementation, and assisted with experimental procedures. J.L. created multimedia materials. T.C. assisted with image creation. Y.W. and Z.H. revised the manuscript. F.G.  provided key writing guidance and crucial experiment design suggestions and provided funding.       
\textbf{Competing interests:}
The authors declare that they have no competing interests.
\textbf{Data and materials availability:}
All data needed to evaluate the conclusions in the paper are present in the paper or the Supplementary Materials.  The data for this study has been deposited in the database DOI: 10.5281/zenodo.18005929.

\newpage

\begin{center}
\section*{Supplementary Materials for\\ \scititle}

% Author list for the supplement
% Indicate the corresponding authors, but do NOT include institutions here
% It would be nice if the template auto-generated this, but doing so is complicated...
	Tianyue Wu$^{\dagger}$,
	Guangtong Xu$^{\dagger}$,
	Zihan Wang,
        Junxiao Lin,
        Tianyang Chen, \and
        Yuze Wu,
        Zhichao Han, 
        Zhiyang Liu,
        Fei Gao$^{*}$ \\
\small$^\ast$Corresponding author. Email: fgaoaa@zju.edu.cn\\
\small$^\dagger$These authors contributed equally to this work.

\end{center}

% Fill out the numbers for each type of supplementary material,
% and delete any lines that aren't applicable.
% These are just example numbers that don't match the rest of this template.
\subsubsection*{This PDF file includes:}
Supplementary Sections S1 to S15\\
Figures S1 to S13\\
Tables S1 to S4\\
Algorithm S1

\subsubsection*{Other Supplementary Materials for this manuscript:}
Movies S1 to {S6}

\newpage
\section*{Supplementary Sections}
\renewcommand{\thefigure}{S\arabic{figure}}
\renewcommand{\thetable}{S\arabic{table}}
\setcounter{figure}{0}  
\setcounter{table}{0}   

\subsection*{S1  \ \ Problem Formulation}
We formulate the control problem addressed in this research in this section. The quadrotor vehicle is modeled as a single rigid body. The \emph{collider} $\mathcal{C} (\cdot)$ of this rigid body, which depends on its current state $\mathbf{x}_k$ in the special Euclidean group ($\mathrm{SE(3)}$), is used to determine whether a collision occurs between the quadrotor and its environment. The quadrotor is treated as a discrete-time dynamical system with continuous states and control inputs. The objective of this control problem is to determine the optimal control sequence that guides the quadrotor through a plane $\mathcal{P}^g$ without collision. A closed region with geometry $\mathcal{G}$ forms the only passable space on the plane, referred to as the \emph{gap} in this research. Accordingly, the free space is divided into three distinct components: the passable region $\mathcal{F}^g$ on $\mathcal{P}^g$, the initial space $\mathcal{F}^0$ where the quadrotor starts, and the target region $\mathcal{F}^1$. These spatial relations are illustrated in Fig. \ref{fig:S6}.

Since the passable region on the plane can be quite narrow, the quadrotor must adjust its attitude to leverage a momentary tilted attitude and its asymmetric body geometry to traverse the plane without collision. The goal of this research is to develop closed-loop policies to solve this control problem, which is formulated as follows:
\begin{equation*}
	\label{eq:wbc}
	\begin{aligned}
		&\exists \ T < \infty, \ \ \text{subject to:} \vspace{-0.05cm}\\
		&\quad \mathbf{x}_0 = \mathbf{x}_{\textnormal{init}}, \ \mathbf{x}_{k+1} = \mathbf{x}_k + f(\mathbf{x}_k, \mathbf{u}_k), \vspace{-0.05cm}\\
		&\quad \mathcal{C} \left( \mathbf{x}_k \right) \in \mathcal{F} :=\mathcal{F}^g\cup \mathcal{F}^0\cup \mathcal{F}^1, \forall k, \vspace{-0.05cm}\\
		&\quad \mathcal{C} \left( \mathbf{x}_0 \right) \in \mathcal{F}^0, \ \mathcal{C} \left( \mathbf{x}_T \right) \in \mathcal{F}^1, \vspace{-0.05cm}\\
		&\quad \mathbf{u}_k\sim\pi(\cdot|o_{0:k}).
	\end{aligned}
\end{equation*}
Here, $f(\cdot)$ is the system's discrete-time dynamics function, $\{\mathbf{u}_k\}$ is the control sequence along time as the decision variable, $\pi$ is the policy for decision-making conditioned on historical observations $o_{0:k}$.

\subsection*{S2  \ \ Reward Formulas}
We formulate the reward function based on the geometric relationship illustrated in Fig. \ref{fig:S6}. Specifically, we define a coordinate axis $X^g$ that is perpendicular to the plane of the gap to be traversed. For consecutive gap traversal, $X^g$ represents the coordinate axis established on the gap to be traverse next. The origin of this axis, denoted as $O^g$, is the point where the axis intersects the gap plane. The reward function after each control step is the sum of the following components.
\vspace{-0.1cm}
\begin{itemize}
	\item  \emph{Traversing reward:}
	\begin{equation*}
	    \label{eq:traver reward}
		\lambda^{\mathrm{traver}}\cdot\mathbb{I} \left[ \left| x_{k}^{g} \right|\leq l^{\mathcal{C}}
		\text{ and } \mathcal{C} \left( \mathbf{x}_k \right) \in \mathcal{F} \right] \cdot ( \textnormal{min}(l^{\mathcal{C}}, x_{k}^{g})-\textnormal{max}(-l^{\mathcal{C}}, x_{k-1}^{g})),
	\end{equation*}
    where $\mathbb{I} \left[ \cdot \right]$ is the indicator function, $x^{g}$ represents the 1-axis coordinate along the $X^g$ axis, and $l^\mathcal{C} > 0$ is a threshold determined by the size of the collider, which is set to 0.2m in our implementation. Rather than rewarding only upon full traversal, the agent can earn this reward when the quadrotor is either close to or actively traversing the gap plane. For rolled gap traversal, we set $\lambda^{\mathrm{traver}}$ as $\mathrm{10}$, while for pitched gap traversal, $\lambda^{\mathrm{traver}}=10 \cdot c^\theta$ to scale this reward, where $c^\theta=\exp (-{{\left| \theta _k-\theta ^g \right|}\big/{20\degree}})$. We add the specialized factor $c^\theta$ to encourage the quadrotor to pitch the vehicle to adapt to the tilted angle of the gap; otherwise, training tends to result in a conservative policy. We analyze the difference between pitched gap traversal and the rolled one: the quadrotor can still successfully pass through the gap with a relatively significant pitch misalignment, whereas the traversal demonstrates lower tolerance for roll misalignment. 
	\item \emph{Shaping reward:} 
	\begin{equation*}
		\lambda^{\mathrm{shaping}}\cdot\mathbb{I} \left[  x_{k}^{g} < 0 \right] \cdot \left( \left\| \mathbf{p}_{k-1}-\mathbf{p}^g \right\| -\left\| \mathbf{p}_k-\mathbf{p}^g \right\| \right),
	\end{equation*}
	where $\mathbf{p}^g$ is the geometric center of the passable region to be traversed next and $\lambda^{\mathrm{shaping}}=0.3$.
	\item \emph{Smoothness penalty:} 
	\begin{equation*}
		-\sum_{i=1}^{4}{\lambda^{\mathrm{mag}}}^{(i)}\left\| \left. \mathbf{a}_k^{(i)} \right\| \right. -\sum_{i=1}^{4}\lambda^{\mathrm{var}(i)}\left\| \mathbf{a}_k^{(i)}-\mathbf{a}_{k-1}^{(i)} \right\|, 
	\end{equation*}
    where {${\lambda^{\textnormal{mag}}}^{(i)}$ , ${\lambda^{\textnormal{var}}}^{(i)}$} greater than 0 are the penalty coefficients for the magnitude and variation of action components. These penalties are used to encourage a natural and smooth motion. {In our implementation, $[{\lambda^{\textnormal{mag}}}^{(0)}, {\lambda^{\textnormal{mag}}}^{(1)}, {\lambda^{\textnormal{mag}}}^{(2)}, {\lambda^{\textnormal{mag}}}^{(3)}] =$ $ [0.04/{(\textnormal{m/s}^2)},0.05\textnormal{/rad},0.04$ $\textnormal{/rad}, 0.02\textnormal{/rad}]$, and $[{\lambda^{\textnormal{var}}}^{(0)}, {\lambda^{\textnormal{var}}}^{(1)}, {\lambda^{\textnormal{var}}}^{(2)}, {\lambda^{\textnormal{var}}}^{(3)}] = [0.06/{(\textnormal{m/s}^2)},0.015\textnormal{/rad},0.01\textnormal{/rad},0.00\textnormal{/rad}]$.}
	\item \emph{Speed constraint:}
	\begin{equation*}
		\lambda^{\mathrm{speed}}\cdot\mathbb{I} \left[ \left\| \mathbf{v}_k \right\| \leq \mathrm{v}_{\max} \right] \cdot \left( -\exp \left( \left\| \mathbf{v}_k \right\| -\mathrm{v}_{\max} \right) +1 \right).
		\vspace{-0.2cm}
	\end{equation*}
	Here $\mathrm{v}_{\max}$ is set as 4m/s and $\lambda^{\mathrm{aggressive}}=0.05$. This reward is designed to softly constrain the velocity of the motions.
	\item \emph{Distillation regularization:} 
	\begin{align}
		\lambda^{\mathrm{distill}}\cdot\mathbb{I} \left[ x_{k}^{g} < -l^\mathrm{approach} \right] \cdot  (\langle \boldsymbol{\nu }_{k}^{b|x}, \boldsymbol{\nu }_{k}^{g} \rangle - \langle\boldsymbol{\nu }_{k-1}^{b|x}, \boldsymbol{\nu }_{k-1}^{g} \rangle),
	\end{align}
	where $\boldsymbol{\nu }^{b|x}$ represents the direction of the quadrotor's body x-axis (the directed centerline of the field of view, or FoV), and $\boldsymbol{\nu }^{g}$ denotes the direction pointing from the quadrotor's body to the center of the gap geometry. In our implementation, $\lambda^{\mathrm{distill}}$ is set as 0.15 for rolled gap traversal and 0.1 for the pitched one. The parameter $l^\mathrm{approach}$ is a manually defined threshold used to switch between this reward and the subsequent regularization reward. In our implementation, $l^\mathrm{approach}$ is set to 0.6m, though values in the range of 0.4m to 0.8m yield similar performance. This term keeps the gap within the FoV during approach, reducing the observation discrepancy between gap-point and pixel inputs at distillation time.
    \item \emph{Approaching and traversing regularization:}
    \begin{align*}
		\hspace{-0.3cm}\mathbb{I} \left[ x_{k}^{g} \geq -l^\mathrm{approach} \right] \cdot  (\lambda_v (\langle \boldsymbol{v}^{b|x}_k, \boldsymbol{n}^{g} \rangle - \langle\boldsymbol{v}^{b|x}_{k-1}, \boldsymbol{n}^{g}\rangle)-\lambda_\theta|\theta_k-\theta^g|-\lambda_\psi|\psi_k-\psi^g|),
    \end{align*}
    where $\boldsymbol{v}^{b|x}_k$ represents the quadrotor's velocity in the body frame, and $\boldsymbol{n}^{g}$ is the normalized vector indicating the positive direction of $X^g$.  This term regularizes approach and traversal behavior: the quadrotor is expected to traverse the gap with a velocity perpendicular to the gap plane while maintaining its nose oriented toward the plane. Without this specification, the quadrotor’s behavior can result in a larger discrepancy between the success rates observed in real-world experiments and simulations.  In our implementation, for rolled gaps, $\lambda_v$ is set as $\mathrm{0.375}$, and  $\lambda_\theta$ and $\lambda_\phi$ are set as $\mathrm{0.25}$ and $\mathrm{0.5}$, respectively, where the unit of the angle in this formula is $\mathrm{rad}$. For pitched gap traversal, these weights are 0.8 times the above values.

\end{itemize}

\subsection*{S3 \ \ Quadrotor Dynamics Model}
The quadrotor is considered a rigid body that is actuated by four propellers, which generate four parallel lift forces $f_1$, $f_2$, $f_3$, and $f_4$. The state of the quadrotor 
is described by $\boldsymbol{s} = [\boldsymbol{p}, \boldsymbol{q}, \boldsymbol{v}, \boldsymbol{\omega}]$, and its simplified kinematic equations  are expressed as:
\begin{align}
\dot{\boldsymbol{p}} &= \boldsymbol{v}, \ \ \dot{\boldsymbol{v}} = {\mathbf{R}(\boldsymbol{q})}(\boldsymbol{z}_{B}- \boldsymbol{f}_{\mathrm{drag}}) + \boldsymbol{g},\nonumber\\
\dot{\boldsymbol{q}} &=\frac{1}{2}[\boldsymbol{\omega}]_\times \cdot \boldsymbol{q}, \ \ \dot{\boldsymbol{\omega}} = \mathbf{J}^{-1}(\boldsymbol{\tau}-\boldsymbol{\omega}\times\mathbf{J}\boldsymbol{\omega}). \label{eq:dynamics}
\end{align}
Here, $\boldsymbol{p}$, $\boldsymbol{v}$, and $\boldsymbol{q} = [q_w, q_x, q_y, q_z]$ denote the position, linear velocity, and unit quaternion of the quadrotor expressed in the world frame, respectively. { $\mathbf{R}(\boldsymbol{q})$ represents the rotation matrix computed from quaternion $\boldsymbol{q}$.} $\boldsymbol{\omega} = [\omega_x, \omega_y, \omega_z]$ is the body rate of the quadrotor. The notation $[\cdot]_\times$ indicates the skew-symmetric matrix form of a vector. $\boldsymbol{g} = [0,0,-9.81]^{\rm T}$ is the constant gravitational acceleration, and $\mathbf{J}$ is the inertia matrix. $\boldsymbol{z}_B=[0,0,T]^{\rm T}$ represents the mass-normalized thrust in the quadrotor's body frame, and $\boldsymbol{\tau}=[\tau_x, \tau_y, \tau_z]$ is the body torque. Once the mass-normalized lift forces $[f_1, f_2, f_3, f_4]$ are determined, the collective thrust $T$ and body torque $\boldsymbol{\tau}$ can be directly calculated using the Newton-Euler equations. In our simulation, the drag force $\boldsymbol{f}_{\mathrm{drag}}$ is modeled as {$ \boldsymbol{f}_{\mathrm{drag}} = \boldsymbol{d} \odot \boldsymbol{v}_{\text{b}} + \boldsymbol{k} \odot |\boldsymbol{v}_{\text{b}}| \odot \boldsymbol{v}_{\text{b}}$,    where $\boldsymbol{d}=[d_x, d_y, d_z]^T$ and $\boldsymbol{k}=[k_x, k_y, k_z]^T$ are the linear and quadratic drag coefficient vectors, respectively,} $\boldsymbol{v}_{\text{b}}$ is the linear velocity in the quadrotor's body frame, { $|\boldsymbol{v}_{\text{b}}| = [|v_{bx}|, |v_{by}|, |v_{bz}|]^T$ denotes the element-wise absolute value, and $\odot$ represents the Hadamard (element-wise) product.}

\subsection*{S4 \ \ Low-Level Control of the Quadrotor System}
\subsubsection*{Collective thrust and bodyrates control interface}
In the presented system, the control commands the policy output are the collective thrust and bodyrates, a widely used interface that is supported by the common flight controllers. This control interface is demonstrated to not only exploit the maneuverability of the system without position or velocity feedback, but also facilitates the sim-to-real transfer compared to direct control of the single-motor thrust \cite{kaufmann2022benchmark}. In particular, in the presented system, the onboard computer transmits the desired thrust $T$ to the PX4 flight controller as a throttle $\hat{T} \in [0,\ 1]$. A precise thrust mapping relationship is then devised for each aircraft, modified from the official PX4 Autopilot documentation \cite{px4autopilot}, which is presented in the following.
\begin{equation*}
    T = \lambda_1 V^{\lambda_2} \left( \lambda_3 \hat{T}^2 + (1 - \lambda_3) \hat{T} \right),
\end{equation*}
where $V$ is the current battery voltage and $\lambda_i$ for $i = 1, 2, 3$ is the system parameters to be calibrated. Accurate calibration of $\lambda_i$ is crucial, as the policy outputs mass-normalized thrust in simulation but requires throttle conversion for real-world execution.

We implement a thrust model calibration technique as follows. Incremental payloads are attached to the drone, and the voltage and throttle under hover conditions controlled by a Proportional-Integral-Derivative (PID) controller are recorded continuously, from a fully charged state to battery depletion. At a hovering state, the drone's weight is considered the effective thrust. Then, we employ MATLAB’s curve fitting toolbox\footnote{https://www.mathworks.com/products/curvefitting.html} to fit the parameters. The specific parameters and the fitting results are shown in Fig. \ref{fig:S7}.

\subsubsection*{Optical flow-based control in recovery mode}
We use Position Mode of the PX4 Autopilot flight controller \cite{px4autopilot} to achieve attitude recovery and hovering. The optical flow module, including a downward-facing camera and Time of Flight (ToF) distance sensor, is required by the flight controller firmware\footnote{https://docs.px4.io/main/zh/sensor/optical\_flow.html}.

\subsection*{S5 \ \  Quadrotor Platform Setup}

The quadrotor features a 250 mm wheelbase diameter and has a total mass of 820 g. It is powered by four F60PROV brushless motors rated at 2550 KV\footnote{kv2550. https://store.tmotor.com/product/f60prov-fpv-motor.html}, each paired with 5-inch propellers to achieve a thrust-to-weight ratio of 5.81. The aircraft incorporates an NxtPX4v2 flight controller\footnote{https://micoair.cn/docs/nxtpx4} integrated with a four-in-one electronic speed controller\footnote{http://www.hikrc.com/pd.jsp?fromColId=103\&id=62\#\_pp=103\_435} rated for up to 60A continuous current. For perception, the system employs a Hikrobotics MV-CB013-A0UM-S monocular camera\footnote{https://www.hikrobotics.com/cn/machinevision/productdetail/?id=9706} \footnote{https://www.hikrobotics.com/en/machinevision/productdetail/?id=5908} and an Upixels optical flow sensor with integrated time-of-flight (ToF) ranging capability\footnote{http://www.upixels.com/hnyx2019/vip\_doc/15917119.html}. We select NVIDIA Orin NX \cite{NX}, a computer tailored for embedded and edge systems, as the onboard computer to implement the control policy.

\subsection*{S7 \ \ Procedure of Informed Reset}
To reset the quadrotor to informative states at the start of an episode, we first construct an offline dataset containing states generated through trajectory optimization \cite{wang2022geometrically}. Specifically, this dataset is created by uniformly sampling the gap's pose and the quadrotor's initial position within the defined task space, ensuring comprehensive coverage of the entire task space. We observe that the employed trajectory optimization method is highly effective in planning feasible and smooth trajectories in most cases for single-gap traversal, using a conservatively designed SFC visualized in Fig. \ref{fig:S8}. More specifically, the SFC consists of $2n+1$ polyhedrons that constrain the $2n+1$ segments of the resulting polynomial trajectory, where $n$ is the number of gaps to be traversed. All trajectories are generated with zero initial velocity, as we observe that non-zero velocities can cause trajectory distortion and infeasibility due to the local nature of the optimizer and the optimization formulation of the $\mathrm{SE(3)}$ planner \cite{wang2022geometrically}.

After constructing the informative state dataset, we introduce a probability $p = 0.5$ for the training program to randomly sample a state as the quadrotor's initial state, including its position, velocity, and attitude, along a trajectory in the dataset, with the corresponding gap pose. We formalize this improved RL training procedure in Algorithm \ref{alg:ir}.

\subsection*{S8 \ \ Segmentation Model Implementation}
\subsubsection*{{Network architecture}}
Our network adopts an encoder-decoder architecture for real-time gap segmentation on resource-constrained devices, as illustrated in Fig. \ref{fig:S2}. It comprises a MobileNetV3-Small \cite{howard2019searching} backbone (in particular, \texttt{mobilenet\_v3\_small} implemented in PyTorch\footnote{{https://pytorch.org/vision/main/models/mobilenetv3.html}}) , a simplified ASPP module \cite{chen2018encoder} with five parallel branches, and a decoder that fuses high-level ASPP features with low-level backbone features.

% \subsubsection*{Dataset creation}
% We collect 8000 real-world images using the same camera equipped on the quadrotor. The images are collected in different scenarios with objects randomly placed.

% The mask labels are annotated using three methods: prompting the released Segment Anything Model (SAM) \cite{kirillov2023segment} (in particular, the \texttt{vit\_h} model \cite{segment-anything}), manually identifying 8 vertices to define a complete gap frame within FoV, manually identifying edge lines to define an incomplete gap frame within FoV. We observe that while SAM achieves near-perfect segmentation at close range (e.g., 0-3m), its performance remains suboptimal at greater distances or with even slight motion blur. Even repeated prompts fail to yield satisfactory segmentation results, necessitating the use of the latter two annotation methods in some scenarios.

\subsubsection*{{Training}}
{The model is supervised with a multi-component loss function combining Cross-Entropy (CE) loss, Morphological loss, and Dice loss with weights of 1.0, 0.1, and 0.3, respectively. The Morphological loss computes Mean Squared Error (MSE) between morphologically-transformed predictions and targets, where erosion is implemented as \texttt{-max\_pool2d(-x)} and dilation as \texttt{max\_pool2d(x)} using a 5×5 kernel, then combined Morphological loss as $L_{\textnormal{morph}} = 0.5 \cdot L_{\textnormal{erosion}} + 0.5 \cdot L_{\textnormal{dilation}}$. The Dice loss optimizes the Dice coefficient with a smoothing factor of 1e-5. The model is optimized using the Adam optimizer with a fixed learning rate of 0.0005 over 400 epochs. Training employs a batch size of 128.}

\subsection*{{S9 \ \ Implementation and Discussion of the  Baseline Systems}}
\subsubsection*{{Implementation}}
{
The trajectory optimization method in Baseline 1 \cite{wang2022geometrically} is previously described in the \emph{Procedure of Informed Reset} section, which is implemented using the open-source code\footnote{{https://github.com/ZJU-FAST-Lab/GCOPTER}}. The maximum speed constraint and bodyrate constraint are aligned with the demonstrated policies in the \emph{Result} section. The tracking controller is implemented as in \cite{wang2021geometrically}, which is a carefully tuned PID controller.  }

{
The \emph{approach trajectory} (see Section II-C of \cite{falanga2017aggressive}) generation method in Baseline 2 is implemented with the open-source code\footnote{{https://github.com/markwmuller/RapidQuadrocopterTrajectories}}, the official implementation of the trajectory planner employed \cite{mueller2015computationally}. This planner requires the specification of an initial state, a terminal state, and the trajectory execution time. Dynamic constraints, including maximum and minimum thrust as well as bodyrate limits, cannot be imposed during trajectory optimization. Therefore, the approach to finding feasible trajectories involves sampling execution times and verifying trajectory feasibility. In our implementation, we design a sampling method as follows: given an initial execution time guess $t_\textnormal{approach}$, the planner samples between $\alpha_\textnormal{min}$ and $\alpha_\textnormal{max}$ times of $t_\textnormal{approach}$ at a resolution of $r$. The initial value of $t_\textnormal{approach}$, i.e., the execution time of the first planned trajectory, is manually tuned. Subsequently, the planner sets $t_\textnormal{approach}$’s initial value to the remaining execution time of the currently executed trajectory at the current timestamp when replanning is triggered. If no feasible trajectory is found within this time window, the controller continues to execute the last trajectory. Replanning is attempted immediately after each visual feedback update. In the planner, the feasible net thrust range is set from 0.41 (i.e., 4 m/s²) to 2.04 times the acceleration due to gravity (i.e., 20 m/s²), while bodyrate is set at 8 rad/s. This limit is chosen because we observe that 6 rad/s significantly constrained the system's replanning capability. Throughout all experiments, $\alpha_\textnormal{min}$ and $\alpha_\textnormal{max}$ are set to 50\% and 200\%, respectively, and $r$ is 5\%. The \emph{traverse trajectories} are generated by solving the problem (6) in \cite{falanga2017aggressive} using the NLopt library\footnote{{http://ab-initio.mit.edu/nlopt}}. The parameters $\textnormal{v}_{\textnormal{0,max}}$ and $d_{\textnormal{min}}$ are set as 4 m/s and 0.3 m, respectively.  The trajectories are tracked using the same PID controller as in Baseline 1.}

{
Both state estimation approaches, motion capture-based and vision-based, operate within an absolute global coordinate frame. For the motion capture system, this frame is established by the motion capture infrastructure; for vision-based estimation, it is defined relative to the static gap landmark. Each system provides 6D pose estimates (by motion capture system or the PnP algorithm implemented in OpenCV\footnote{{https://opencv.org/}}) that are fused with onboard IMU data using Extended Kalman Filtering (EKF) similar to \cite{lynen2013robust}. The fusion produces estimates of position, velocity, and yaw angle. Roll and pitch estimates, however, are obtained directly from the flight controller's onboard IMU processing without fusion. These direct roll and pitch estimates are treated as sufficiently accurate, as experimental observations show they differ from motion capture-fused attitude measurements by less than 1\degree \ on average, indicating these attitude components are not the dominant error source in the system.}

\subsubsection*{{A discussion on the trajectory generation methods}}
{Both planners face fundamental trade-offs that constrain their online planning capabilities. Wang et al.'s planner \cite{wang2022geometrically} formulates the problem under full $\mathrm{SE(3)}$ constraints but requires numerical optimization on a nonconvex optimization whose quality depends on manually designed corridors and the optimization's initial guess. As a result, solution quality is severely influenced by the corridor design and can easily fail with a non-zero initial velocity and acceleration. The trajectory planning method employed in Falanga's system \cite{mueller2015computationally} achieves efficient computation by specifying a termination state without explicitly imposing collision-avoidance and dynamical feasibility constraints, while such a formulation inherently fails to guarantee the solution's feasibility.}

{To investigate the planning limitations of \cite{falanga2017aggressive}, we simulate the planner under varying state estimation noise levels, motivated by the experimental findings in the \emph{Performance Evaluation with Reference Baseline Systems} section. Following \cite{falanga2017aggressive}, state estimation uncertainty is modeled to increase quadratically with distance from the gap. Simulation results (Fig. \ref{fig:S9}) reveal that even when considering only position and yaw estimation inaccuracy, the planned trajectories themselves can result in collisions with the gap frame without trajectory tracking errors. The collision probability increases with longer initial distances from the gap, larger gap orientation angles, and higher noise levels. This failure mode arises because the planning formulation cannot generate dynamically feasible corrective trajectories when attempting to leverage the improved state estimates available near the gap where visual accuracy is highest. Consequently, the system must rely on the last successfully computed trajectory, which is planned at a greater distance where state estimates contain more uncertainty and may be not a collision-free trajectory.

\subsection*{S10 \ \ Ablation Study in Position-Only Constrained Setup}
In this section, we trained privileged-information-based policies with only positional constraints in a consecutive traversal scenario, mimicking standard drone racing setups. Specifically, we set the centers of the positional constraint ranges to coincide with the centers of the narrow gaps in Tracks 4 and 5 (detailed in the Table S1). The drone's center is constrained within a 5cm radius ring when passing through the gate plane, corresponding to the tolerance along the quadrotor's height during narrow gap traversal. We evaluated this position-only constrained policy both in its native training setting and in the consecutive narrow gap scenario and also the gap traversal policies in these position-only constrained track.
	
Results (Fig. S10) demonstrate that position-only constrained policies fail to traverse most narrow gaps, with the sole exception of the first gap on Track 5, and are unable to complete a gap track. This single highly successful gap traversal (as noted in Fig. S10(B)) occurs largely by coincidence: despite the policy's unawareness of the full SE(3) constraints, the drone's traversal pose happens to closely match the gap's tilted orientation, with attitude errors often within 10°. However, when attempting the second or third narrow gaps on Track 5, attitude errors frequently reach 20°, resulting in failure. Conversely, policies trained strictly under narrow-gap geometric constraints exhibit considerable success rates when evaluated on tracks requiring only positional constraints. Failures in this reverse evaluation primarily occur because the SE(3)-constrained policies learn a wider displacement tolerance along the gap's long edge, causing some otherwise successful trajectories to violate the strict 5 cm positional constraint at the center.

\subsection*{S11 \ \ Implementation of Simulation and Domain Randomization}
\subsubsection*{Dynamics and observation simulation method}
We develop our dynamics simulation for training based on the open-source Flightmare simulator \cite{song2021flightmare}, utilizing CPU multiprocessing to compute transition function during both the RL and distillation stages. Gap point observations and other state information required in the RL stage are generated using the CPU, while the binary image observations needed for the distillation stage are computed on the GPU, implemented with the NVIDIA Warp library\footnote{https://github.com/nvidia/warp}. For non-curved gaps, such as rectangular or triangular gaps, we triangulate the gap instance and perform ray tracing with the Warp library. If the ray distance for a pixel is less than 10 meters, the pixel value is set to 1, indicating the pixel is on the gap mask; otherwise, the pixel value is set to 0. For curved gaps, such as elliptical or arch-shaped gaps, we create a large rectangular gap plane and perform ray tracing. A pixel value is set to 1 only if the intersection of the ray corresponding to the pixel and the plane satisfies the geometric equation of the gap. 

\subsubsection*{Implementation and parameters of domain randomization}
\begin{itemize}
\item  \emph{Perturbation force:} At each control step, the program applies a random perturbation force with probability $p_f$ if the following conditions are met: (i) no current perturbation force is being applied to the quadrotor,
(ii) the quadrotor's distance from the gap plane exceeds 1.5m, and (iii)
the angular velocity of the quadrotor is below 3 rad/s. If a perturbation force is applied, it lasts 20 control steps (i.e., 1/3 of a second).  The perturbation force is generated by uniformly sampling a three-dimensional acceleration vector in the global frame $\mathbf{a}_f$ (in $\mathrm{m/s^2}$) within a range of $(0, a_{\mathrm{max}}^x]$, $(0, a_{\mathrm{max}}^y]$, and $(0, a_{\mathrm{max}}^z]$ for the $x$, $y$, and $z$ dimensions, respectively. In our implementation, this sampling method ensures that the perturbation effectively challenges the quadrotor’s control response without violating system constraints. In our implementation, for single gap traversal, $p_f=\mathrm{0.1}$, $a_{\mathrm{max}}^x=\mathrm{2m/s^2}$, $a_{\mathrm{max}}^y=\mathrm{1m/s^2}$, and $a_{\mathrm{max}}^z=\mathrm{1m/s^2}$, for RL training, while $a_{\mathrm{max}}^x=\mathrm{1.5m/s^2}$ for the distillation stage. For consecutive gap traversal, in the distillation stage, $p_f=\mathrm{0.05}$, $a_{\mathrm{max}}^x=\mathrm{1m/s^2}$, $a_{\mathrm{max}}^y=\mathrm{0.5m/s^2}$, and $a_{\mathrm{max}}^z=\mathrm{0.5m/s^2}$, while the parameter setup of the RL stage is consistent with the one of single gap traversal.  

\item \emph{Response randomization:} Once the current response is computed according to the method in the \emph{Response simulation for the flight controller} section, it is then multiplied with a random factor $\boldsymbol{c}:=[c^{(0)}, c^{(1)}, c^{(2)}, c^{(3)}]$. The same $\boldsymbol{c}$ lasts for $n_c$ {times of policy outputs}, where $n_c$ is uniformly sampled from a range of [30, 90] in our implementation, until it is updated as a new value. The parameter representing response latency, i.e., {$h^{(0)}$ donating thrust latency, $h^{(1)} \sim h^{(3)}$ donating bodyrate latency}, are randomized at 40\%, 30\%, respectively.
\item \emph{Air drag:} We largely randomize the air drag coefficients at 50\%.
\end{itemize}

\subsection*{{S12 \ \ Failure Modes in Dynamic Gap Traversal}}
{The \emph{Reactive Dynamic Gap Traversal} demonstrates that policies merely trained on static gaps can traverse moving gaps through reactive control, with performance significantly enhanced by domain randomization during training. However, since the policy cannot predict future gap motion and is not explicitly trained on dynamic scenarios, its behavior is purely reactive—responding to instantaneous observations without anticipating motion patterns. This reactive nature reveals its fundamental limitations} and creates exploitable failure modes.
	
{A critical failure case occurs when the gap's motion pattern changes abruptly as the quadrotor approaches. While the policy can react to the updated gap position based on relative visual observations, the tight $\mathrm{SE(3)}$ constraints may prevent finding a feasible traversal solution when the change occurs too close to the gap. To demonstrate this, we design experiments where the gap moves laterally at constant velocity $\|v_{\text{gap}}\| = 1$ m/s along the y-axis until the quadrotor-gap distance decreases to threshold $d_\text{min}$, at which point the gap halts abruptly: $v_{\text{gap}} = \mathbb{I}[d > d_\text{min}] \cdot [0, 1, 0]$ m/s, where $\mathbb{I}[\cdot]$ is the indicator function and $d$ is the distance between the gap center and the quadrotor. The gap maintains constant height throughout. Initial conditions match those in Fig. 3C. and Fig. \ref{fig:S10} visualizes trajectories for different $d_\text{min}$.}
	
{Results show that collision depends critically on when the motion change occurs. For large $d_\text{min}$ (e.g., 2 m, middle panel in Fig. \ref{fig:S10}), the quadrotor has sufficient reaction time and distance to adapt its maneuvers, successfully traversing the now-stationary gap. For very small $d_\text{min}$ (e.g., 0.5 m, rightmost panel), the motion change occurs so late that the quadrotor's state remains well-aligned with the gap despite the halt, also enabling successful traversal. However, intermediate $d_\text{min}$ values (e.g., 1 m, leftmost panel) create a critical failure zone: the motion change occurs close enough that geometric constraints prevent corrective maneuvering, yet far enough that the quadrotor's position alignment is significant. This mismatch leads to collision with the gap plane.}
	
{A second fundamental failure mode arises from the quadrotor's limited directional control authority during traversal. When passing through narrow gaps at tilted orientations, the vehicle's attitude constraints restrict acceleration to specific directions. Gaps moving fast inevitably cause positional constraint violations and collisions. As shown in Fig. 3C, successful traversal requires gap speeds below certain values for different tilted orientations of the gap. These speed limits represent fundamental boundaries of the reactive approach and cannot be overcome without explicit training or planning on specified high-speed dynamic scenarios.}

\subsection*{{S13 \ \ Failure Modes in Consecutive Gap Traversal}}
\subsubsection*{{Training Failures in Exploration From Scratch}}
{As expected, when the feasible solution space become narrow in consecutive gap tracks, e.g., with narrower inter-gap spacing or larger orientation differences between adjacent gaps, RL training can fail to discover feasible policies even with informed reset (IR), as demonstrated in Fig. \ref{fig:S11}.} {This failure mode exposes a fundamental tension in RL for the precise aggressive flight problem at hand: as task difficulty increases, two effects compound—(i) the planner itself struggles, providing lower-quality reference trajectories with more frequent collisions, and (ii) the actual feasible solution space shrinks—that make exploration even harder.}
	
{A pragmatic workaround is curriculum-based initialization: training a policy on a slightly easier track configuration that shares structural similarity, then use its weights to initialize training on the target track. Fig. \ref{fig:S11} shows a successful example where this bootstrapping enables fast convergence. We postpone the description of experiment setup in the caption of Fig. \ref{fig:S11}. However, this approach lacks principled scalability because finding appropriate initialization policies requires manual trial-and-error, and the structural similarity requirements are poorly understood. Ultimately, scaling up training, developing principled curricula, or enhancing exploration capabilities of the training algorithm will be necessary.
	
\subsubsection*{{Deployment Failures due to Sim-to-Real Performance Gap}}
{Even when policy training succeeds, policies may complete tracks in simulation yet fail in physical deployment. This failure mode becomes  probable on tracks with refined behavior, such as Track 5 in the \emph{Consecutive Narrow Gap Traversal} section, which demands coordinated lateral corrections between gaps and attitude changes approaching 70\degree. Fig. \ref{fig:S11} visualizes representative failure trials in HIL experiments.}
	
{One of the reasons for the increased deployment fragility on consecutive tracks compared to single gap traversal can by temporal trajectory shift accumulation coupled with diminishing recovery margins.  For instance, the quadrotor's state after the first gap becomes an increasingly poor match for the simulated state distribution the policy is trained on, making subsequent gaps progressively more challenging. However, as evident from the trajectories, the simulation and real quadrotor states shows no significant differences after traversing the first gap, yet failure occurs at the second gap or third gap. Therefore, we hypothesize another two contributing factors. First, the policy's internal representation of the situation—maintained through recurrent processing of the observation history—may diverge between simulation and deployment even when instantaneous states appear similar due to distribution shift. Second, consecutive gap traversal inherently provides less room for error than single gap scenarios. When approaching a single gap, the policy can recover from small mistakes by adjusting the approach trajectory. However, in consecutive scenarios, the constrained exit states from the previous passed gap can narrow the feasible solution space for traversing the next gap. Under these circumstances, the impact of policies' suboptimality on success is greatly magnified.} 

\subsection*{{S14 \ \ Performance under Different Speed Constraints}}
{The main results in this work are obtained with a soft speed constraint of 4 m/s (see the \emph{Reward Formulas} section). To understand the impact of this  choice, we evaluate policies trained under different speed limits: 3 m/s, 4 m/s, 5 m/s, and unconstrained (where peak speeds reach approximately 6.5 m/s). Both simulation and HIL results are presented in Fig. \ref{fig:S12}.}
	
{The results reveal several notable patterns. The RL policies with privileged state information maintain consistently high success rates as speed limits are relaxed, indicating that precise maneuvering is achievable at higher speeds when accurate state feedback is available.} {However, the performance gap between vision-based distilled policies and RL policies widens progressively as speed increases. This degradation likely stems from two interrelated factors: First, the policy must infer motion information such as velocity from temporal visual sequences, but faster motion reduces the information can be captured per unit flight distance with a fixed observation frequency of 60 Hz. Second, higher flight speeds reduce the temporal margins available for correcting trajectory deviations, thereby amplifying the impact of both perception uncertainty and imitation errors from the distillation process.}
	
{As expected, real-world performance also degrades as speed increases. Beyond the policies' performance degradation in simulation, higher speeds amplify the effects of inaccurate system and sensor dynamics in the simulator under a low control tolerance. While domain randomization mitigates some of these effects, residual distribution mismatches become more consequential at elevated speeds. The chosen 4 m/s constraint thus represents a practical balance between demonstrating aggressive maneuvering capabilities and maintaining robust real-world deployment, primarily driven by vision system limitations and sim-to-real transfer considerations rather than fundamental control constraints.}

\subsection*{{S15 \ \ Policies Unification}}
{In the \emph{Results} section, we train separate policies for different task variants such as single gap traversal and consecutive gap traversal. In this section, we attempt to unify multiple policies. In the \emph{Traversal through a Rectangular Narrow Gap with Low Clearances} section, we develop distinct RL policies for traversing rolled gaps and pitched gaps, as directly merging these two task spaces in RL training is challenging. Here we employ the well-known multi-expert distillation method \cite{wan2023unidexgrasp++} to distill the two RL policies and obtain a unified, deployable vision-based policy.} 

{During distillation, the student policy learns to mimic the appropriate expert based on the gap orientation: when encountering a rolled gap, it imitates the rolled-gap expert; for pitched gaps, it follows the pitched-gap expert. This approach can be extended to any number of task subsets by training corresponding expert policies.}

{Simulation results are presented in Table \ref{tab:S4}.  The unified policy achieves performance comparable to the separate specialist policies in most cases. However, at a pitch of 20$\degree$, we observe a relatively significant drop in the success rate of the unified policy. We hypothesize this degradation likely occurs because moderately pitched gaps represent boundary regions between the two expert domains. At these boundaries, visual observations may sometimes be ambiguous, making it difficult for the unified policy to reliably determine which expert behavior to follow.}

\newpage

\vspace*{\fill}
\begin{figure}[H]
    \centering  % 添加这行
    \vspace{-0.3cm}
    \includegraphics[width=1\textwidth]{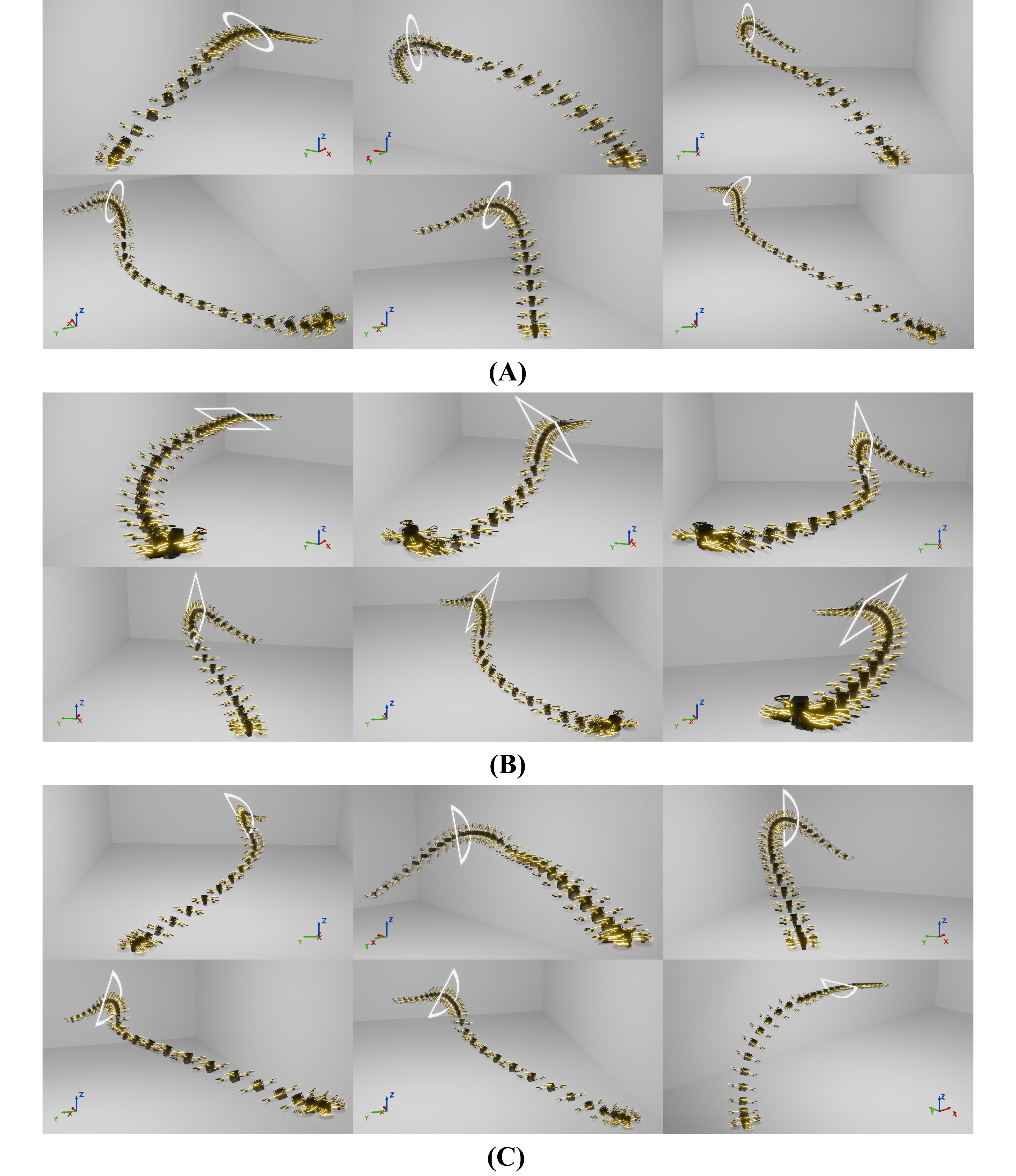}
    \vspace{-0.3cm}
    \caption{\textbf{Simulation trajectories of traversal through gaps with various geometries.} Results are shown for gaps with ellipse shape \textbf{(A)}, diamond shape \textbf{(B)}, and arch shape \textbf{(C)}.}
    \label{fig:S1}
\end{figure}
\vspace*{\fill}

\begin{figure}[H]
    \includegraphics[width=1\textwidth]{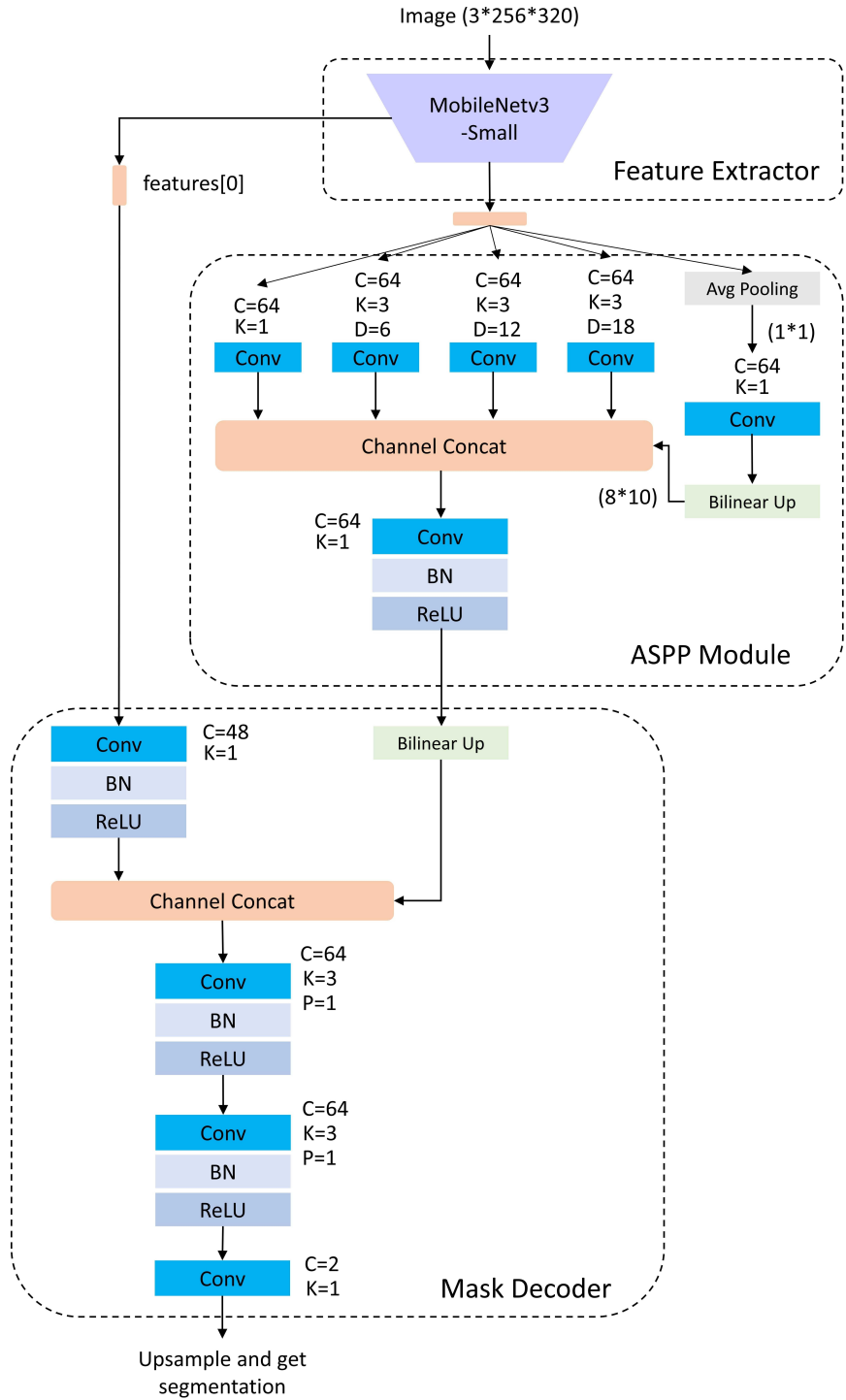}
    \vspace{-0.5cm}
    \caption{\textbf{The detailed neural network architecture of the lightweight segmentation model.}  ``C", ``K", ``S", and ``P" are the input channel, kernel size, stride length, the padding, respectively,  of the (2D) CNN. “BN” is batch normalization. ``Avg Pooling" is global average pooling. ``Bilinear Up" is bilinear upsampling. ``feature[0]" represents the first feature in the feature stack (in a form of the \texttt{Sequential} data structure of PyTorch) output by the MobileNetv3 encoder.}
    \label{fig:S2}
\end{figure}

\newpage

\vspace*{\fill}
\begin{figure}[H]
    \centering
    \includegraphics[width=0.9\textwidth]{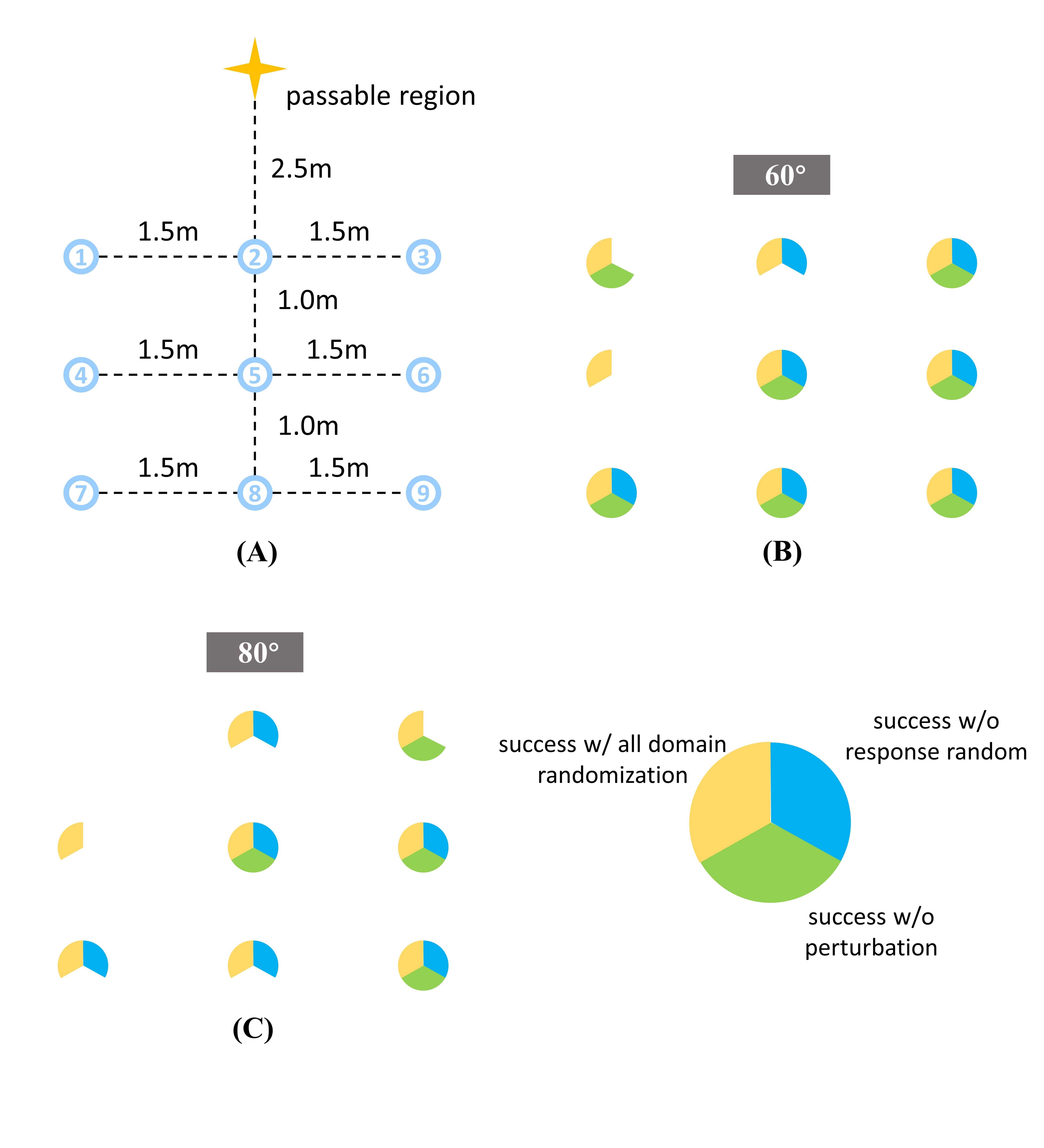}
    \caption{\textbf{Test points and selected results of the ablation study in the \emph{Key Ingredients for Policy Learning and Sim-to-Real Transfer} section.} \textbf{(A)} The test point layout for a pose of gap. \textbf{(B)} Results of the ablation on a gap with a 60\degree orientation. \textbf{(C)} Results of the ablation on a gap with a 60\degree \  orientation. Sectors of different colors represent whether different configurations can successfully traverse the narrow gap at the corresponding test point; if successful, the color sector is present at that point, and vice versa if unsuccessful.}
    \label{fig:S3}
\end{figure}
\vspace*{\fill}

\newpage

\vspace*{\fill}
\begin{figure}[H]
    \includegraphics[width=1\textwidth]{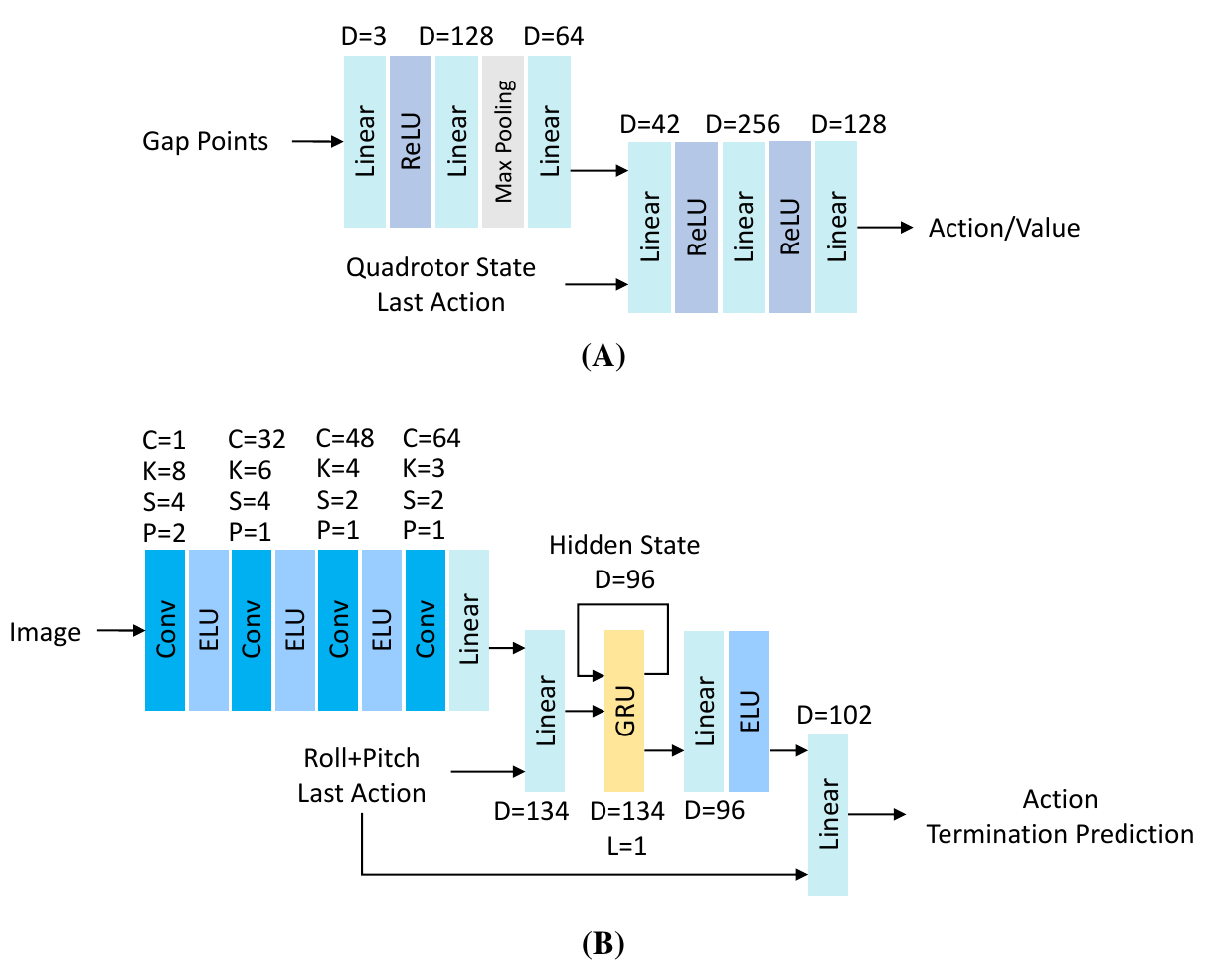}
    \caption{\textbf{Detailed neural network architectures in for the policies.} \textbf{(A)} The neural network architecture of the actor and critic used in RL training. ``D" is the input dimension of the corresponding linear layer. \textbf{(B)} The neural network architecture of the deployed policy in distillation. ``C", ``K", ``S", and ``P" are the input channel, kernel size, stride length, the padding, respectively,  of the (2D) CNN. D and L are the input dimension and layer number of the GRU, respectively. ``D" labeled after the hidden state is its dimension.}
    \label{fig:S4}
\end{figure}
\vspace*{\fill}

\newpage

\vspace*{\fill}
\begin{figure}[H]
    \centering  % 添加这行
    \includegraphics[width=1\textwidth]{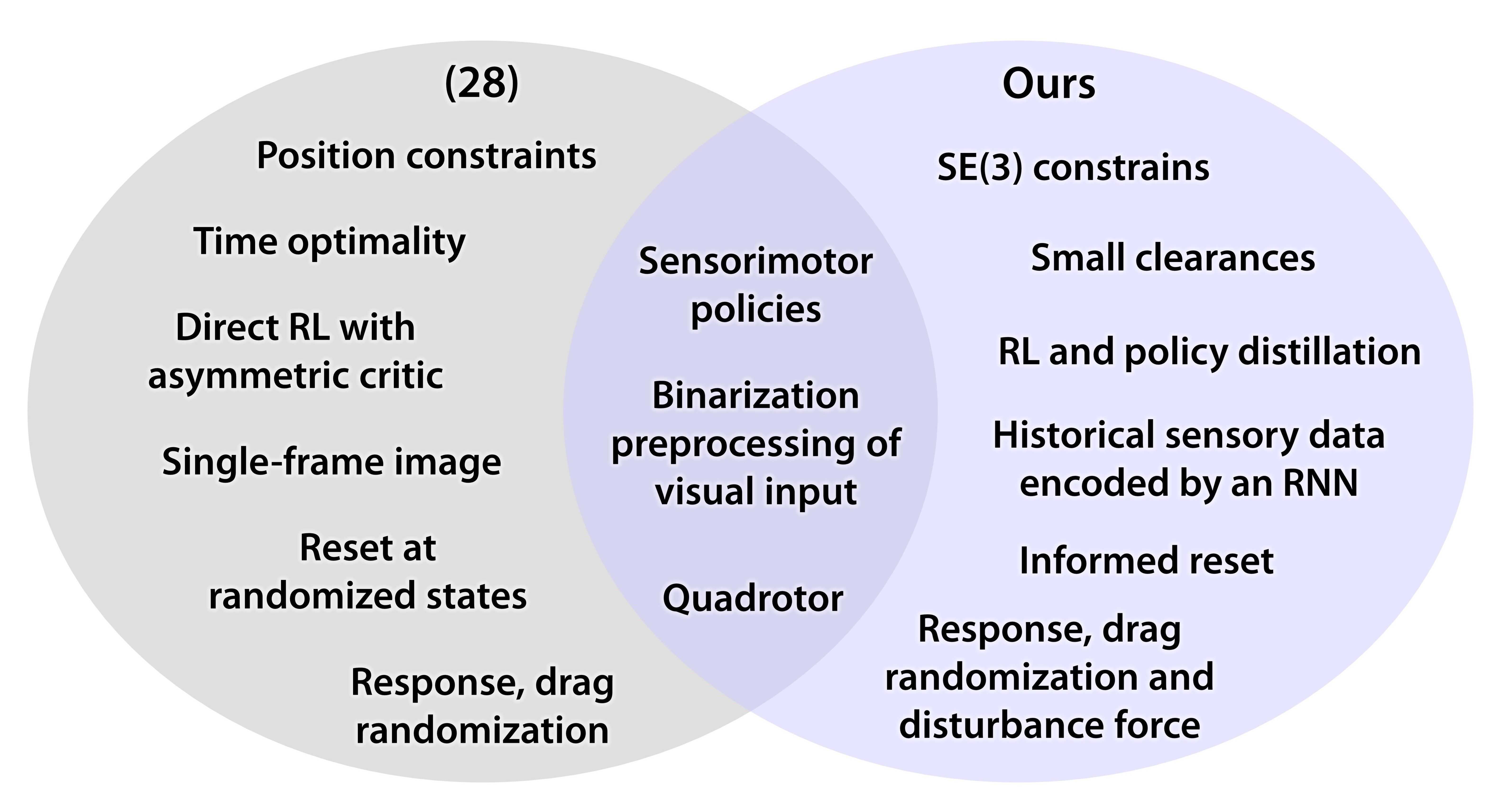}
    \caption{\textbf{Technical comparison between this work and vision-based drone racing \cite{geles2024demonstrating}.} The two ellipses represent the technical components of the two research. The overlapping area indicates commonalities or similarities between the two works, while the non-overlapping area represents their differences.}
    \label{fig:S5}
\end{figure}
\vspace*{\fill}

\newpage

\vspace*{\fill}
\begin{figure}[H]
    \centering  % 添加这行
    \includegraphics[width=0.8\textwidth]{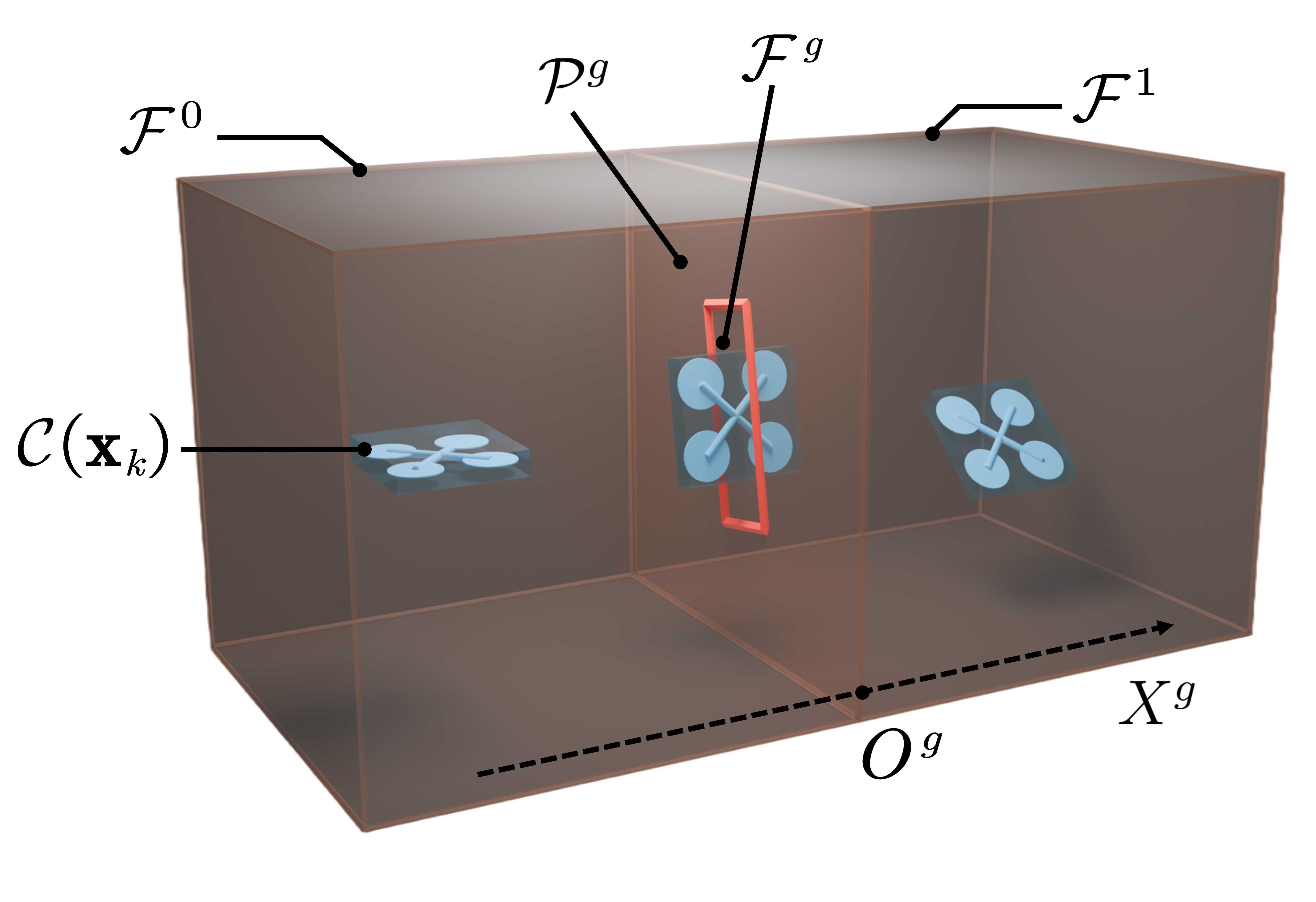}
    \caption{\textbf{Illustration of the geometry relationship of the problem studied in this research.} A coordinate system is established with its axis parallel to the normal vector of the gap plane $\mathcal{P}^g$ to facilitate a clear description of the reward functions.}
    \label{fig:S6}
\end{figure}
\vspace*{\fill}

\newpage

\vspace*{\fill}
\begin{figure}[H]
    \centering
    \includegraphics[width=0.6\textwidth]{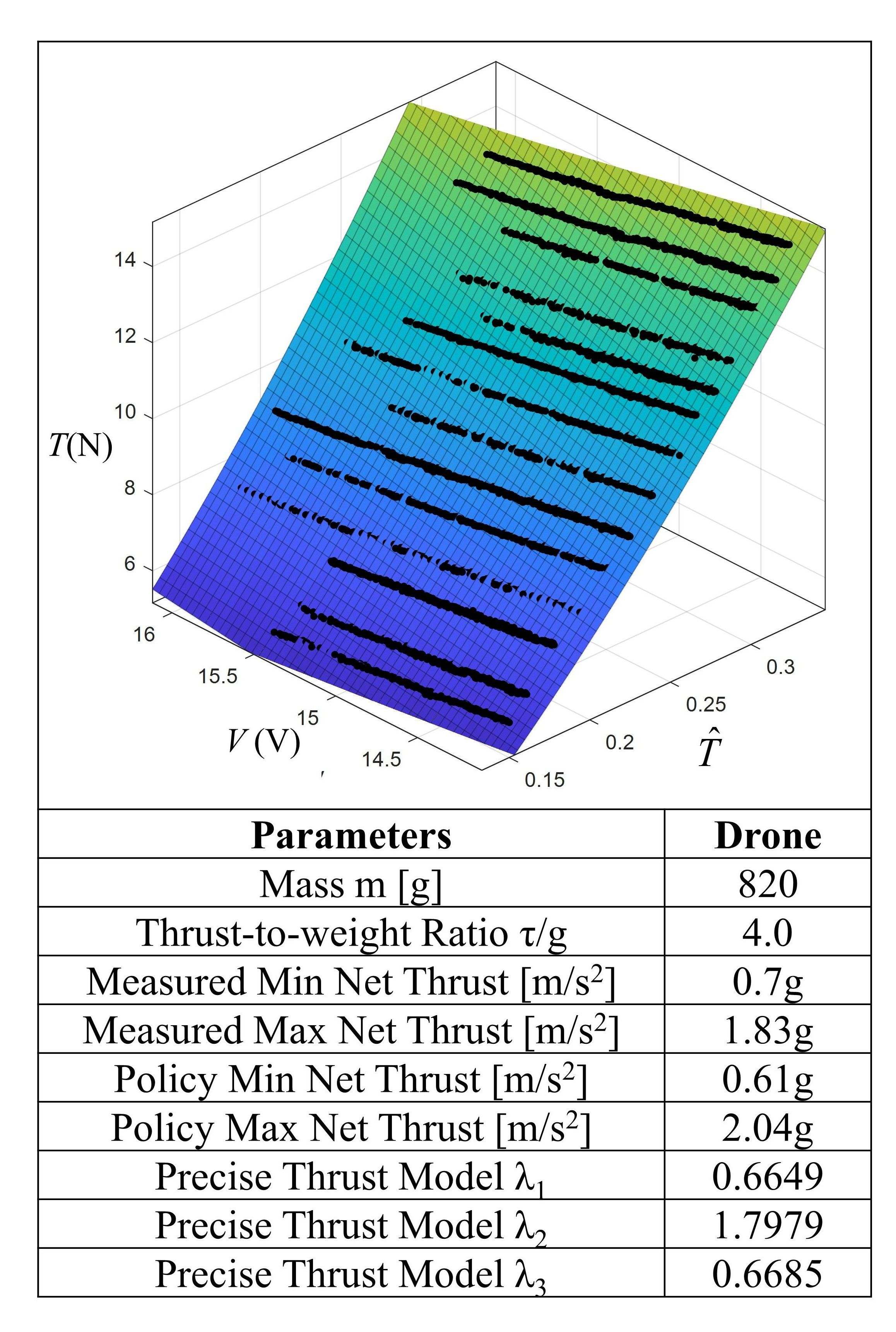}
    \caption{\textbf{Throttle-thrust mapping parameters calibration.} The black points represent recorded calibration data, while the colorful surface illustrates the final fitting results. We list the relevant information about the calibration in a table.}
    \label{fig:S7}
\end{figure}
\vspace*{\fill}

\newpage

\vspace*{\fill}
\begin{figure}[H]
    \centering
    \includegraphics[width=0.85\textwidth]{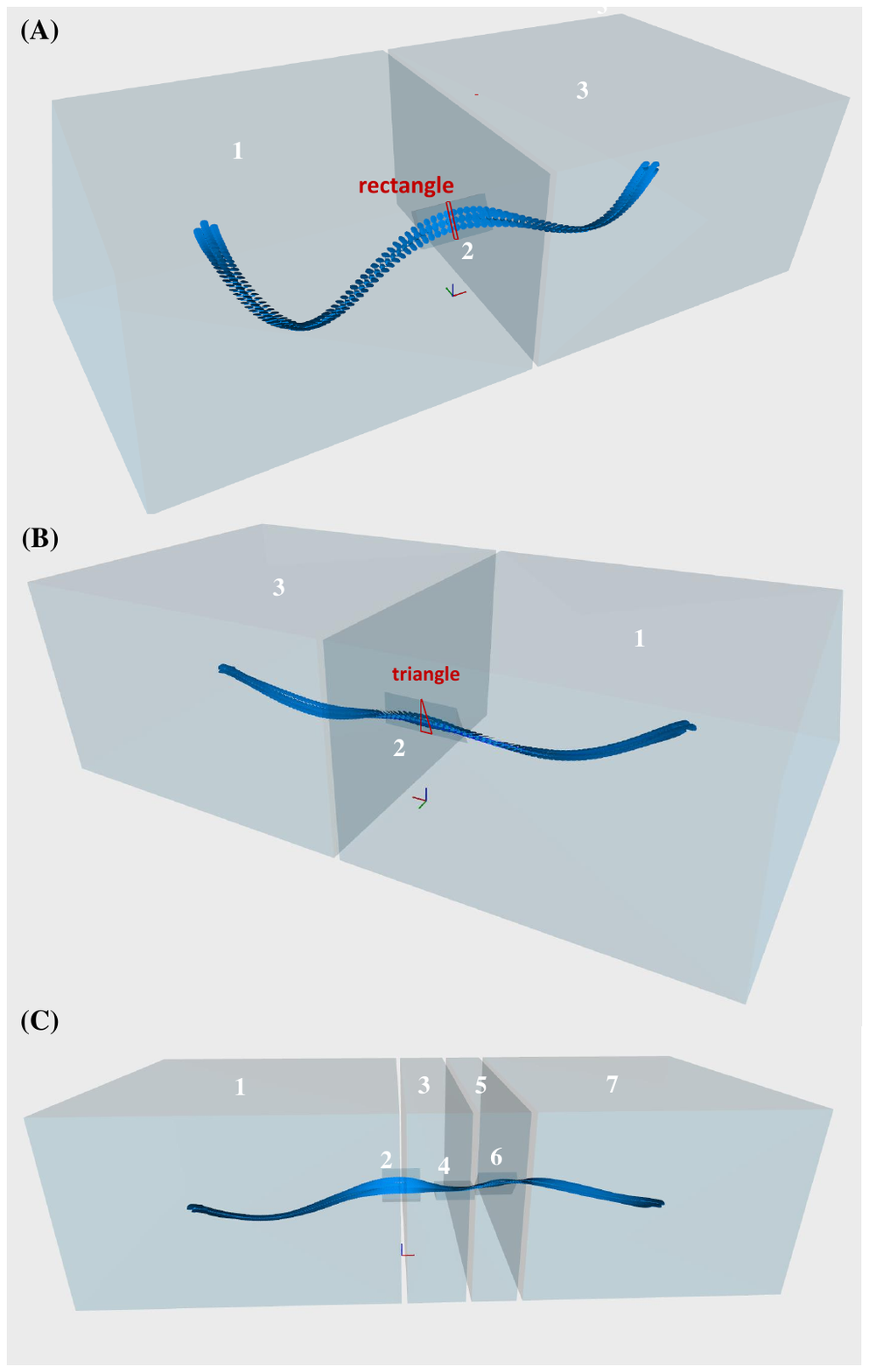}
    \vspace{-1cm}
    \caption{\textbf{Visualization of the SFC layout and planned $\mathrm{SE(3)}$ trajectory.}
    The numbers labeled on the picture represent the indices of the convex polyhedrons that make up the SFC. \textbf{(A)} The SFC and a planned trajectory for traversal through a planar rectangular passable region. \textbf{(B)} The SFC and a planned trajectory for traversal through a planar triangular passable region. \textbf{(C)} The SFC and a planned trajectory for traversal through three consecutive passable regions on planes.}
    \label{fig:S8}
\end{figure}
\vspace*{\fill}

\vspace*{\fill}
\begin{figure}[H]
    \centering
    \includegraphics[width=1.0\textwidth]{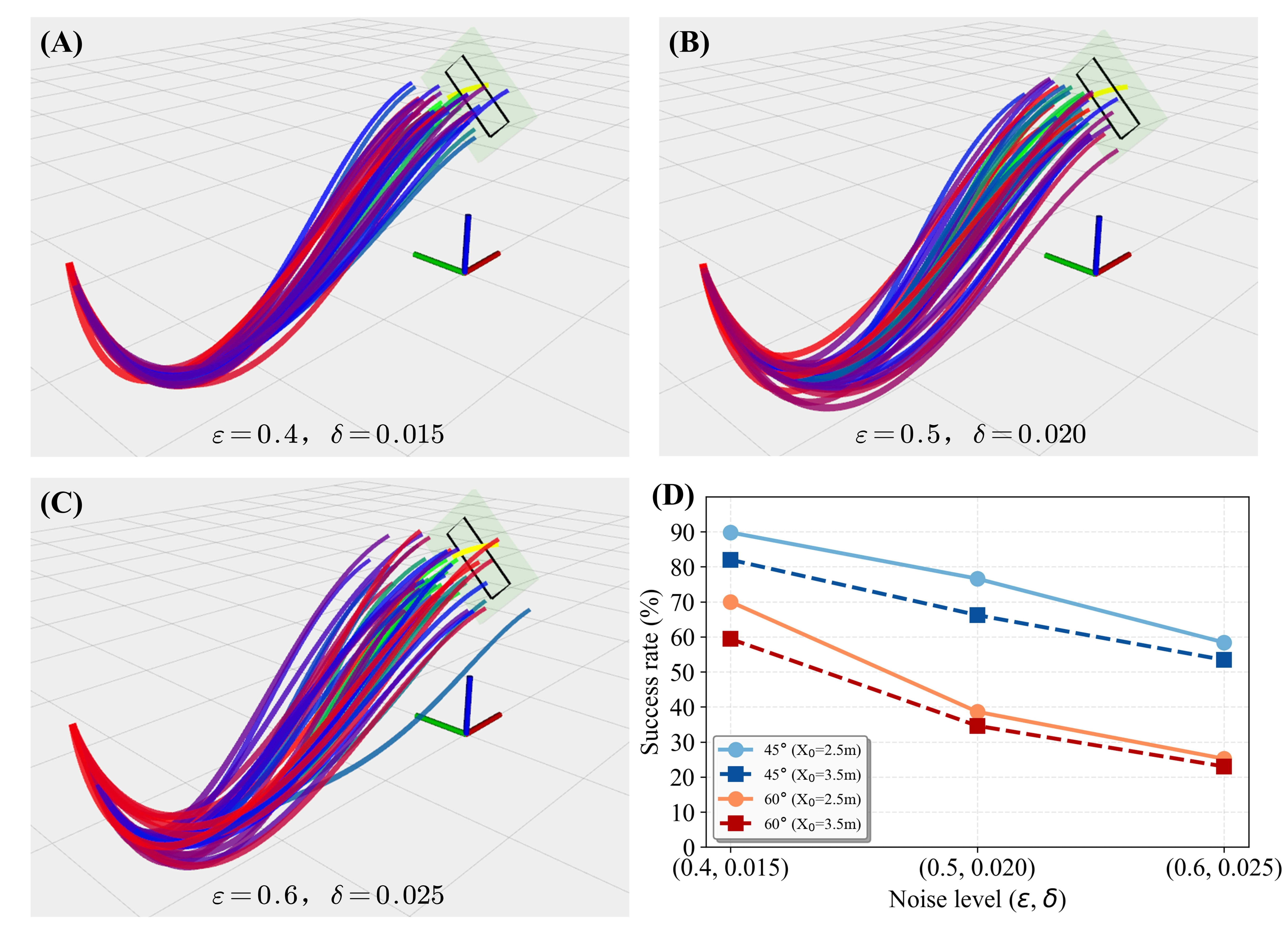}
    \caption{{\textbf{Trajectory generation performance of the planner in \cite{falanga2017aggressive} under simulated state estimation noise.}
     Figures (A), (B), and (C) visualize the replanned trajectories under different state estimation errors, where the initial distance to the gap plane $X_0$ is 2.5 m. Trajectories replanned at different times during flight are represented by distinct colors, while the yellow trajectory at the end of the final successfully planned trajectory denotes the traverse trajectory described in \cite{falanga2017aggressive}. Here, the yaw estimation noise is simulated as $\mathrm{n}_\psi(d)=\mathcal{U}(-\varepsilon 
 d^2, \varepsilon  d^2)$(\degree), and position error is simulated as $\mathrm{n}_p(d)=[\mathcal{U}(-\delta d^2, \delta d^2), \mathcal{U}(-\delta d^2, \delta d^2), \mathcal{U}(-\delta d^2, \delta d^2)]$(m) where $d$ is the distance from the quadrotor to the gap center (unit: meters), $\varepsilon$ and $\delta$ are noise-level coefficients, and $\mathcal{U}(\cdot)$ represents uniform distribution. Figure (D) shows  success rates under different state estimation error levels. The corresponding value $X_0$ at the start point, the tilted angle $\phi_\textnormal{gap}$ of the gap, and the noise coefficients are labeled in the figures.}}
    \label{fig:S9}
\end{figure}
\vspace*{\fill}

\vspace*{\fill}
\begin{figure}[H]
	\centering
	\includegraphics[width=0.9\textwidth]{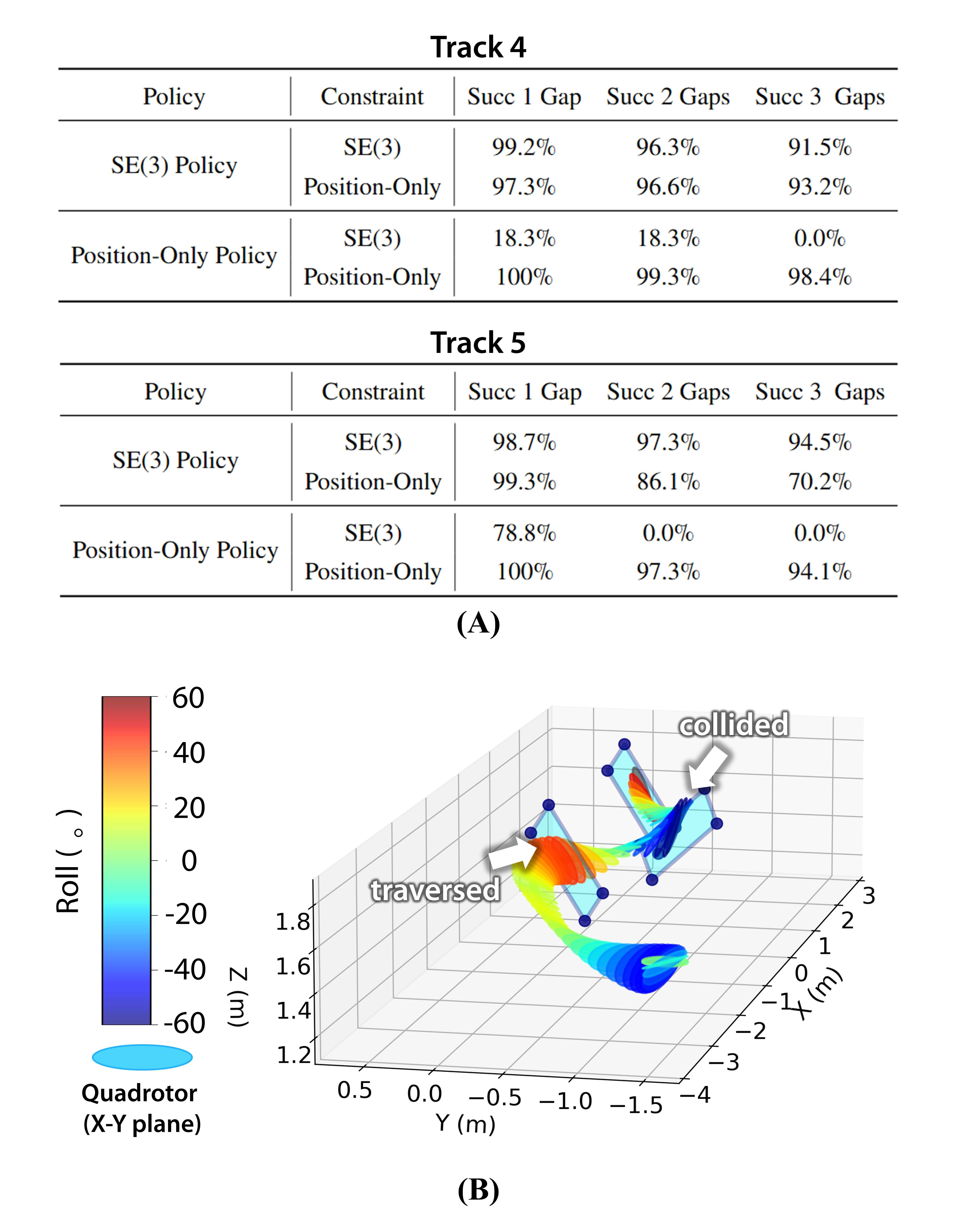}
	\caption{\textbf{Ablation study on SE(3) constraints in policy learning and track configuration.} \textbf{(A)} Success rates of policies trained with and without full SE(3) constraints, evaluated on Track 4 and Track 5 under both SE(3) and position-only success criteria (see Section S10 for details). ``Succ $n$ Gap(s)" denotes the success rate of traversing $n$ consecutive gap(s) within the track. \textbf{(B)} Trajectory visualization of a position-only constrained policy on Track 5. Although the policy is trained without awareness of full SE(3) constraints, the drone successfully traverses the first gap by coincidence, while subsequent gaps result in collision.}
    \label{fig:S95}
\end{figure}
\newpage

\newpage

\vspace*{\fill}
\begin{figure}[H]
    \centering
    \includegraphics[width=1.0\textwidth]{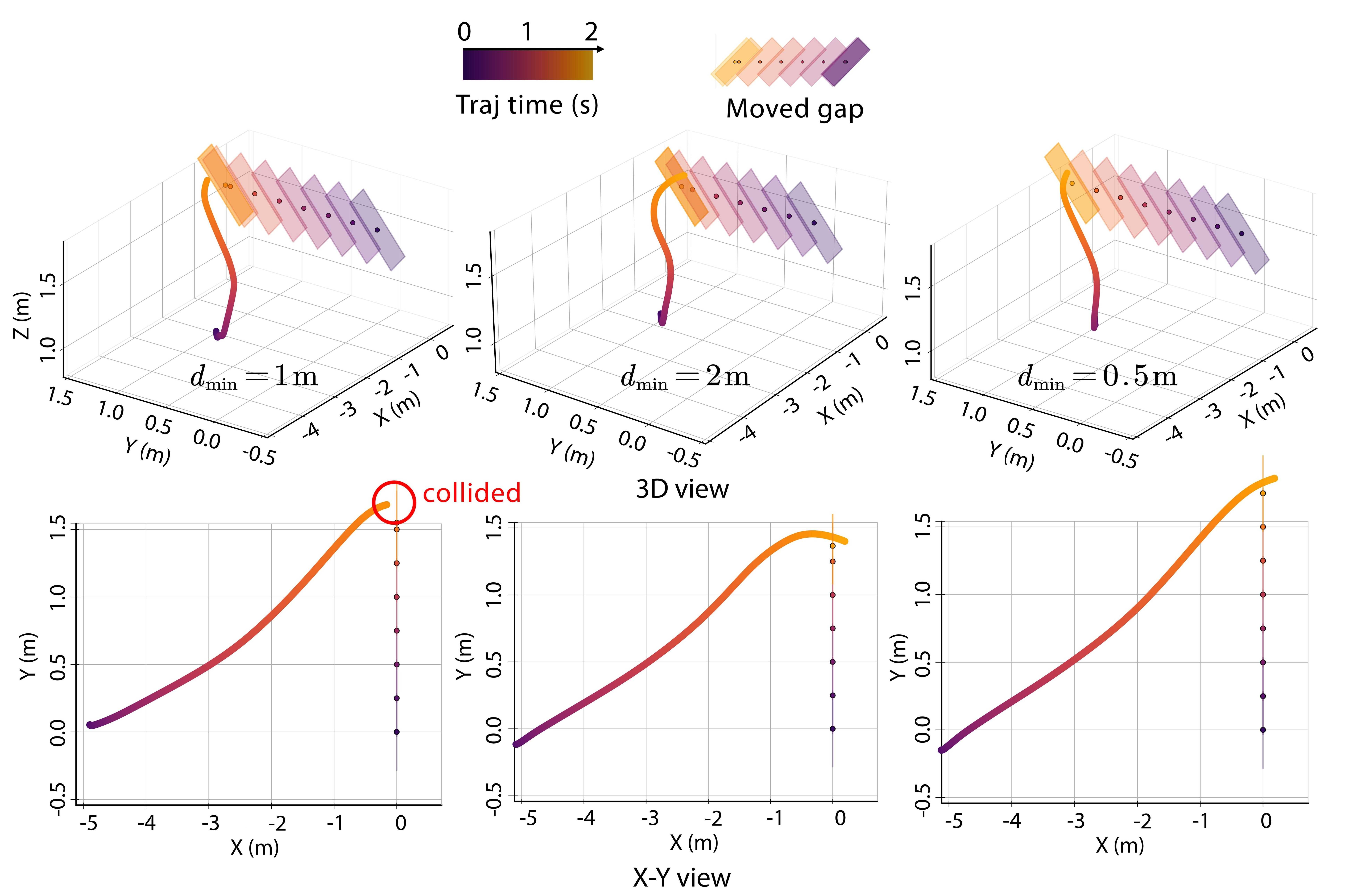}
    \caption{\textbf {Autonomous rollout trajectories with abrupt halting moving gap.}
    The top row of images illustrate the 3D trajectories with $d_\text{min}$ being 1 m (left), 2m (middle) and 0.5 m (right), respectively. The bottom row of images are the corresponding X-Y view of the trajectories.}
    \label{fig:S10}
\end{figure}
\vspace*{\fill}

\newpage

\vspace*{\fill}
\begin{figure}[H]
    \centering
    \includegraphics[width=1.0\textwidth]{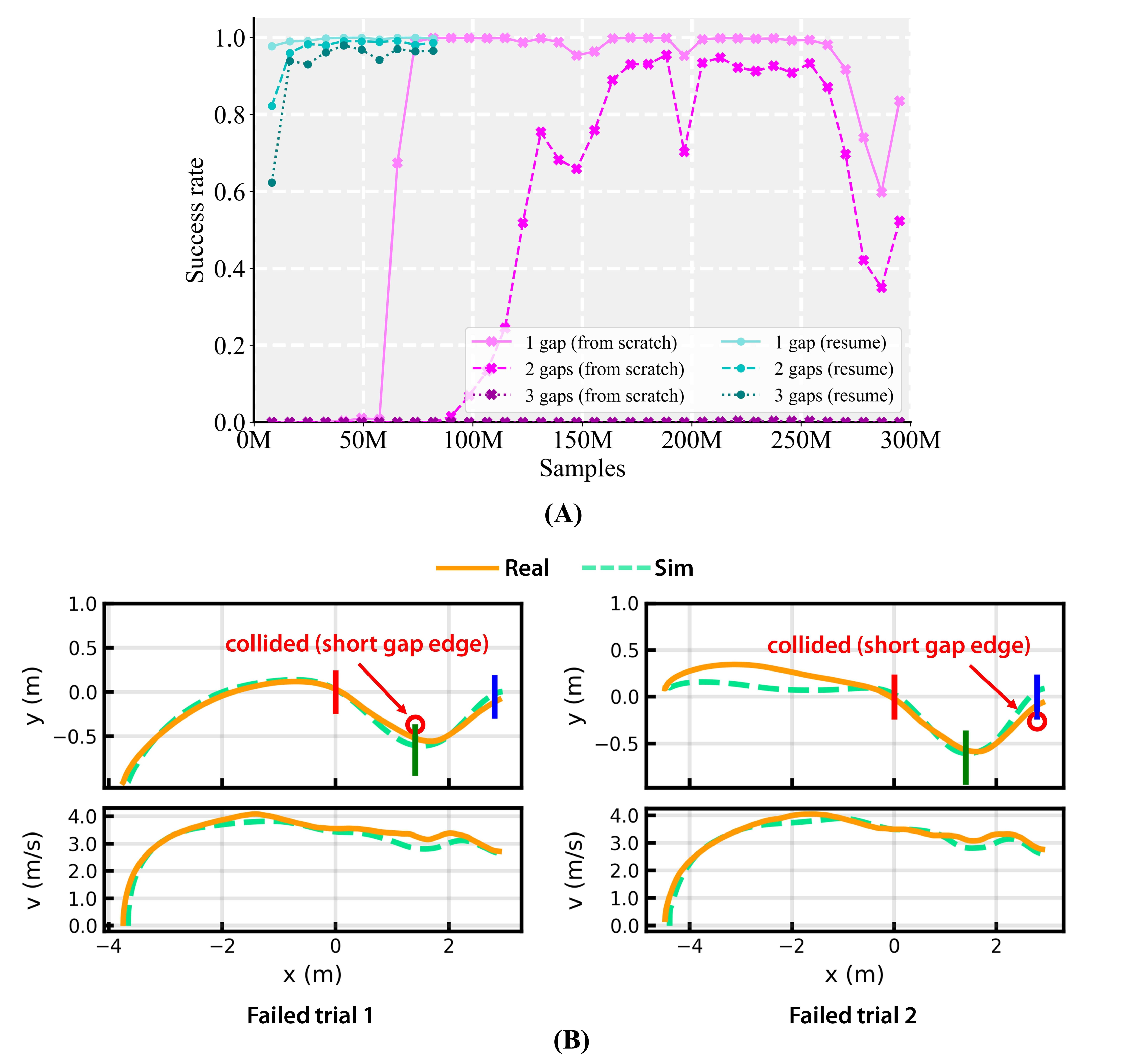}
    % \vspace{0.1cm}
    \caption{\textbf {Failure modes of consecutive gap traversal.}
    (A) Policy learning with Informed Reset on a more challenging variant of Track 5 (see Table \ref{tab:S1}, where the second and third gaps are positioned at x-ranges of 1.1$\sim$1.2 m and 2.3$\sim$2.4 m, respectively. Success rate evolution curves compare training from scratch (donated as ``from scratch") versus curriculum-based initialization using the policy trained on the original Track 5 (donated as ``resume"). (B) Sim-to-real performance degradation on Track 5. Left and right panels show representative failure cases where the quadrotor collides with the short edge of the second and third gaps, respectively.}
    \label{fig:S11}
\end{figure}
\vspace*{\fill}

% \subsection*{Figure S6: Useless of Informed Reset on DAgger.}
\newpage

\vspace*{\fill}
\begin{figure}[H]
    \centering
    \includegraphics[width=0.8\textwidth]{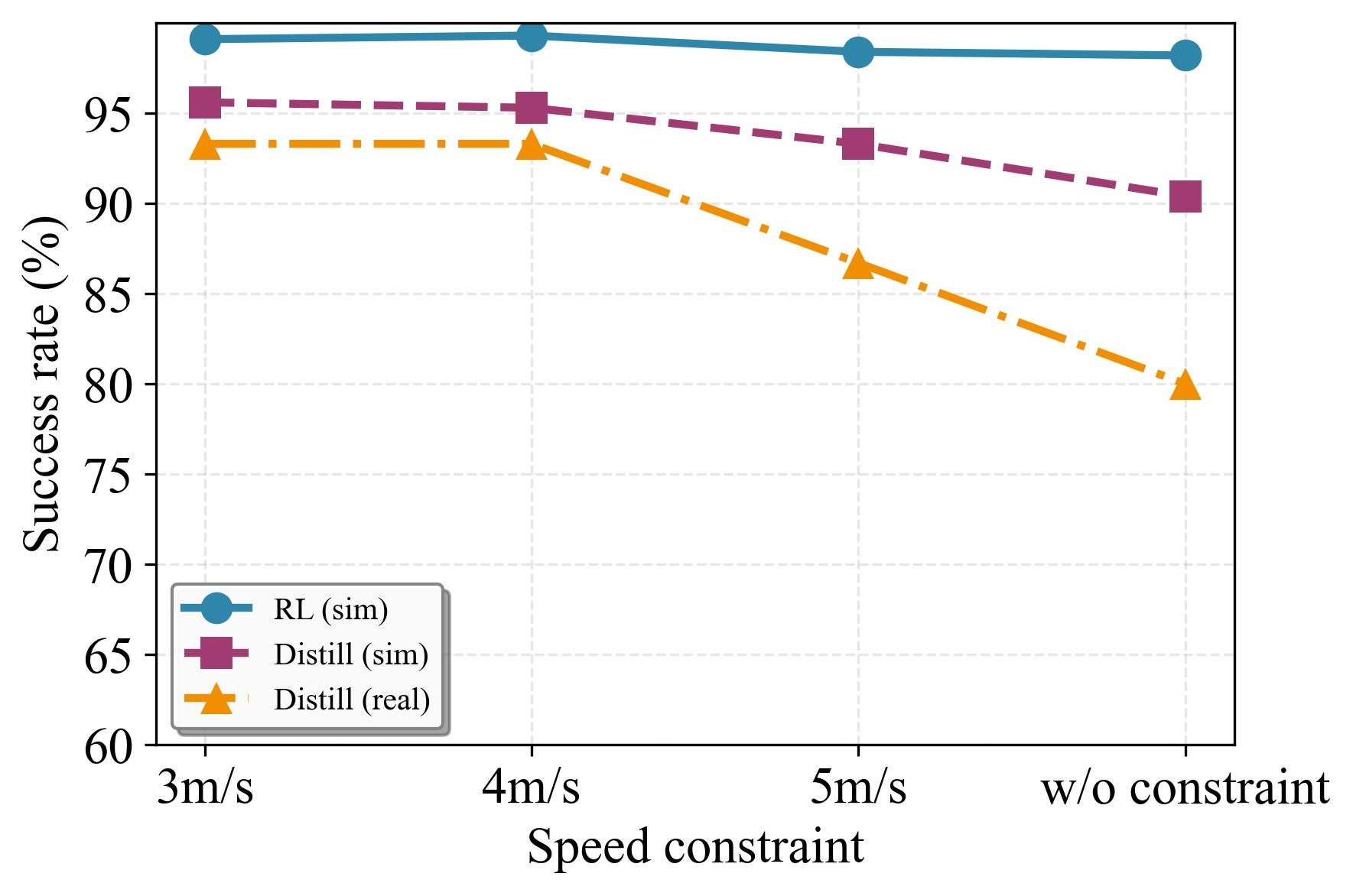}
    \vspace{0.1cm}
    \caption{\textbf {Policy performance with 60\degree~tilted gap under different speed constraints.}
    The numbers labeled on the picture represent the indices of the convex polyhedrons that make up the SFC. \textbf{(A)} The SFC and a planned trajectory for traversal through a planar rectangular passable region. \textbf{(B)} The SFC and a planned trajectory for traversal through a planar triangular passable region. \textbf{(C)} The SFC and a planned trajectory for traversal through three consecutive passable regions on planes.}
    \label{fig:S12}
\end{figure}
\vspace*{\fill}

\newpage

\vspace*{\fill}
\begin{table}[H]
\centering
\caption{\textbf{Training configurations for different tracks in the \emph{Traversal through Consecutive Narrow Gaps with Low Clearances} section.} The x-direction of the following coordinate information is defined as the normal direction of the gap plane, with the positive direction pointing from one side of the starting position to the other. The positive direction of the z-axis is opposite to the direction of gravity and, together with the positive directions of the x-axis and y-axis, forms a right-handed coordinate system. Assume that the ground on which the frame is placed is perpendicular to the direction of gravity and that each gap plane is parallel to each other. The following coordinate information contains the distribution range of the x, y, and z axis coordinates.}
\begin{tabular}{c|c|c|c}
\toprule
Track & Gap & Position ($\mathrm{m}$) & Gap roll ($\mathrm{rad}$) \\
\midrule
\multirow{2}{*}{Track 1} 
& Gap 1 & $(0, 0, 1.5)$ & $\pi/4.3 \sim \pi/3.7$ \\
& Gap 2 & $(0.80\sim0.90, -0.05\sim0.05, 1.45\sim1.55)$ & $\pi/7 \sim \pi/6$ \\
\midrule
\multirow{2}{*}{Track 2} 
& Gap 1 & $(0, 0, 1.5)$ & $\pi/3.3 \sim \pi/2.7$ \\
& Gap 2 & $(0.85\sim0.95, -0.10\sim0.00, 1.35\sim1.45)$ & $\pi/6.5 \sim \pi/5.5$ \\
\midrule
\multirow{2}{*}{Track 3} 
& Gap 1 & $(0, 0, 1.5)$ & $\pi/4.3 \sim \pi/3.7$ \\
& Gap 2 & $(1.30\sim1.40, -0.75\sim-0.65, 1.45\sim1.55)$ & $-\pi/5.8 \sim -\pi/6.2$ \\
\midrule
\multirow{3}{*}{Track 4} 
& Gap 1 & $(0, 0, 1.5)$ & $\pi/4 \sim \pi/3.5$ \\
& Gap 2 & $(1.00\sim1.10, -0.10\sim0.00, 1.35\sim1.45)$ & $-\pi/18 \sim -\pi/36$ \\
& Gap 3 & $(1.80\sim1.85, 0.00\sim0.05, 1.35\sim1.40)$ & $-\pi/3.7 \sim -\pi/4.3$ \\
\midrule
\multirow{3}{*}{Track 5} 
& Gap 1 & $(0, 0, 1.5)$ & $\pi/4.3 \sim \pi/3.7$ \\
& Gap 2 & $(1.3\sim1.4, -0.70\sim-0.60, 1.40\sim1.45)$ & $-\pi/5.8 \sim -\pi/6.2$ \\
& Gap 3 & $(2.7\sim2.8, -0.05\sim0.05, 1.45\sim1.55)$ & $\pi/4.3 \sim \pi/3.7$ \\
\midrule
\multirow{3}{*}{Track 6} 
& Gap 1 & $(0, 0, 1.5)$ & $\pi/6.2 \sim \pi/5.8$ \\
& Gap 2 & $(0.9\sim0.95, -0.45\sim-0.40, 1.45\sim1.55)$ & $-\pi/5.8 \sim -\pi/6.2$ \\
& Gap 3 & $(1.75\sim1.8, -0.05\sim0.00, 1.45\sim1.55)$ & $\pi/6.2 \sim \pi/5.8$ \\
\bottomrule
\end{tabular}
\label{tab:S1}
\end{table}
\vspace*{\fill}
\newpage

\vspace*{\fill}
\begin{table}[H]
\centering
\caption{\textbf{Geometry of the gaps in the \emph{Traversal through Narrow Passable Regions with Various Geometries} section.} The 2D coordinates in the table describe the geometric parameters of the passable region corresponding to each gap.}

\begin{tabular}{|m{3cm}|m{10cm}|}
    \hline
    \centering\textbf{Gap} & \centering\textbf{Geometry} \tabularnewline
    \hline
    \centering Triangle & \centering\includegraphics[width=10cm]{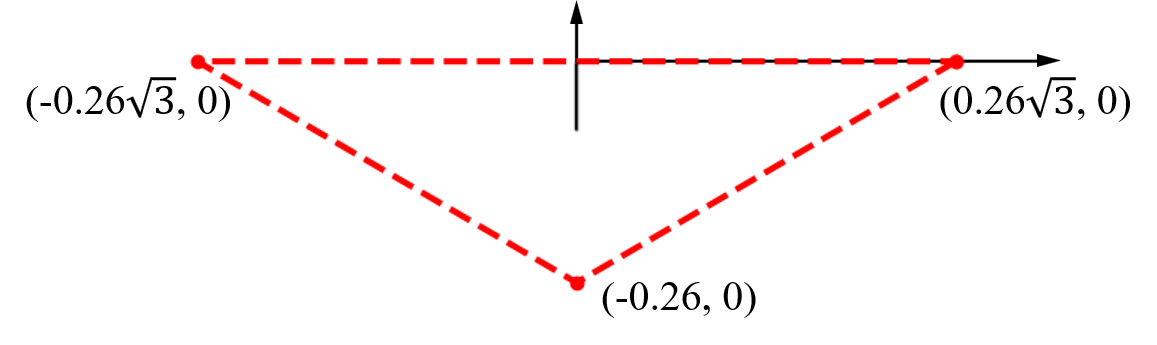} \tabularnewline
    \hline
    \centering Parallelogram & \centering\includegraphics[width=10cm]{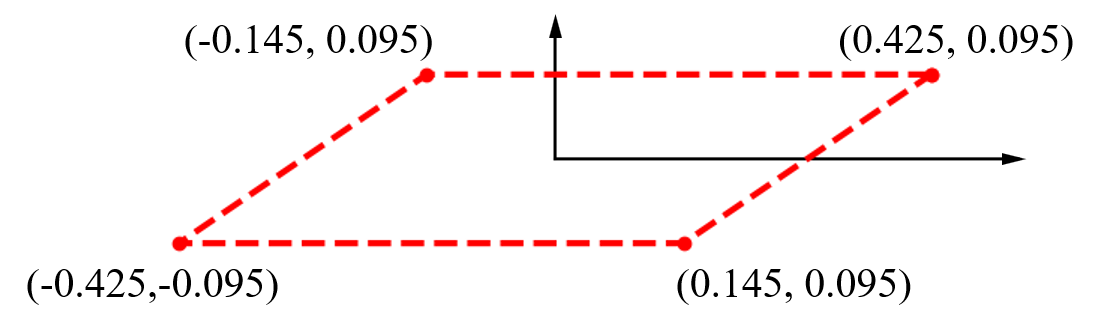} \tabularnewline
    \hline
    \centering Ellipse & \centering\includegraphics[width=9cm]
    {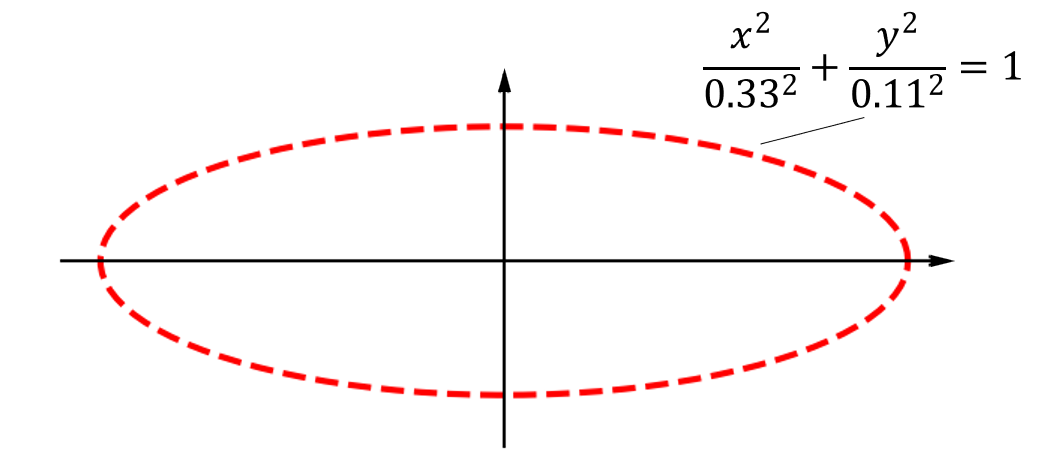} \tabularnewline
    \hline
    \centering Diamond & \centering\includegraphics[width=10cm]{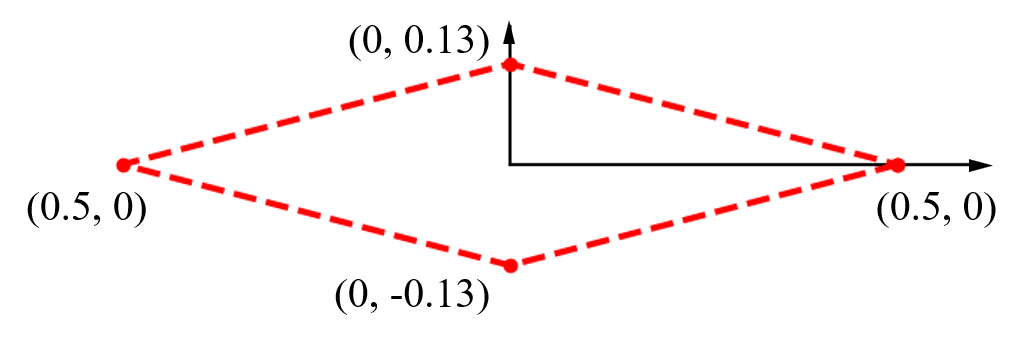} \tabularnewline
    \hline
    \centering Arch & \centering\includegraphics[width=10cm]
    {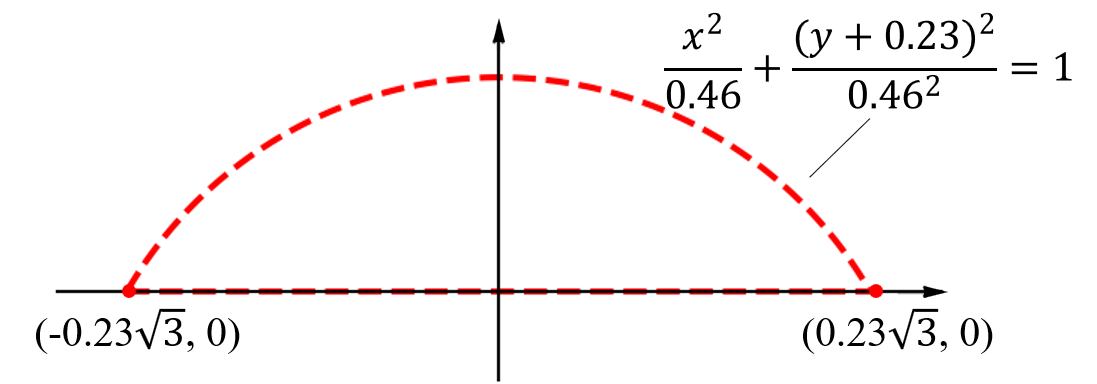} \tabularnewline
    \hline
\end{tabular}
\label{tab:S2}
\end{table}
\vspace*{\fill}

\newpage

\begin{table}[p]
\begin{center}
\caption{\textbf{Hyperparameter setup of policy optimization algorithms.} The annotation (single/consecutive) indicates that the parameters for the single gap traversal task and the consecutive gap traversal are different.}
\begin{tabular}{cccc}
\hline
\multicolumn{4}{c}{\textbf{PPO}} \\
\hline
\textbf{Hyperparameter} & \textbf{Value} & \textbf{Hyperparameter} & \textbf{Value} \\
\hline
\# of envs & 2048 & batch size & 100000 (steps) \\
$\lambda_\mathrm{GAE}$ & 0.95 & Reward discount & 0.99 \\
Actor learning rate & 0.0003 & Critic learning rate & 0.0005 \\
 State normalization trick \cite{engstrom2020implementation} & True & Gradient clip norm \cite{engstrom2020implementation} & 1.5 \\
 \begin{tabular}[c]{@{}c@{}}\# of rollout steps of each env \\ between networks updates\end{tabular} & 200 &  & \\
\hline
\multicolumn{4}{c}{\textbf{DAgger}} \\
\hline
\# of envs (single/consecutive) & 275/200 & batch size (single/consecutive) & 25/22 (envs) \\
learning rate (single/consecutive) & 0.0002/0.0001 &  \begin{tabular}[c]{@{}c@{}}\# of rollout steps of each env \\ between network updates\end{tabular} & 135/180\\
 \begin{tabular}[c]{@{}c@{}}\# of learning epochs\\ in each network update\end{tabular} & 3 & Gradient clip norm & 3.0 \\
\hline \\

\end{tabular}
\label{tab:S3}
\end{center}
\end{table}
\newpage

\newcolumntype{Y}{>{\centering\arraybackslash}X}
\newcolumntype{C}[1]{>{\centering\arraybackslash}p{#1}}
\newcommand{\bcheck}{$\boldsymbol{\checkmark}$}
\newcommand{\bcross}{$\boldsymbol{\times}$}
\newcommand{\thickhline}{
	% 创建额外的垂直空间，确保竖线断开
	\noalign{\vskip 2pt}  
	% 设置为粗线
	\noalign{\global\arrayrulewidth=1.6pt}
	\hline
	% 恢复正常线宽
	\noalign{\global\arrayrulewidth=0.4pt}
	% 创建额外的垂直空间，确保竖线断开
	\noalign{\vskip 2pt}
}
                          % 使用小号字体
	
\begin{table}[htbp]
	\centering
    \caption{Success rates of unified and separate student policies at various tilt angles. Starting from the second column, each column represents the success rates at the specified tilt angle.}
	\setlength{\tabcolsep}{8pt}      % 设置列间距
	\setlength{\extrarowheight}{2pt} % 增加行高
	\small                           % 使用小号字体
	
	\begin{tabular}{lcccccc}
		% 第一根粗横线
		\thickhline
		
		% 表头
		Policy & Roll $30\degree$ & Roll $60\degree$ & Roll $80\degree$ & Pitch $20\degree$ & Pitch $45\degree$ & Pitch $60\degree$ \\
		
		% 第二根粗横线
		\thickhline
		
		% 数据行
		Unified & 100\% & 95.7\% & 88.6\% & 90.8\% & 76.8\% & 74.4\% \\
		
		Separate & 100\% & 97.5\% & 92.3\% & 100\% & 81.8\% & 76.8\%\\
		
		% 普通中间横线
		\noalign{\vskip 2pt}
		\thickhline
	\end{tabular}
	\label{tab:S4}
\end{table}
\newpage

\renewcommand{\thealgorithm}{S\arabic{algorithm}}

\begin{algorithm}[p]
\caption{Online Learning in MDP with Informed Reset}
\label{alg:ir}
\begin{algorithmic}[1]
  \Require informative states buffer $\mathcal{B}$ including quadrotor's states and gap poses, task space $D$, informative state sampling probability $p$, horizon $H$, maximum iteration number $k_{\textnormal{max}}$, $\dots$ 
  \Ensure policy $\pi$
  \For{iteration $k \in 0,1,\dots, k_{\textnormal{max}}$}
        \For{step $i \in 0,1,\dots, H$}
            \State rollout $\pi^{(k)}$ to get
              $(s_{i},a_{i},r_{i}, \textnormal{done}_i)$ sample
            \State append the sample to rollout buffer $\mathcal B_{\pi}$
            \If {$\textnormal{done}_i$}
              \If {$u \in \mathcal{U}(0,1) < p$}
                \State sample a state from $\mathcal{B}$ to reset the agent and  environment  \texttt{// informed reset}
              \Else
                \State randomly sample a state and gap pose in $D$ to reset the agent and the environment
               \EndIf
            \EndIf
        \EndFor
      \State update $\pi^{(k)}$ (and value network $V$ if necessary) using data in $\mathcal B_{\pi}$
  \EndFor
  \State \Return $\pi$
\end{algorithmic}
\end{algorithm}
\newpage

\end{document}